\documentclass[chap]{thesis}
\usepackage{url}
\usepackage{tabularx}
\usepackage{adjustbox}
\usepackage{booktabs}
\usepackage{wrapfig}
\usepackage{listings, paralist}
\usepackage[titletoc]{appendix}
\usepackage{float}
\usepackage{textcase}
\usepackage{notoccite}

\usepackage{amsthm}
\theoremstyle{definition}
\newtheorem{definition}{Definition}[section]

\newcommand{\numproto}{$20$}
\newcommand{\dm}{\texttt{T2DM}}
\newcommand{{\ckd}}{\texttt{CKD}}

\usepackage{longtable}
\usepackage{multirow}
\usepackage{xurl}

\usepackage{color,soul}
\definecolor{lightBlue}{RGB}{0, 153, 255}
\definecolor{lightRed}{RGB}{204, 0, 255}

\usepackage{algorithm} 
\usepackage{algpseudocode}
\usepackage{amsmath}
\usepackage{soul,color}
\usepackage{tcolorbox}

\soulregister\cite7
\soulregister\ref7
\soulregister\pageref7

\algdef{SE}[VARIABLES]{Variables}{EndVariables}
   {\algorithmicvariables}
   {\algorithmicend\ \algorithmicvariables}
\algnewcommand{\algorithmicvariables}{\textbf{global variables}}

\newcommand\blfootnote[1]{%
  \begingroup
  \renewcommand\thefootnote{}\footnote{#1}%
  \addtocounter{footnote}{-1}%
  \endgroup
}

\algblock{Input}{EndInput}
\algnotext{EndInput}
\algblock{Output}{EndOutput}
\algnotext{EndOutput}
\newcommand{\Desc}[2]{\State \makebox[6em][l]{#1}#2}
\newcommand{\explanation}{\mathcal{E}}

\lstdefinelanguage{Manchester}
{
    sensitive = true,
    keywords = [1]{Class, EquivalentTo, SubClassOf},
    morekeywords = [2]{and, or},
    morekeywords = [3]{some, value},
    keywordstyle=[2]\textbf,
    keywordstyle=[2]\color{lightBlue}\textbf,
    keywordstyle=[3]\color{lightRed}\textbf,
    morestring=[b]''
}

\begin{document}
%
    
%
\thesistitle{\bf An Ontology-Enabled Approach For User-Centered and Knowledge-Enabled Explanations of AI Systems}        
\author{Shruthi Chari}        
\degree{Doctor of Philosophy}        
\department{Computer Science} 
     
\signaturelines{5}     
\thadviser{Deborah L. McGuinness}
\cothadviser{Oshani Seneviratne}    

\memberone{James A. Hendler}        
\membertwo{Pablo Meyer}
\memberthree{Prithwish Chakraborty}

\submitdate{August 2024}
\copyrightyear{2024}   
\titlepage   
\copyrightpage         
\tableofcontents        
\listoftables          
\listoffigures         

 
\specialhead{ACKNOWLEDGMENT}
 
This thesis was funded under two project grants: the IBM-RPI Health, Empowerment by Analytics, Learning and Semantics (HEALS) project, and the IARPA HIATUS grant. 

\textbf{Funding Acknowledgments:}

This work is partially supported by IBM Research AI through the AI Horizons Network. Also, this research is partially supported in part by the Office of the Director of National Intelligence (ODNI), Intelligence Advanced Research Projects Activity (IARPA), via the HIATUS Program contract \#2022-22072200002. The views and conclusions contained herein are those of the authors and should not be interpreted as necessarily representing the official policies, either expressed or implied, of ODNI, IARPA, or the U.S. Government. The U.S. Government is authorized to reproduce and distribute reprints for governmental purposes, notwithstanding any copyright annotation. 

\textbf{Professional and Personal Acknowledgements:}

I thank my committee members, my advisor, Prof. Deborah L. McGuinness, my co-advisor, Prof. Oshani Seneviratne, Dr. James A. Hendler, Dr. Pablo Meyer, and Dr. Prithwish Chakraborty for their valuable inputs, direction and collaboration. I have had the pleasure of working with and being advised by all my committee members. I thank my excellent collaborators over the years from multiple teams at the Center for Computational Health (CCH) at IBM Research including Dr. Amar K. Das, Dr. Daniel M. Gruen, Morgan Foreman, Daby Sow, Dr. Ching-Hua Chen, Dr. Pablo Meyer, Dr. Prithwish Chakraborty, Dr. Mohammed Ghalwash, Dr. Fernando Suarez-Saiz and Dr. Elif Eyigoz, RPI including Dr. Henrique Santos, Dr. Jamie McCusker, Dr. Sabbir Rashid, Dr. Mia Qi, Dr. Nneka N. Agu, Gregorios Katsios and Dr. Neha Keshan, and LEIDOS including Dr. Noelie Creaghe and Dr. Alex Rosenfield, for their excellent inputs and direction.

Finally, I am grateful to my family, including my mother, Rekha Chari; my father, G. Srishail Chari; my brother, Sanjay Chari; my husband, Vijay Sadashivaiah; my in-laws Uma and Sadashivaiah and brother-in-law, Vishwas Sadashivaiah, for supporting me. I am incredibly fortunate to be surrounded by people with career interests/careers in Computer Science in my closest family circles, including my parents, husband, and brothers; it helped me learn from their strengths and also helped me have conversations about fundamental topics when need be. I am also grateful to my grandparents, Captain C.V. Prabhakar and Kumuda Prabhakar and Pushpa Bhashyam, who have passed, and I dedicate this thesis to them. I would also like to thank my friends and extended family for providing me with support and company when I needed it. Finally, I am fortunate to have persevered through this PhD journey; I have learned a lot and grown as a researcher, and I am excited to contribute to trustworthy applications of AI.   
 
\specialhead{ABSTRACT}
Explainability has been a well-studied problem in the Artificial Intelligence (AI) community through several AI ages, ranging from expert systems to the current deep learning era, to enable AI's safe and robust use. Through the ages, the unique nature of AI approaches and their applications have necessitated the explainability approaches to evolve as well. However, these multiple iterations of explaining AI decisions have all focused on helping humans better understand and analyze the results and workings of AI systems. 

In this thesis, we seek to further the user-centered explainability sub-field in AI by addressing several challenges around explanations in the current AI era, which is characterized by the availability of many machine learning (ML) explainers and neuro-symbolic AI approaches. Previous research in user-centered explainability has mainly focused on what needs to be explained and less so on implementations for them. Additionally, there are challenges to supporting explanations in a manner that humans can easily interpret due to the lack of a unified framework for different explanation types and methods to support domain knowledge from authoritative literature. We address the three challenges or research questions around user-centered explainability: \textit{How can we formally represent explanations with support for interacting AI systems (AI methods in applied ecosystem), additional data sources, and along different dimensions?} \textit{How useful and feasible are such explanations for clinical settings? Is it feasible to combine explanations from multiple data modalities and AI methods?} 

For the first research question, we design an Explanation Ontology (EO), a general-purpose semantic representation that can represent fifteen different literature-derived explanation types via their system-, interface- and user- related components. We demonstrate the utility of the EO in representing explanations across different use cases, supporting system designers to answer explanation-related questions via a set of competency questions, and categorizing explanations to be of supported explanation types within our ontology. 

For the second research question,  we focus on key explanation dimension, that is, contextual explanation and conduct a case study on supporting contextual explanations from an authoritative knowledge source, clinical practice guidelines (CPGs). Here, we design a clinical question-answering (QA) system to address CPG questions to provide contextual explanations to help clinicians interpret risk prediction scores and their post hoc explanations in a comorbidity risk prediction setting. For the QA system, we leverage large language models (LLMs) and their clinical variants and implement knowledge augmentations to these models to improve semantic coherence of the answers. We evaluate both the feasibility and value of supporting these contextual explanations. For feasibility, we use quantitative metrics to report the performance of the QA system and do so across different model settings and data splits. To evaluate the value of the explanations, we report findings from showing the results of our QA approach to an expert panel of clinicians. 

Finally, for the last research question, we design a general-purpose and open-source framework, Metaexplainer, capable of providing natural-language explanations to a user question from the several explainer methods registered to generate explanations of a particular type. The Metaexplainer is a three-stage (Decompose, Delegate, and Synthesis) modular framework through which each stage produces intermediate outputs that the next stage ingests. In the Decompose stage, we input user questions and identify what explanation type can best address them and generate actionable machine interpretations; in the Delegate stage, we run explainer methods registered for the identified explanation type and pass filters if any, from the question and finally, in the synthesis stage we generate natural-language explanations along the explanation type template. For the Metaexplainer, we leverage LLMs, the EO, and explainer methods to generate user-centered explanations in response to user questions. We evaluate the Metaexplainer on open-source tabular datasets, but the framework can be applied to other modalities with code adaptations. 

Overall, through this thesis, we aim to design methods that can support knowledge-enabled explanations across different use cases, accounting for the methods in today's AI era that can generate the supporting components of these explanations and domain knowledge sources that can enhance them. We demonstrate the efficacy of our approach in two clinical use cases as case studies but design our methods to be applied to use cases outside of healthcare as well. By implementing approaches for knowledge-enabled explainability that leverage the strengths of symbolic and neural AI, we take a step towards user-centered explainability to help humans interpret and understand AI decisions from different perspectives.  




 
\chapter{INTRODUCTION}
 
Artificial Intelligence (AI) has evolved over the years from having limited applications in originally thought-about application domains such as the military to having more widespread use to being available to assist humans in both high-precision tasks such as healthcare, military and financial decisions to more everyday tasks such as helping in web search, weather alerts to navigation. Through these applications and improvements in computing technology and AI research, AI methods have evolved to support various applications and better computing infrastructure. Moreover, with the evolution of AI, different methods have emerged, such as rule-based expert systems to pattern-based machine learning (ML) and deep learning methods. In today's AI ecosystem, we see a co-existence of both neural and symbolic approaches or neurosymbolic approaches. As humans, we tend to trust AI better if we can understand the reasoning of how the AI came to a decision or connect an AI decision to what we are familiar with~\cite{swartout1991explanations},~\cite{miller2019explanation}. 
Clearly, explainability or explainable AI (XAI) has been one of the first conceived thrusts of what we know today as trustworthy AI~\cite{trustworthy-ai-nist}, from early works such as Mycin~\cite{shortliffe1974mycin} that were developed to explain an expert system in a medical setting to now the plethora of post-hoc explainability methods that provide reasoning for features that were found to be important by somewhat opaque ML models. In essence, explanations in AI have had to evolve in approaches and outputs with the advancements in AI methods.  

\begin{figure}
\centering
\includegraphics[width=0.7\linewidth]{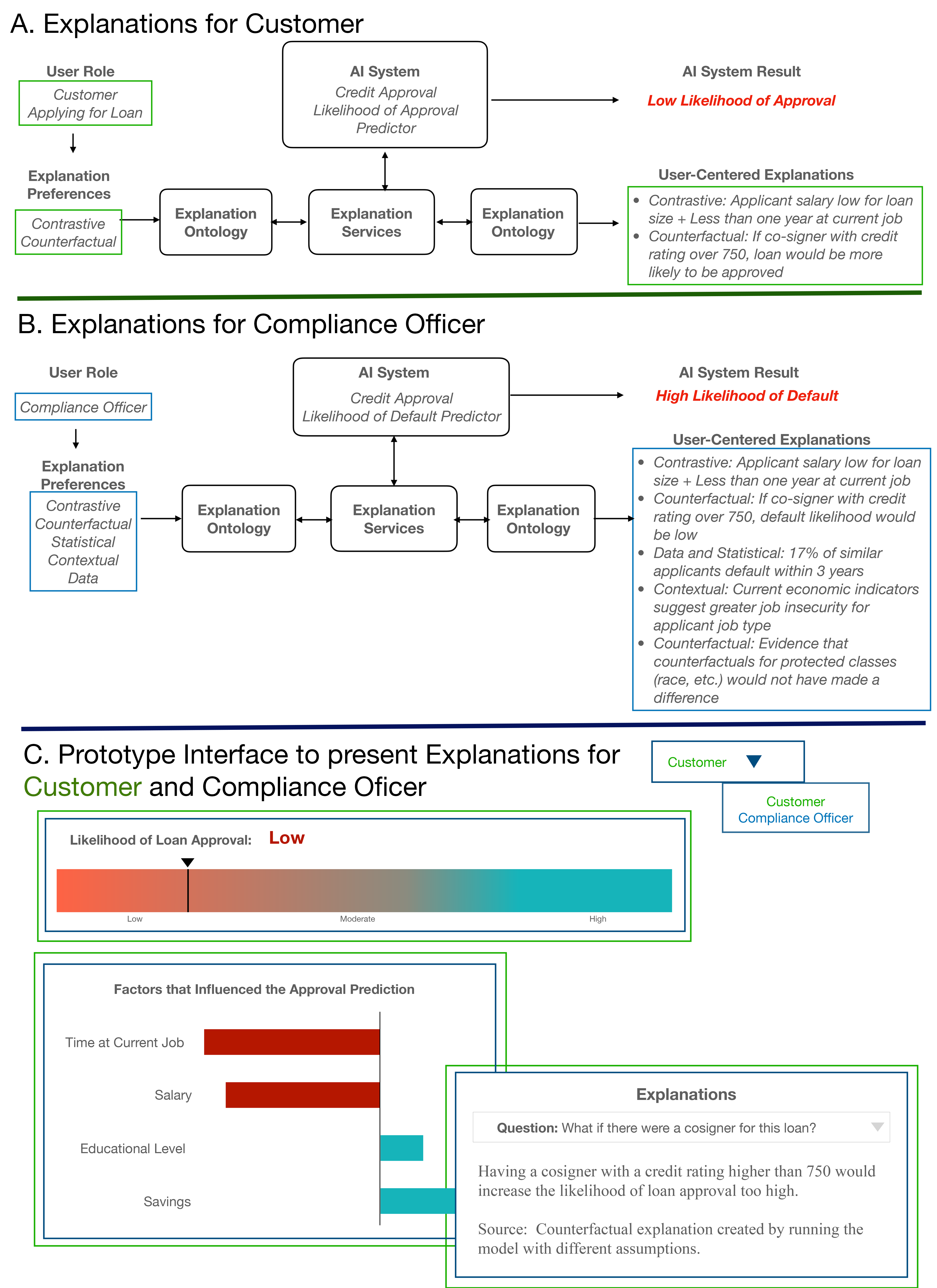}
\caption{Users needs for explainability are diverse and depend on the use case and question being asked by domain experts / end-users in that use case. Illustrated here is different needs for explanations by two different user groups of A) Customer and B) Loan Officer, and also seen is how an explanation interface, C) needs to be tailored for different explanation types to show for these two user groups. Reproduced from: S. Chari, O. Seneviratne, M. Ghalwash, S. Shirai, D.M. Gruen, P. Meyer, P. Chakraborty and D.L. McGuinness, ``Explanation ontology: A general-purpose, semantic representation for supporting user-centered explanations,'' \textit{Semantic Web J.}, vol. pre-press, pp. 1 - 31, May 2023, doi: 10.3233/SW-233282, with permission from IOS Press. \copyright2023}
\label{fig:needforexplanations}  
\end{figure}

Additionally, explanations can be multi-dimensional in that, as humans, we reason through different paths before we trust a decision (Fig. \ref{fig:needforexplanations}). For example, when a clinician is deciding, ``which treatment option is suitable for a patient?'' they refer to multiple factors, including patient details such as their drug-related allergies, their current treatment, and literature to understand what is recommended for the patient. In summary, explanations are often reactive to user questions~\cite{dey2022human},~\cite{gunning2017explainable},~\cite{doshi2017accountability}, and often have multiple forms and types such as the ``What ifs'' or counterfactuals, ``What evidence'' or scientific and ``What data'' or data-based~\cite{gilpin2018explaining},~\cite{mittelstadt2019explaining},~\cite{wachter2017counterfactual}. However, given the increase in complexity of AI and ML methods, they have become more opaque, and a lot of explainability research has focused on model explanations~\cite{arya2022ai},~\cite{ribeiro2016should} alone. Several researchers~\cite{miller2019explanation},~\cite{mittelstadt2019explaining},~\cite{gilpin2018explaining} have posited the need for conversational user-centered explainability, which is of multiple types and supported by various sources such as data, knowledge, and context. However, many publications in user-centered explainability have either been position statements~\cite{dey2022human},~\cite{doshi2017accountability},~\cite{miller2019explanation} or survey papers~\cite{ribera2019can},~\cite{biran2017explanation},~\cite{tiddi2022knowledge, hasan2012explanation}, with fewer implementations that can be applied across use cases~\cite{lakkaraju2022rethinking}. Hence, through this thesis, we see an opportunity to provide flexible designs and implementations to support various types of user-centered explanations that build off of multiple AI results, including ML decisions and model explanations, data fragments, and domain and contextual knowledge, in various use cases. Specifically, we address the following research questions: 
\begin{itemize}
    \item \textit{How can we formally represent explanations with support for interacting AI systems, additional data sources, and along different dimensions?}
    \item \textit{How useful and feasible are such explanations for clinical settings?} 
    \item \textit{Is it feasible and generate combine explanations from multiple data modalities and AI methods?} 
\end{itemize} 

Furthermore, in this thesis, we strive to support explanations following the well-cited definition, we proposed upon a thorough review of directions and foundations of explainable knowledge-enabled systems. Our definition is:

Explanations are:
``an account of the system, its workings, the  \textit{implicit and explicit} knowledge used 
in its reasoning processes and the specific decision, 
that is \textit{sensitive} to the end-user's \textit{understanding, context, and current needs}''~\cite{chari2020foundationschapter},~\cite{chari2020directionschapter}. 

Semantic web technologies such as ontologies and knowledge graphs serve as tools to represent different explanation types from their contributing components to support such user-centered and knowledge-enabled explanations. Hence, as a first contribution, within this thesis, we design an Explanation Ontology~\cite{chari2020explanation}, a vocabulary representing different user-centered and literature-derived explanation types. Additionally, context is often mentioned as an important dimension to situate explanations in the application domain and use case~\cite{ghassemi2021false},~\cite{dey2022human, dey1998cyberdesk}, hence, as a second contribution, we design a method to extract and support contextual explanations in a clinical setting of real-world importance. Finally, since explanations that humans prefer are conversational~\cite{lakkaraju2022rethinking}, they often can be composed of several other explanations. As a third contribution, we are developing a metaexplainer capable of combining explanations from multiple ML and post-hoc explainers, data, and knowledge sources. Given the user-centered focus on explainability, we engage domain experts at multiple stages in the development of the contributions, primarily through requirements gathering, formulating the problem statements, and evaluating the methods' results. Other researchers~\cite{lakkaraju2022rethinking},~\cite{wang2019designing} also find that engaging the end-users helps build explainable AI systems that are more attuned to user needs. We provide a brief description of the contributions towards furthering user-centered explainability below: 

\section{Contributions}
    \subsection{Explanation Ontology} (Chapt. \ref{chapt:explanation_ontology}) We design an Explanation Ontology (EO), a general-purpose semantic representation that can represent fifteen different literature-derived explanation types via their system-, interface- and user- related components. We showcase the utility of the EO's model to represent explanation types across 7 different use cases ranging from domains of finance, food to healthcare. We design competency questions that our target end-user, a system developer, would ask when using the EO. Further, we have released two versions of the EO with added support for a wider range of commonly used explainer methods in EO V2.0. The EO is open-sourced and available at: \url{https://tetherless-world.github.io/explanation-ontology/index}. 
    
    \subsection{Contextualizing Model Explanations via a \\ Knowledge-augmented Question-Answering Method}  (Chapt. \ref{chapt:qa_contextualization}) We design a clinical question-answering (QA) system to address questions from clinical practice guidelines (CPGs) to provide contextual explanations to help clinicians interpret risk prediction scores and their post-hoc explanations in a comorbidity risk prediction setting. Here, we chose a setting of clinical and real-world importance, such as comorbidity risk prediction of a chronic disease, and refined the use case in consultation with a clinician. In addition, we identified dimensions of interest in the use case, along with which contextual explanations would be helpful for clinicians to interpret the scores and patient features better. From an implementation standpoint, we developed a QA framework to extract and support contextual explanations from CPGs. We leverage large language models and their clinical variants for the QA and build knowledge augmentations to improve the semantic coherence of the answers to the questions. We conduct a quantitative evaluation to demonstrate the QA's efficacy and the feasibility of supporting contextual explanations from authoritative literature. We also engaged clinicians in expert panel sessions to understand if contextual explanations are helpful and where else they would require support to use them in their practice. Additionally, while this contextualization method is developed for a clinical setting - our general approach from unstructured text extraction to QA can be applied to explain post-hoc explanations in other literature rich settings such as finance and policy.
    
    \subsection{MetaExplainer: A Method to Combine Multiple Explanations}  (Chapt. \ref{chapt:Metaexplainer}) We design a general-purpose metaexplainer framework capable of providing multiple explanations to an end-user question (e.g., that of a clinician). Within the MetaExplainer, we first want to break down a user question into sub-questions addressed by explanation types supported in the EO. We then can invoke explainers registered to these types to generate individual explanations. Overall, through the MetaExplainer, we generate natural language explanations from their individual data, knowledge, and method output components. We evaluate the explanations at each stage by using metrics~\cite{zhou2021evaluating} that the XAI community has proposed.
\section{Outline}
In the rest of thesis, we will first explore the related work to each contribution (Chapter \ref{chapt:related_work}), and in individual chapters (Chapters \ref{chapt:explanation_ontology}, \ref{chapt:qa_contextualization} and \ref{chapt:Metaexplainer}) discuss details of each contribution covering the motivation, methods and results of them. Finally, we discuss the challenges this thesis addresses and potential avenues for future work in the Discussion chapter (Chapt. \ref{chapt:Discussion}).


 
\chapter{RELATED WORK} \label{chapt:related_work}\blfootnote{Portions of this chapter previously appeared as: S. Chari, O. Seneviratne, M. Ghalwash, S. Shirai, D.M. Gruen, P. Meyer, P. Chakraborty and D.L. McGuinness, ``Explanation ontology: A general-purpose, semantic representation for supporting user-centered explanations,'' \textit{Semantic Web J.}, vol. Pre-press, pp. 1 - 31, May 2023, doi: 10.3233/SW-233282. Reprinted with permission from IOS Press. \copyright2023}
\blfootnote{Portions of this chapter previously appeared as: S. Chari, P. Acharya, D.M. Gruen, O. Zhang, E.K. Eyigoz, M. Ghalwash, O. Seneviratne, F.S. Saiz, P. Meyer, P. Chakraborty, D.L. McGuinnesss, ``Informing clinical assessment by contextualizing post-hoc explanations of risk prediction models in type-2 diabetes,'' \textit{Artificial Intelligence in Medicine J.}, vol. 137, Mar. 2023, Art. no. 102498, doi: 10.1016/j.artmed.2023.102498.}
This research combines different components - an ontology, a question-answering (QA) method for contextual explanations and a method to combine explanations, contributing to the overarching goal of user-centered explainability. Hence, below, I will review related work to the three contributions and present a critical analysis of how our approach differs or adds to these works. 

\section{Explanation Ontology}
Ontologies and Knowledge Graphs (KGs) capture an encoding of associations between entities and relationships in domains, and hence, can be used to inform upstream tasks~\cite{gaur2021semantics},~\cite{raghu2021learning}, guide/constrain ML models~\cite{beckh2021explainable}, and structure content for the purpose of organization~\cite{tiddi2015ontology},~\cite{mcguinness2004explaining}. User-centered explanations are composed of different components, such as outputs of Artificial Intelligence (AI) / Machine Learning (ML) methods and prior knowledge, and are also populated by content annotated by different domain ontologies and KGs (of which there are many). The former proposition of composing explanations from components has been attempted less frequently~\cite{tiddi2014dedalo},~\cite{tiddi2015ontology}, ~\cite{chari2020explanation},~\cite{dalvi2021explaining} and at different degrees of content abstraction, hence providing open challenges to represent explanations semantically. Additionally, there have been two multidisciplinary, comprehensive, and promising reviews~\cite{tiddi2022knowledge},~\cite{lecue2020role} highlighting the applications of KGs to explainable AI (XAI), either solely as the data store to populate explanations, or as aids to explain AI decisions from the knowledge captured by the KG encodings. Efforts to use ontologies and KGs to improve the explainability of AI models will become increasingly popular since several publications point out that single scores from ML models are hard for subject matter experts (SMEs) to interpret directly~\cite{ghassemi2021false},~\cite{park2020evaluating}, ~\cite{challener2019proliferation}. Here, we review semantic efforts to represent explanations and highlight how the EO is different in terms of its overall goal and representation.

One of the early efforts of an ontology to represent explanations was by Tiddi et al.~\cite{tiddi2015ontology}, who designed the Explanation Patterns (EP) ontology to model attributes of explanations from a philosophy perspective of explanation and its dependencies on what phenomenon generated the explanation and what events they are based on. They based their ontology design on ontological realism~\cite{smith2010ontological}, i.e., building their ontology to be as close to how it is defined in theory. The primary use of the EP ontology was to define explanations in multi-disciplinary domains, such as cognitive science, neuroscience, philosophy, and computer science, with a simple general model. The authors mention how the EP model can be applied in conjunction with a graph traversal algorithm like Dedalo~\cite{tiddi2014dedalo} to find three components of an explanation, including the \textbf{A}ntecedent event (A), the \textbf{T}heory they are based on (T), and the \textbf{C}ontext in which they are occurring (C), in KGs. However, we found that for the user-centered explanation types (Chapter \ref{chapt:explanation_ontology}), the (A, T, C) is often not sufficient when some explanations either do not need all the (A, T, C) components or when other explanations require more than these components. For example, (A, T, C) components are not sufficient in \emph{case-based explanations}, where what cases the explanation is dependent upon needs to be modeled. Additionally, we found that this simplicity was insufficient to support explanations generated by AI methods. Therefore, we added additional classes to support the dependencies of explanations on the methods that produce them and the users that consume them. However, in the spirit of reusing existing ontologies, we leverage certain classes and properties of the EP ontology in our EO ontology model as described in Chapter \ref{chapt:explanation_ontology}.

In a recent paper, Dalvi and Jansen et al.~\cite{dalvi2021explaining} released an Entailment Bank for explanations and applied natural-language processing methods to identify trees for the facts that are most relevant to a Question-Answer (QA) pair. While they are trying to identify entailments, or the facts that are most pertinent to an answer in a QA setting, several publications~\cite{lakkaraju2022rethinking},~\cite{dan2020designing},~\cite{dey2022xai} find that domain experts are often aware of the base knowledge that support explanations and do not always appreciate the additional theory. In the EO, we model a comprehensive set of literature-derived explanation types that can address various questions by system designers. These explanation types can provide users the varied support~\cite{lakkaraju2022rethinking},~\cite{dan2020designing} they seek in terms of information to help them better understand the AI methods output. For example, contextual explanations situate answers and scientific explanations provide evidence to reason about the supporting literature. As for composing the explanations, we describe how the description of sufficiency conditions on these explanation types allows them to be built from KGs (Chapter \ref{chapt:explanation_ontology}). 
In a similar vein, Teze et al.~\cite{tezeengineering} present style templates to combine assertional, terminological and ontology terms to support seven different user-centered explanation types including statistical, contextual, data-driven, simulation-based, justification-based, contrastive and counterfactual. However, their work doesn't use traditional ontology languages like RDF and OWL and is therefore less interoperable with standard semantic frameworks. Additionally, they focus more on integrating the outputs of logical reasoners to populate explanation types, instead of a wider breadth of AI explanation methods, such as those listed in Zhou et al.~\cite{zhou2021evaluating} and Arya et al.~\cite{arya2019one},~\cite{arya2022ai}. In the EO, we design a simple, yet comprehensive, model to represent user-centered explanation types that account for their generational needs and impact on the user.

\section{Contextualizing Model Explanations via a \\ Knowledge-augmented Question-answering (QA) \\ Method}
Our methods 
build on both expert feedback and past efforts to leverage clinical domain knowledge for generating explanations within AI assistants. 
Some 
notable and relevant past works include: MYCIN~\cite{shortliffe1974mycin}, where domain literature was encoded as rules and trace-based explanations, which addressed `Why,' `What,' and `How,' were provided for the treatment of infectious diseases; the DESIREE project~\cite{seroussi2018implementing}, where case, experience, and guideline-based knowledge was used to generate insights relevant to patient cases; and a mortality risk prediction effort~\cite{raghu2021learning} of cardiovascular patients, where a probabilistic model was utilized to combine insights from patient features, and
domain knowledge, to ascertain patient conformance to the literature. However, these approaches are either not flexible nor scalable for the ingestion of new domain knowledge~\cite{shortliffe1974mycin},~\cite{seroussi2018implementing}, or are narrowly focused in their approach to explanations along limited dimensions~\cite{raghu2021learning}. 
We attempt to allow clinicians to probe the supporting evidence systematically and thoroughly while asking holistic 
questions about the supporting evidence(s) 
to understand their patients better.


On the guideline QA front, there have been several efforts on representation formats
for guidelines and more recent work on applying ML and large language model (LLM)~\cite{devlin2018bert} approaches on guidelines for upstream tasks other than QA~\cite{hussain2021text},~\cite{hematialam2021identifying},~\cite{schlegel2019clinical}. Guideline representation efforts attempt to model guidelines as rules that can then be checked against patient data for conformance. While rule engineering is more accurate than applying ML models, it is not scalable without human effort. In a more scalable effort, Schlegel et al.~\cite{schlegel2019clinical}, have shown how a standard Natural Language Processing (NLP) pipeline of tools like named entity recognizers, syntactic, semantic, and dependency parsers can be applied to convert guideline text into annotated text snippets. However, their system, ClinicalTractor, is not available for reuse yet, and hence we could not use it for the semantic annotation portion of our QA pipeline. Another similar effort is from Hussain et al.~\cite{hussain2021text}, where they use heuristic patterns to identify different composition patterns in guideline sentences. These guideline natural language understanding efforts while useful, still require significant effort to be used upstream by QA approaches and could instead be used to augment QA approaches such as ours with additional information that can help improve semantic and syntactic coherence of answers.


On the other hand, with the rise of LLMs, several papers have been published on adapting LLMs to the biomedical and clinical domains by the pre-training of these models on large biomedical corpora~\cite{lee2020biobert},~\cite{gu2021domain}. There are currently very few efforts on the applications of these domain adapted LLMs to medical guidelines~\cite{hematialam2021identifying}. Hussein and Woldek~\cite{hematialam2021identifying} have applied LLMs to identify condition-action statements from three medical guidelines, and they find that the combination of syntactic and semantic features from Metamap can boost the performance of LLMs like BioBERT. However, it is not immediately clear how the extraction of condition-action statements can be used in a QA setting where the range of question types like those we support goes beyond condition-action pairs. For example, questions asking about diagnoses features don't always have a condition to be searched against.  Contrarily, Sarrouti et al.~\cite{sarrouti2020sembionlqa}, have designed a semantic biomedical QA system that achieves the state-of-the-art results on the BioASQ challenge~\cite{tsatsaronis2015overview} by using Unified Medical Language System (UMLS)~\cite{bodenreider2004unified} similarity scores and a novel passage retrieval algorithm to find candidate answers from Pubmed documents. In their future work, they list the needs for large training samples as a limiting factor to use a deep learning algorithm. While we agree, we have shown how knowledge augmentation algorithms can improve the performance of deep learning LLMs in new settings such as unseen guidelines. More recently, Nori et al.~\cite{nori2023capabilities} put forth a review paper demonstrating the performance of various LLMs on medical challenge datasets and presented how methods such as prompting and fine-tuning significantly improve the performance of the baseline models. However, guidelines are relatively smaller data settings and approaches like fine-tuning must be supplemented by methods such as weak supervision~\cite{ratner2017snorkel} to produce sufficiently large training data. 

Several studies have also tried to utilize patient data to query the literature for applicable evidence or treatment suggestions. However, very few of these studies combine multiple modalities and sources of data and knowledge for querying. 
Agosti et al.~\cite{agosti2019analysis}, conducted an analysis of query reformulation techniques for precision medicine.
Natarajan et al.~\cite{natarajan2010analysis}, conducted an analysis of clinical queries in an electronic health record search utility and found that queries on diseases and lab results were most searched.
 Patel et al.~\cite{patel2007matching}, matched patient records to clinical trials using ontologies and used a purely logical A-box and T-box approach to query literature.


Finally, several studies have hinted that model explanations alone are not sufficient and indicate that context can be an important dimension to make these explanations more useful. We summarize a few studies which either contextualize model explanations with links to knowledge or can provide context around risk prediction scores to make them more useful. 
 Rieger et al.~\cite{rieger2020interpretations}, found that interpretations are useful by penalizing explanations to align neural networks with prior knowledge.
 Zhang et al.~\cite{zhang2021context}, presented context-aware and time-aware attention-based model for disease risk prediction with interpretability. They used disease code hierarchies as context in RNN network's attention layer. Weber et al.~\cite{weber2021knowledge}, attempted a knowledge-based XAI through case-based reasoning and found that there is more to explanations than models can tell. Yao et al.~\cite{yao2021refining}, refined LLMs with Compositional Explanations by aligning LLM and post-hoc output with human knowledge and Tonekaboni et al.~\cite{tonekaboni2019clinicians}, analyzed what clinicians want and found that contextualizing explainable ML for clinical end use is helpful.
 
\section{MetaExplainer: A Method to Combine Multiple Explanations}
While several model explainer~\cite{arya2022ai},~\cite{ribeiro2016should},~\cite{lundberg2017unified} methods exist to provide post-hoc explanations of model decisions, or even counterfactuals and contrastive explanations, no single method can provide the composite view that users require~\cite{lakkaraju2022rethinking}. Additionally, recent papers point out that current model explainability~\cite{liao2022connecting},~\cite{ghassemi2021false} does not align with the expected human-comprehensible explanations that domain experts expect when using AI aids in their practice:  explanations required in real-world applications are multi-modal and conversational in nature~\cite{doshi2017accountability},~\cite{lakkaraju2022rethinking}. 

In an effort to produce explanations directly for user questions, Slack et al.~\cite{slack2023explaining} propose using utterances or cues to prompt LLMs to create filters that can be used to serve as inputs for model explainers and thus generate model explanations in-line with user questions. While this approach is promising, it mainly works for model explainers and not for other methods that borrow from domain knowledge. In our metaexplainer approach, we plan to develop methods that can provide users with multiple user-centered explanations supported within the EO and also produce these explanations from different modalities, including data, method outputs, and knowledge. Our closest related work is by Slack et al~\cite{slack2023explaining} who propose a TalkToModel or Megaexplainer framework where users can ask questions and are provided natural-language explanations of model explanations including contrastives or feature-importance and data explanations or details about data processing and analysis. However, there framework only supports a limited set of explanation types and is limited by the types of questions that can be addressed owing to the rule-based generation process of questions. Our MetaExplainer is more modular and is adaptable to more explanation methods (i.e., developers can add support for more explainer methods under the Delegate stage of the MetaExplainer) and we currently support five user-centered explanation types (Case based, Contrastive, Counterfactual, Data and Rationale)  which is more than the four types of explanation types (Contrastive, Counterfactual, Data and System Performance) that Megaexplainer supports. We reused the design principles of Megaexplainer  where possible.

 Additionally, some researchers such as Krishna et al.~\cite{krishna2023towards} are also looking at when it is wise to produce explanations and when not, they propose a Robust Counterfactual Explanations under the Right to be Forgotten (ROCERF) framework. Further, Ghassemi et al.~\cite{ghassemi2021false} also allude to the false hope of explanations in their article. While ROCERF is a start at analyzing whether explanations provide value, it is hard to apply the method to problems beyond the field of counterfactual explanations studied. In the MetaExplainer, we rank explanation outputs using well-studied explanation metrics~\cite{zhou2021evaluating}. Finally, McGuinness and Silva~\cite{mcguinness2007explaining},~\cite{mcguinness2004explaining} have early work on composing explanations from components in task-based environments. However, their work predates the explainer methods that we use today but is still relevant in the various modules such as dispatchers, constraint, and knowledge explainers contributing to explanations. We leverage their modular design while designing the MetaExplainer so that the different explanation composition stages are modular and easy to swap.
 
\chapter{EXPLANATION ONTOLOGY} \label{chapt:explanation_ontology} \blfootnote{This chapter previously appeared as: S. Chari, O. Seneviratne, M. Ghalwash, S. Shirai, D.M. Gruen, P. Meyer, P. Chakraborty and D.L. McGuinness, ``Explanation ontology: A general-purpose, semantic representation for supporting user-centered explanations,'' \textit{Semantic Web J.}, vol. Pre-press, pp. 1 - 31, May 2023, doi: 10.3233/SW-233282. Reprinted with permission from IOS Press. \copyright2023}
\blfootnote{Portions of this chapter previously appeared as: S. Chari, O. Seneviratne, D.M. Gruen, M.A. Foreman, A.K. Das and D.L. McGuinness, ``Explanation ontology: A model of explanations for user-centered AI,'' in: Intl. Semantic Web Conf., Springer, 2020, pp. 228–243.}
\section{Overview}
In recent years, with principles introduced in global policy around AI such as in Europe's General Data Protection Regulation (GDPR) act~\cite{gdpr} and the White House's National AI Research Resource (NAIRR) Task Force~\cite{whitehouseaitaskforce}, there has been a growing focus on trustworthy AI. This focus has reflected the need for transparency and trust around the vast amounts of data collected by parties in multiple domains and has brought to light potential concerns with the AI models increasingly used on such data to affect important decisions and policy. In the trustworthy AI age, several position statements ~\cite{mittelstadt2019explaining}, ~\cite{miller2019explanation}, ~\cite{gilpin2018explaining}, ~\cite{doshi2017accountability} focused on directions to move towards explainable, fair, and causal AI. These papers inspired computational efforts to improve trust in AI models. For example, to enable explainability in composition to provide confidence in model usage, IBM released the AIX-360 toolkit~\cite{arya2019one}, ~\cite{arya2022ai}, with multiple explainer methods capable of providing different types of model explanations, including local, global, and post-hoc explanations. At the same time, there have been user studies~\cite{gilpin2018explaining}, ~\cite{wang2019designing}, ~\cite{liao2020questioning}, ~\cite{chari2020explanation} on the explanation types that users require and the questions that they want to be addressed, illustrating that user-centered explainability is question-driven, conversational and contextual. 

Inspired by these studies, we reviewed the social sciences and computer science literature~\cite{chari2020directionschapter} regarding explanations and cataloged several user-centered explanation types that address different types of user questions and the associated informational needs. We conducted expert panel studies with clinicians ~\cite{dan2020designing} to understand which of these explanation types were most helpful in guideline-based care.  We found that clinicians prefer a combination of holistic, scientific, and question-driven explanations connected to broader medical knowledge and their implications in context, beyond typical model explanations which focus only on the specific data and AI mechanisms used. In line with our findings,  Dey et al.~\cite{dey2022human} recently illustrated a spectrum of clinical personas and their diverse needs for AI explainability. Also, recent papers point out that current model explainability~\cite{GHASSEMI2021e745} does not align with the human-comprehensible explanations that different domain experts expect when using AI aids in their practice.  This reinforces the idea that explanations in real-world applications need to involve domain knowledge and be sensitive to the  prior knowledge, usage situations, and resulting informational needs of the users to whom they are delivered. 

These studies point to still unmet needs in bridging the gap between explanations that AI models can generate and what users want. 
This gap motivated us to design the EO to represent user-centered explanation types and their dependencies.
We aimed to provide system designers with a single resource that could be used to identify critical user-centered explanation requirements, and further to provide building blocks for them to use in their designs. 
In our modeling of the EO (Sec. \ref{sec:eomodel}), we take into account our learnings from our literature review for various explanation types and include terminology that is typically associated with explanations both from the perspective of end-users (e.g., from the clinicians we interviewed) that consume them, and the AI methods that generate them.

\begin{figure}[hbt!]
\centering
\includegraphics[width=1.0\linewidth]{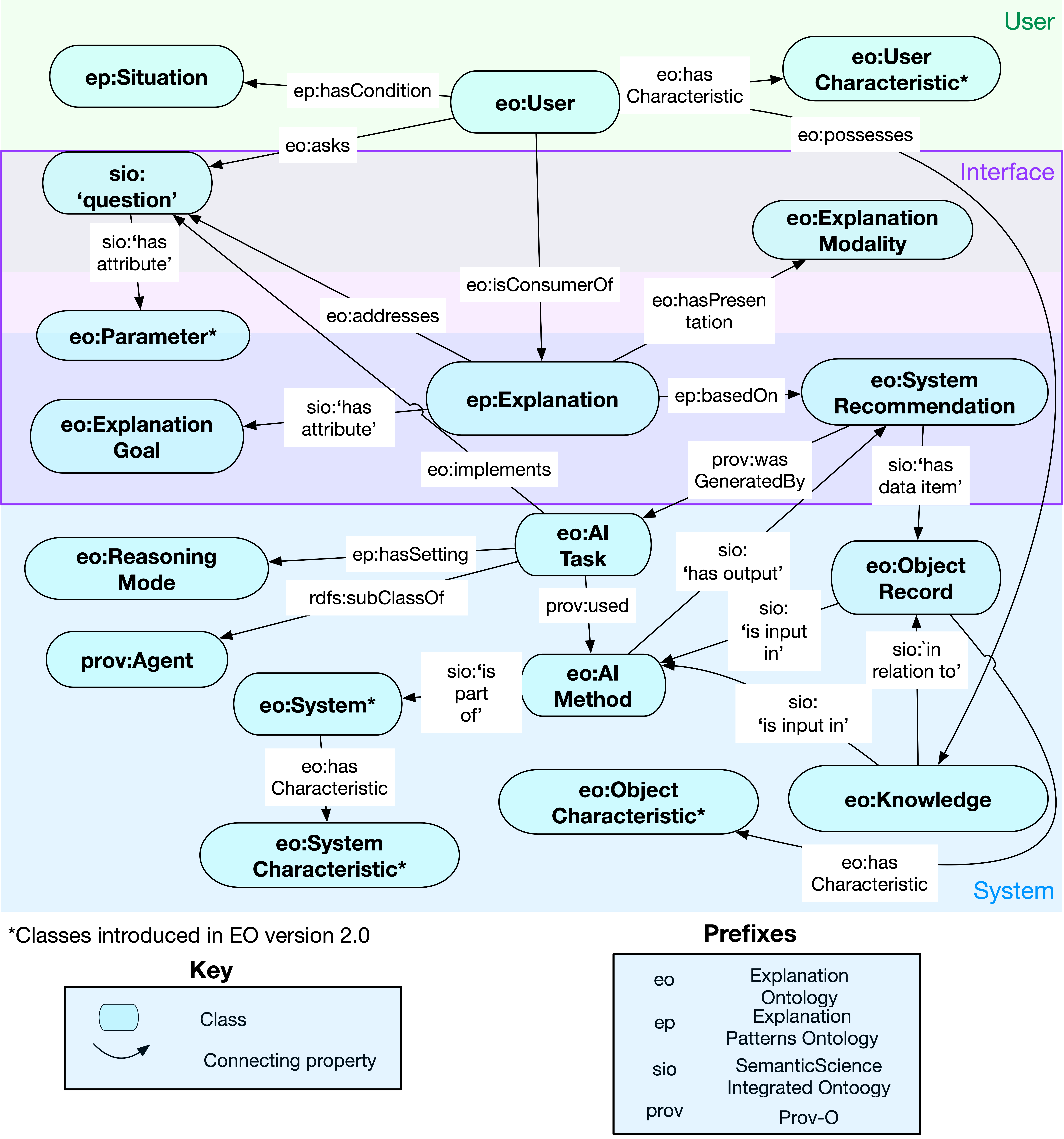}
\caption{Core model of the Explanation Ontology that ties together system-, interface- and user- dependencies of explanations. Reproduced from S. Chari, O. Seneviratne, M. Ghalwash, S. Shirai, D.M. Gruen, P. Meyer, P. Chakraborty and D.L. McGuinness, ``Explanation ontology: A general-purpose, semantic representation for supporting user-centered explanations,'' \textit{Semantic Web J.}, vol. Pre-press, pp. 1 - 31, May 2023, doi: 10.3233/SW-233282, with permission from IOS Press. \copyright 2023}
\label{fig:eomodel}  
\end{figure}
\section{Methods} \label{sec:eomodel}
\subsection{Explanation Ontology Model}
In the EO, we capture the attributes that explanations build upon, such as a `system recommendation' and `knowledge,' and model their interactions with the `question' (s) they address. We can broadly categorize explanation dependencies into the system, user, and interface attributes as seen in Fig. \ref{fig:eomodel}. These categorizations help us cover a broad space of attributes that explanations directly or indirectly interact with. 

\textbf{User Attributes}: User attributes are the concepts that are related to a `user' who is consuming an explanation. These include the `question' that the user is asking, `user characteristic's that describe the user, and the user's `situation'. 
Explanations that the user is consuming are modeled to address the user's `question,' and may also take into account factors such as the user's `situation' or their `user characteristic' such as their education and location.

\textbf{System Attributes}: System attributes encapsulate the concepts surrounding the AI `system' used to generate recommendations and produce explanations. `Explanation's are based on `system recommendation's, which in turn are generated by some `AI task .'`AI task's are analogous to the high-level operations that an AI system may perform (e.g., running a reasoning task to generate inferences or running an explanation task to generate explanations about a result). The `AI task' relates to user attributes as it addresses the `question' that a user is asking. Another important class we capture as part of system attributes, is that of the `AI method', that `AI tasks' use, to produce the `system recommendation' that an `explanation' is dependent on. Further, from a data modeling perspective, the `object record' class helps us include contributing objects in the `system recommendation' that could further have `knowledge' and `object characteristic' of their own, contributing to explanations. For example, in a patient's risk prediction (an instance of `system recommendation'), the patient is an object record linked to the prediction, and an explanation about the patient's risk refers to both the prediction and the patient. 
Also, within the system attributes, various additional concepts such as `reasoning mode' and `system characteristic' are modeled to capture further details about the AI system's operations and how they relate to each other. Maintaining such system provenance helps system designers debug explanations and their dependencies.

\textbf{Interface Attributes}: Lastly, interface attributes capture an intersection between user and system attributes that can be directly interacted with on a user interface (UI). From an input perspective, these attributes consist of the `question' that the user asks, the `explanation goal' they want to be fulfilled by the explanation, and the `explanation modality' they prefer. From the content display perspective, we capture the system attributes that might be displayed on the UI, such as the `explanation' itself and the `system recommendation' it is based on.

\begin{figure}[hbt!]
\centering
\includegraphics[width=0.9\linewidth]{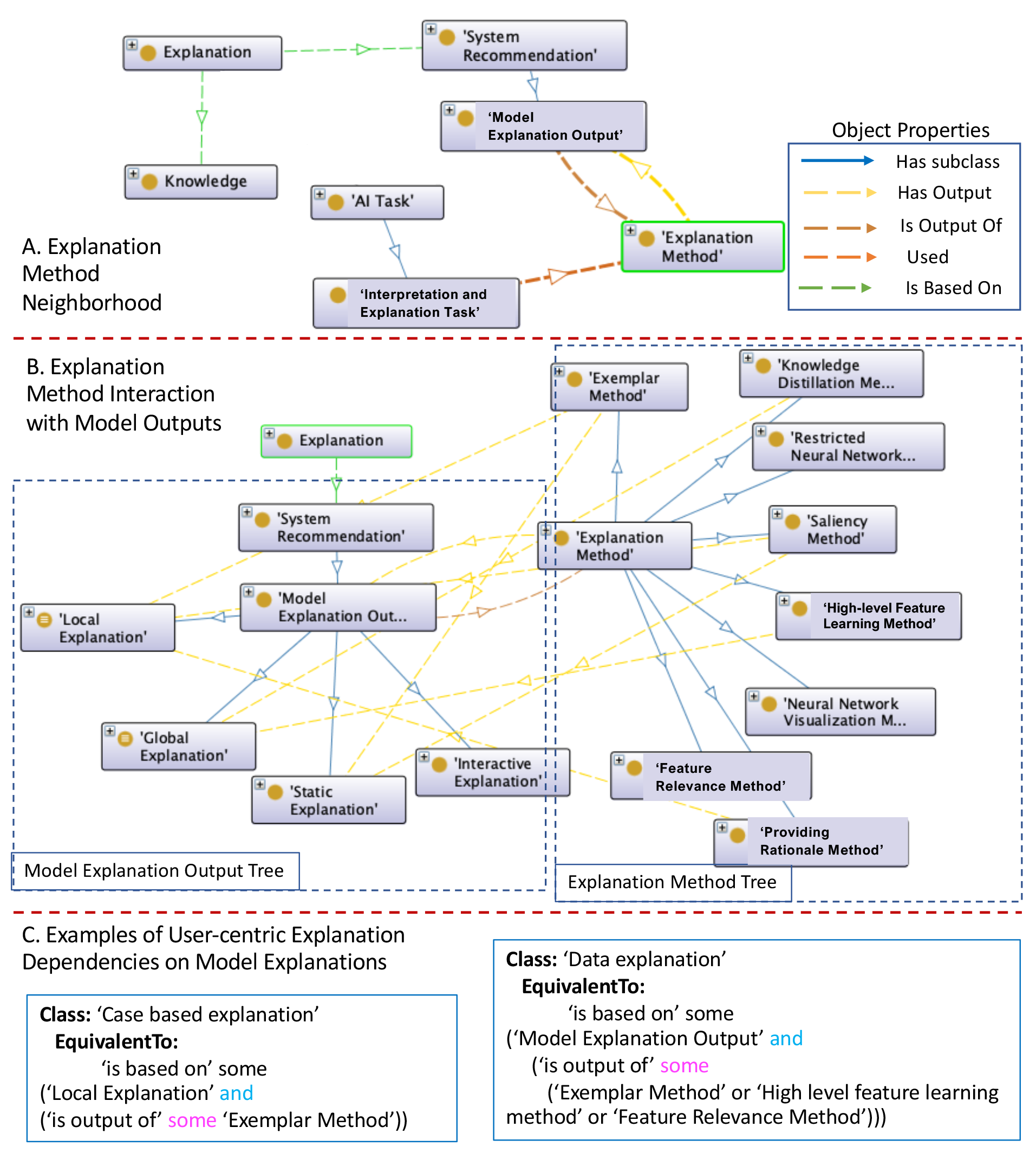}
\caption{Model explanations and their linked methods supported within the EO. Shown here in A) is the dependency of explanation methods (model explainers) on other classes in the EO model, B) the variety of model explainers and the explanation outputs they generate (e.g., saliency methods provide local explanations) and finally C) the dependence of a user-centered explanation, contrastive explanation, on the model explanation outputs. Reproduced from S. Chari, O. Seneviratne, M. Ghalwash, S. Shirai, D.M. Gruen, P. Meyer, P. Chakraborty and D.L. McGuinness, ``Explanation ontology: A general-purpose, semantic representation for supporting user-centered explanations,'' \textit{Semantic Web J.}, vol. Pre-press, pp. 1 - 31, May 2023, doi: 10.3233/SW-233282, with permission from IOS Press. \copyright 2023}
\label{fig:eoexplanations}  
\end{figure}

This mid-level model of the explanation space can be further extended by adding sub-class nodes to introduce more specific extensions for the entities where they exist. For example, both the `Knowledge' and `AI Method' classes have several sub-classes to capture the different types of `knowledge' and `AI methods' that explanations can be dependent on and are generated by, respectively. An example of the `AI Method' hierarchy and its interactions to support the generation of system recommendations upon which explanations are based can be seen in Fig. \ref{fig:eoexplanations}.

\begin{table}[hbt!]
\centering
\caption{Statistics on the composition of the Explanation Ontology. These counts are taken by loading the EO into the ontology editing tool, Protege{~\cite{protegeweb}} and choosing the active closure of the ontology with imported ontologies view option{~\cite{protegemetrics}}. We calculated the counts of classes and properties we introduced by commenting out the import statements in the ontology file. The introduced classes and properties are indicated by a $^*$ in the table.}
\label{tab:ontology-metrics}
\begin{tabular}{|l|l|}
\hline
\textbf{Metrics}        & \textbf{Count} \\ \hline
Classes                 & 1707           \\ \hline
Classes Introduced$^{*}$                 & 135          \\ \hline
Object Properties       & 283            \\ \hline
Object Properties Introduced$^{*}$     & 52            \\ \hline
Data Properties         & 8              \\ \hline
Data Properties Introduced$^{*}$       & 2              \\ \hline
Equivalent Class Axioms & 85             \\ \hline
Equivalent Class Axioms Introduced$^{*}$ & 35             \\ \hline
Instances               & 17             \\ \hline
Instances Introduced$^{*}$            & 13             \\ \hline
\end{tabular}
\end{table}

\subsection{Supported Explanation Types}
We refer to explanation types  in line with the different kinds of explanations that have been referred to in the explanation sciences literature, arising in such fields as ``law, cognitive science, philosophy, and the social sciences"~\cite{mittelstadt2019explaining}.  These include such things as contrastive, scientific, and counterfactual explanations. We found that these explanations are well defined in adjacent fields of the explanation sciences and more rarely in computer science.  We performed a literature review looking for explanations that serve different purposes, address different questions, and are populated by different components (such as `cases' for case-based explanations and evidence for `scientific explanations') in the hope of refining their definitions to make these explanation types easier to generate by computational means. We previously released these explanation types and their definitions as a taxonomy~\cite{chari2020directionschapter}, and we now encode these explanation types, definitions, and sufficiency conditions in the EO. The explanation types we support in the EO, their definitions, and sufficiency conditions can be browsed in Tab. \ref{tab:explanationtypes}, Tab. \ref{tab:explanationtypes1}, \ref{tab:explanationtypes2}, Tab. \ref{tab:newexplanationtypes}, Tab. \ref{tab:newexplanationtypes1} and Tab. \ref{tab:newexplanationtypes2}. 

In the EO, against each explanation type, we encode the sufficiency conditions as equivalent class axioms, allowing instances that fit these patterns to be automatically inferred as instances of the explanation types. 
We find that defining the equivalent class restrictions using top-level classes allows the subsumption of sub-classes of these top-level classes into the explanation type restrictions as well. Patterns of such subsumptions can be seen in our use case descriptions (Sec. \ref{sec:usecases}). Further, if system designers want to familiarize themselves with the components that are necessary for each explanation type, we suggest that they browse the sufficiency conditions, unless they have familiarity with the OWL ontology language~\cite{baader2004description}. The equivalent class restrictions are a logical translation of the sufficiency conditions, so the modeling of explanations based on an understanding of the sufficiency conditions should prove sufficient. Against some explanation types, we also maintain what `AI methods' can generate them or what types of `knowledge' they are dependent upon, so if a system designer were looking to support certain explanation types in their systems, they can plan ahead to include these methods and/or knowledge types.

Also, in addition to the $nine$ different explanation types that we previously supported in the EO~\cite{chari2020explanation} (Tab. \ref{tab:explanationtypes} and Tab. \ref{tab:explanationtypes1}), we have added $six$ new explanation types mentioned in Zhou et al.'s recent paper~\cite{zhou2021evaluating} (Tab. \ref{tab:newexplanationtypes} and Tab. \ref{tab:newexplanationtypes1}). The addition of these explanation types also prompted the support for explanation methods as subclasses of `AI Method.' Over the past decade, with the focus on trustworthy AI, there have been several developments in explanation methods or model explainers~\cite{van2018contrastive}, ~\cite{wachter2017counterfactual}, ~\cite{lundberg2017unified}, ~\cite{bau2017network}, ~\cite{ribeiro2016should}. 
Part of our goal for the expansion of the EO was to be of use in explainability toolkits, so we reviewed a comprehensive set of explanation methods that are a part of the AI Explainability 360 toolkit~\cite{aix360}. We encode the outputs of these explanation methods, i.e., model explanations and their subtypes (e.g., local, global, static, and interactive explanations) as subclasses of `system recommendation.' The modeling of model explanations and explanation methods can be seen in Fig. \ref{fig:eoexplanations}. We also capture the dependencies on user-centered explanations we define in Tab. \ref{tab:explanationtypes} on these `model explanations.' Further, we observe that the user-centered explanation types can depend on more than one model explanation and users need more context and knowledge to consume these model explanations.

\begin{table}[hbt!]
\centering
\caption{An overview of $2/9$ previously supported explanation types, their simplified descriptions, example questions they can address (in bold, within the Description column), and their sufficiency conditions expressed in natural language.}
\label{tab:explanationtypes}
\begin{adjustbox}{max width=1.0\textwidth,center}
\begin{tabular}{|l|l|l|}
\hline
\multicolumn{1}{|m{0.15\columnwidth}|}{\textbf{Explanation Type}} &
  \multicolumn{1}{m{0.4\columnwidth}|}{\textbf{Description}} &
  \multicolumn{1}{m{0.4\columnwidth}|}{\textbf{Sufficiency Conditions}} \\ \hline

\multicolumn{1}{|m{0.15\columnwidth}|}{\textbf{Case Based}} &
   \multicolumn{1}{m{0.4\columnwidth}|}{Provides solutions that are based on actual prior cases that can be presented to the user to provide compelling support for the system’s conclusions, and may involve analogical reasoning, relying on similarities between features of the case and of the current situation.
  \textbf{``To what other situations has this recommendation been applied?''}}  &
   \multicolumn{1}{m{0.4\columnwidth}|}{Is there at least one other prior case (\textit{`object record'}) similar to this situation that had an \textit{`explanation'}? Is there a similarity between this case, and that other case? } \\
  \hline

\multicolumn{1}{|m{0.15\columnwidth}|}{\textbf{Contextual}} &
  \multicolumn{1}{m{0.4\columnwidth}|}{Refers to information about items other than the explicit inputs and output, such as information about the user, situation, and broader environment that affected the computation. 
  
  \textbf{``What broader information about the current} \textbf{situation prompted the suggestion of this}  \textbf{recommendation?''}} &
  \multicolumn{1}{m{0.4\columnwidth}|}{Are there any other extra inputs that are not contained in the \textit{`situation'} description itself?And by including those, can better insights be included in the \textit{`explanation'}?} \\ \hline

\end{tabular}
\end{adjustbox}
\end{table}

\begin{table}[hbt!]
\centering
\caption{An overview of $2/9$ previously supported explanation types, their simplified descriptions, example questions they can address (in bold, within the Description column), and their sufficiency conditions expressed in natural language.}
\label{tab:explanationtypes1}
\begin{adjustbox}{max width=1.0\textwidth,center}
\begin{tabular}{|l|l|l|}
\hline
\multicolumn{1}{|m{0.2\columnwidth}|}{\textbf{Explanation Type}} &
  \multicolumn{1}{m{0.4\columnwidth}|}{\textbf{Description}} &
  \multicolumn{1}{m{0.4\columnwidth}|}{\textbf{Sufficiency Conditions}} \\ \hline
  
  \multicolumn{1}{|m{0.15\columnwidth}|}{\textbf{Contrastive}} &
  \multicolumn{1}{m{0.4\columnwidth}|}{Answers the question “Why this output instead of that output,” making a contrast between the given output and the facts that led to it (inputs and other considerations),  and an alternate output of interest and the foil (facts that would have led to it). \textbf{``Why choose option A}  \textbf{over option B that I typically choose?''}} &
  \multicolumn{1}{m{0.4\columnwidth}|}{Is there a \textit{`system recommendation'} that was made (let’s call it A)? What facts led to it? Is there another \textit{`system recommendation'} that  could have happened or did occur, (let’s call it B)?  What was the \textit{`foil'} that led to B? Can A and B be compared?} \\ \hline
  
\multicolumn{1}{|m{0.2\columnwidth}|}{\textbf{Counterfactual}} &
   \multicolumn{1}{m{0.4\columnwidth}|}{Addresses the question of what solutions would have been obtained with a different set of inputs than those used. 
   
  \textbf{``What if input A was over 1000?”}} &
  \multicolumn{1}{m{0.4\columnwidth}|}{Is there a different set of inputs that can be considered? If so what is the alternate \textit{`system recommendation'}?} \\ \hline
  
\multicolumn{1}{|m{0.2\columnwidth}|}{\textbf{Everyday}} &
  \multicolumn{1}{m{0.4\columnwidth}|}{Uses accounts of the real world that appeal to the user, given their general understanding and worldly and expert knowledge.  
  
  \textbf{``Why does option A make sense”}} &
  \multicolumn{1}{m{0.4\columnwidth}|}{Can accounts of the real world be simplified to appeal to the user based on their general understanding and \textit{`knowledge'}?} \\ \hline
\end{tabular}%
\end{adjustbox}
\end{table}

\begin{table}[hbt!]
\centering
\caption{An overview of $2/9$ previously supported explanation types, their simplified descriptions, example questions they can address (in bold, within the Description column), and their sufficiency conditions expressed in natural language.}
\label{tab:explanationtypes2}
\begin{adjustbox}{max width=1.0\textwidth,center}
\begin{tabular}{|l|l|l|}
\hline
\multicolumn{1}{|m{0.2\columnwidth}|}{\textbf{Explanation Type}} &
  \multicolumn{1}{m{0.4\columnwidth}|}{\textbf{Description}} &
  \multicolumn{1}{m{0.4\columnwidth}|}{\textbf{Sufficiency Conditions}} \\ \hline
  
  \multicolumn{1}{|m{0.15\columnwidth}}{\textbf{Scientific}} &
  \multicolumn{1}{|m{0.4\columnwidth}|}{References the results of rigorous scientific methods, observations, and measurements.
  
  \textbf{``What studies have backed this recommendation?''} } &
   \multicolumn{1}{m{0.4\columnwidth}|}{Are there results of rigorous \textit{`scientific} \textit{methods'} to explain the situation? Is there \textit{`evidence'} from the literature to explain this \textit{`system recommendation'},  \textit{`situation'} or \textit{`object record'}} \\ \hline

\multicolumn{1}{|m{0.15\columnwidth}}{\textbf{Simulation Based}} &
  \multicolumn{1}{|m{0.4\columnwidth}|}{Uses an imagined or implemented imitation of a system or process and the results that emerge from similar inputs. 
  
 \textbf{``What would happen if this recommendation is followed?''} } &
  \multicolumn{1}{m{0.4\columnwidth}|}{Is there an \textit{`implemented'}  imitation of the \textit{`situation'} at hand? Does that other scenario have inputs similar to the current \textit{`situation'}?} \\ \hline
 
\end{tabular}
\end{adjustbox}
\end{table} 

\begin{table}[hbt!]
\centering
\caption{An overview of $2/9$ previously supported explanation types, their simplified descriptions, example questions they can address (in bold, within the Description column), and their sufficiency conditions expressed in natural language.}
\label{tab:explanationtypes3}
\begin{adjustbox}{max width=1.0\textwidth,center}
\begin{tabular}{|l|l|l|}
\hline
\multicolumn{1}{|m{0.2\columnwidth}|}{\textbf{Explanation Type}} &
  \multicolumn{1}{m{0.4\columnwidth}|}{\textbf{Description}} &
  \multicolumn{1}{m{0.4\columnwidth}|}{\textbf{Sufficiency Conditions}} \\ \hline
  \multicolumn{1}{|m{0.15\columnwidth}}{\textbf{Statistical}} &
  \multicolumn{1}{|m{0.4\columnwidth}|}{Presents an account of the outcome based on data about the occurrence of events under specified  (e.g., experimental) conditions. Statistical explanations refer to numerical evidence on the likelihood of factors or processes influencing the result. 
  
  \textbf{``What percentage of people with } 
  
  \textbf{this condition have recovered?''} } &
  \multicolumn{1}{m{0.4\columnwidth}|}{Is there \textit{`numerical evidence'}/likelihood account of the \textit{`system recommendation'} based on data about the occurrence of the outcome described in the recommendation?} \\ \hline

    \multicolumn{1}{|m{0.15\columnwidth}}{\textbf{Trace Based}} &
   \multicolumn{1}{|m{0.4\columnwidth}|}{Provides the underlying sequence of steps used by the system to arrive at a specific result, containing the line of reasoning per case and addressing the question of why and how the application did something. 
  
 \textbf{``What steps were taken by the system to generate this recommendation?''}}&
  \multicolumn{1}{m{0.4\columnwidth}|}{Is there a record of the underlying sequence of steps (\textit{`system trace'}) used by the \textit{`system'} to arrive at a specific \textit{`recommendation'}} \\ \hline

 

\end{tabular}%
\end{adjustbox}
\end{table}

\begin{table}[hbt!]
\centering
\caption{An overview of $2/6$ new explanation types described in Zhou et al.~\cite{zhou2021evaluating} that we encode in the Explanation Ontology version 2.0. }
\label{tab:newexplanationtypes}
\begin{adjustbox}{max width=1.0\textwidth,center}
\begin{tabular}{|l|l|l|}
\hline
\multicolumn{1}{|m{0.15\columnwidth}|}{\textbf{Explanation Type}} &
  \multicolumn{1}{m{0.4\columnwidth}|}{\textbf{Description}} &
  \multicolumn{1}{m{0.4\columnwidth}|}{\textbf{Sufficiency Conditions}} \\ \hline
\multicolumn{1}{|m{0.15\columnwidth}}{\textbf{Data}} &
  \multicolumn{1}{|m{0.4\columnwidth}|}{Focuses on what the data is and how it has been used in a particular decision, as well as what data and how it has been used to train and test the ML model. This type of explanation can help users understand the influence of data on decisions. 
  
\textbf{"What the data is?", "How it has been used in a particular decision?", "How has the data been used to train the ML model?"}} &
 \multicolumn{1}{m{0.4\columnwidth}|}{Is there a `system recommendation' from an `AI method' that has as input, a `dataset' or part of it?
 
 Is there a `system recommendation' that includes `object records' that are used to train / test the `AI method'?} \\ \hline
 
\multicolumn{1}{|m{0.15\columnwidth}}{\textbf{Rationale}} &
  \multicolumn{1}{|m{0.4\columnwidth}|}{About the “why” of an ML decision and provides reasons that led to a decision, and is delivered in an accessible and understandable way, especially for lay users. If the ML decision was not what users expected, rationale explanations allows users to assess whether they believe the reasoning of the decision is flawed. While, if so, the explanation supports them to formulate reasonable arguments for why they think this is the case.
  
\textbf{"Why was this ML decision made and provide reasons that led to a decision?"}} &
 \multicolumn{1}{m{0.4\columnwidth}|}{Is there a `system recommendation' from an `AI method' that has a 
 `system trace'?
 
 Is there a `local explanation' output that an `explanation' is based on?} \\ \hline

 \end{tabular}%
 \end{adjustbox}
\end{table}

\begin{table}[hbt!]
\centering
\caption{An overview of $2/6$ new explanation types described in Zhou et al.~\cite{zhou2021evaluating} that we encode in the Explanation Ontology version 2.0. }
\label{tab:newexplanationtypes1}
\begin{adjustbox}{max width=1.0\textwidth,center}
\begin{tabular}{|l|l|l|}
\hline
\multicolumn{1}{|m{0.15\columnwidth}|}{\textbf{Explanation Type}} &
  \multicolumn{1}{m{0.4\columnwidth}|}{\textbf{Description}} &
  \multicolumn{1}{m{0.4\columnwidth}|}{\textbf{Sufficiency Conditions}} \\ \hline

  \multicolumn{1}{|m{0.15\columnwidth}}{\textbf{Safety and Performance}} &
  \multicolumn{1}{|m{0.4\columnwidth}|}{Deals with steps taken across the design and implementation of an ML system to maximise the accuracy, reliability, security, and robustness of its decisions and behaviours. Safety and performance explanations help to assure individuals that an ML system is safe and reliable by explanation to test and monitor the accuracy, reliability, security, and robustness of the ML model.
  
\textbf{``What steps were taken to ensure robustness and reliability of system?,'' ``How has the data been used to train the ML model?''}} &
 \multicolumn{1}{m{0.4\columnwidth}|}{Is there a `system recommendation' from an `AI method' that is part of a `system' that exposes its design `plans'?
 
 Is there a `system recommendation' that includes `object records' that are used to train / test the `AI method'?} \\ \hline
 
 \multicolumn{1}{|m{0.15\columnwidth}}{\textbf{Impact}} &
  \multicolumn{1}{|m{0.4\columnwidth}|}{Concerns the impact that the use of a system and its decisions has or may have on an individual and on a wider society. Impact explanations give individuals some power and control over their involvement in ML-assisted decisions. 
  
\textbf{"What is the impact of a system recommendation?", "How will the recommendation affect me?"}} &
 \multicolumn{1}{m{0.4\columnwidth}|}{Is there a `system recommendation' from an `AI method' that has a `statement of consequence'?} \\ \hline

\end{tabular}%
\end{adjustbox}
\end{table}

\begin{table}[hbt!]
\centering
\caption{An overview of $2/6$ new explanation types described in Zhou et al.~\cite{zhou2021evaluating} that we encode in the Explanation Ontology version 2.0.}
\label{tab:newexplanationtypes2}
\begin{adjustbox}{max width=1.0\textwidth,center}
\begin{tabular}{|l|l|l|}
\hline
\multicolumn{1}{|m{0.2\columnwidth}|}{\textbf{Explanation Type}} &
  \multicolumn{1}{m{0.4\columnwidth}|}{\textbf{Description}} &
  \multicolumn{1}{m{0.4\columnwidth}|}{\textbf{Sufficiency Conditions}} \\ \hline
  
 \multicolumn{1}{|m{0.2\columnwidth}}{\textbf{Fairness}} &
  \multicolumn{1}{|m{0.4\columnwidth}|}{Provides steps taken across the design and implementation of an ML system to ensure that the decisions it assists are generally unbiased, and whether or not an individual has been treated equitably. Fairness explanations are key to increasing individuals’ confidence in an AI system. It can foster meaningful trust by explaining to an individual how bias and discrimination in decisions are avoided.
  
\textbf{"Is there a bias consequence of this system recommendation?", "What data was used to arrive at this decision?"}} &
 \multicolumn{1}{m{0.4\columnwidth}|}{Is there a `system recommendation' from an `AI method' that has a `statement of consequence'?
 
 Is there a `dataset' in the `system recommendation' the explanation is based on?} \\ \hline
  \multicolumn{1}{|m{0.2\columnwidth}}{\textbf{Responsibility}} &
  \multicolumn{1}{|m{0.4\columnwidth}|}{Concerns ``who'' is involved in the development, management, and implementation of an ML system, and ``who'' to contact for a human review of a decision. Responsibility explanations help by directing the individual to the person or team responsible for a decision. It also makes accountability traceable.
  
\textbf{``Who is involved in the development, management, and implementation of an ML system?,'' ``Who to contact for a human review of a decision?''}} &
 \multicolumn{1}{m{0.4\columnwidth}|}{Is there a `system recommendation' from an `AI method' that was part of a `system', whose `system developer' is known?
 } \\ \hline
 \end{tabular}%
 \end{adjustbox}
\end{table}

\section{Use Cases Represented Using the EO Model} \label{sec:usecases}
\subsection{Use Cases Represented by the EO}
To demonstrate the utility of the EO as a general-purpose ontology to represent explanations, we show how the EO's model can be used to compose explanations in five different use cases spanning food, healthcare, and finance domains (Tab. \ref{tab:usecasequestions}).  All of these use cases involve data available in the open domain. The first use case is in the food domain and based off of the FoodKG\footnote{\url{https://foodkg.github.io}}~\cite{haussmann2019foodkg}, and the rest of the use cases are from among those listed on the AIX-360 website\footnote{\url{https://aix360.mybluemix.net}}~\cite{aix360}. In AIX-360 usecases,
a suite of `explanation methods' belonging to the AIX-360 toolkit~\cite{arya2019one}, ~\cite{arya2022ai} are run. Additionally, each of the AIX-360 use cases has a technical description or Jupyter notebook tutorial
that system designers can comprehend and utilize to build the instance KGs. 

\begin{table}[hbt!]
\begin{center}
\caption{A listing of example questions and inferred explanation types supported by each use case.}
\label{tab:usecasequestions}

\resizebox{\textwidth}{!}{%
\begin{tabular}{|l|l|l|}
\hline
    \multicolumn{1}{|m{0.2\columnwidth}|}{\textbf{Use case}} & \multicolumn{1}{m{0.5\columnwidth}|}{\textbf{Example Questions}} & \multicolumn{1}{m{0.2\columnwidth}|}{\textbf{Explanation Types Inferred}}\\ \hline
     \multicolumn{1}{|m{0.2\columnwidth}|}{Drug Recommendation} & \multicolumn{1}{m{0.5\columnwidth}|}{Why Drug B and not Drug A?} & \multicolumn{1}{m{0.2\columnwidth}|}{Contrastive}\\ \hline
    \multicolumn{1}{|m{0.2\columnwidth}|}{Food Recommendation} & \multicolumn{1}{m{0.5\columnwidth}|}{Why should I eat spiced cauliflower soup? Why creamed broccoli soup over tomato soup?} & \multicolumn{1}{m{0.2\columnwidth}|}{Contextual and Contrastive}\\ \hline
    \multicolumn{1}{|m{0.2\columnwidth}|}{Proactive Retention} & \multicolumn{1}{m{0.5\columnwidth}|}{What is the retention action outcome for this employee?} & \multicolumn{1}{m{0.2\columnwidth}|}{Rationale}\\ \hline
    \multicolumn{1}{|m{0.2\columnwidth}|}{Health Survey Analysis} & \multicolumn{1}{m{0.5\columnwidth}|}{Who are the most representative patients in this questionnaire? Which questionnaires have the highest number of most representative patients?} & \multicolumn{1}{m{0.2\columnwidth}|}{Case based and Contextual}\\ \hline
     \multicolumn{1}{|m{0.2\columnwidth}|}{Medical Expenditure} & \multicolumn{1}{m{0.5\columnwidth}|}{What are the rules for expenditure prediction? What are patterns for high-cost patients?} & \multicolumn{1}{m{0.2\columnwidth}|}{Data}\\ \hline
      \multicolumn{1}{|m{0.2\columnwidth}|}{Credit Approval} & \multicolumn{1}{m{0.5\columnwidth}|}{What are the rules for credit approval? What are some representative customers for credit? What factors if present and if absent contribute most to credit approval?} & \multicolumn{1}{m{0.2\columnwidth}|}{Data, Case based and Contrastive}\\ \hline

\end{tabular}
}
\end{center}
\end{table}

In each of our six use cases, we assume that AI methods, primarily ML methods, have already been run and generated `system recommendations'.  The output of the AI methods may include enough data or may need to combined with content from a background KG (such as in the food use case) to generate the different explanation types that we support in the EO. Each of the use case has a set of example questions for which different explanation methods and/or ML methods are run. A listing of these example questions, as well as the explanation types inferred from running the reasoner on each use case's knowledge graph, is provided in Tab. \ref{tab:usecasequestions}.
Here, we describe each of the six use cases in a depth sufficient enough so that a system designer who wants to use the EO to instantiate outputs from their own use cases can seek guidance on building their use case KGs from the patterns that we use for these five exemplar KGs. Our use case KG files can also be downloaded from our Github: \url{https://github.com/tetherless-world/explanation-ontology/tree/master/usecases}.

\subsubsection{Drug Recommendation}
We demonstrate the use of EO in the design and operations of an AI system to support treatment decisions in the care of patients with diabetes. We previously conducted a two-part user-centered design study that focused on determining which explanations types are needed within such a system. We discovered that, of the types of explanations listed in Table 1, everyday and contextual explanations were required more than half the time. We noted that clinicians were using a special form of everyday explanations, specifically their experiential knowledge or \textit{`clinical pearls'} \cite{lorin2008clinical} to explain the patient's case. We observed concrete examples of the explanation components being used in the explanations provided by clinicians, such as \textit{`contextual knowledge'} of a patient's condition being used for diagnosis and drug \textit{`recommendations'}. Other examples of  explanation types needed within the system design include \textit{trace-based explanations} in a treatment planning mode, to provide an algorithmic breakdown of the guideline steps that led to a drug recommendation; \textit{`scientific explanations'} in a plan critiquing mode, to provide references to studies that support the drug, as well as \textit{`counterfactual explanations'}, to allow clinicians to add/edit information to view a change in the recommendation; and \textit{`contrastive explanations'} in a differential diagnosis mode, to provide an intuition about which drug is the most recommended for the patient. The results of the user studies demonstrated the need for a diverse set of explanation types and that modeling explanation requires various components to support AI system design. 

\begin{figure}[hbt!]
\centering
\includegraphics[width=1.0\linewidth]{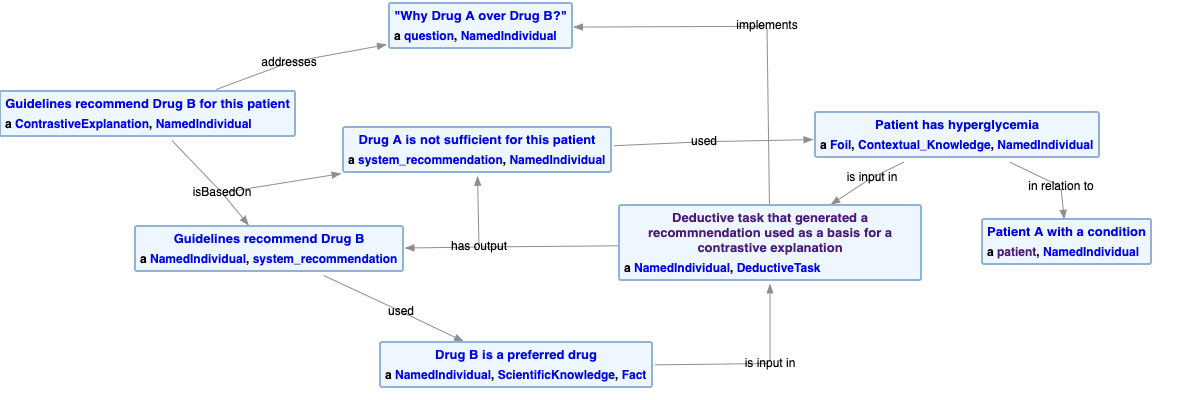}
\caption{A scientific explanation for drug recommendation modelled by the EO.}
\label{fig:eodrugusecae}  
\end{figure}

An example of the EO being used to represent the generation process for a \textit{`contrastive explanation'}, while accounting for the \textit{`reasoning mode,'} \textit{`AI Task'} involved, can be viewed in Fig. \ref{fig:eodrugusecae}. Upon identifying what explanation type would best suit this question, our ontology would guide the system to access different forms of \textit{`knowledge'} and invoke the corresponding \textit{`AI tasks'} that are suited to generate \textit{`contrastive explanations'}. In this example, a deductive AI task (:AITaskExample) is summoned and generates a system recommendation (:SystemRecExampleA) that Drug A is insufficient based on contextual knowledge of the patient record (:ContextualKnowledgePatient). In addition, the deductive task is also fed with guideline evidence that Drug B is a preferred drug, which results in the generation of a recommendation (:SystemRecExampleB) in favor of Drug B. Finally, our ontology would help guide a system to populate the components of a \textit{`contrastive explanation'} from \textit{`facts'} that supported the hypothesis, ``Why Drug B?'' and its \textit{`foil'}, ``Why not Drug A?,'' or the facts that ruled out Drug A. 

\subsubsection{Food Recommendation}

\begin{figure}[hbt!]
\centering
\includegraphics[width=1.0\linewidth]{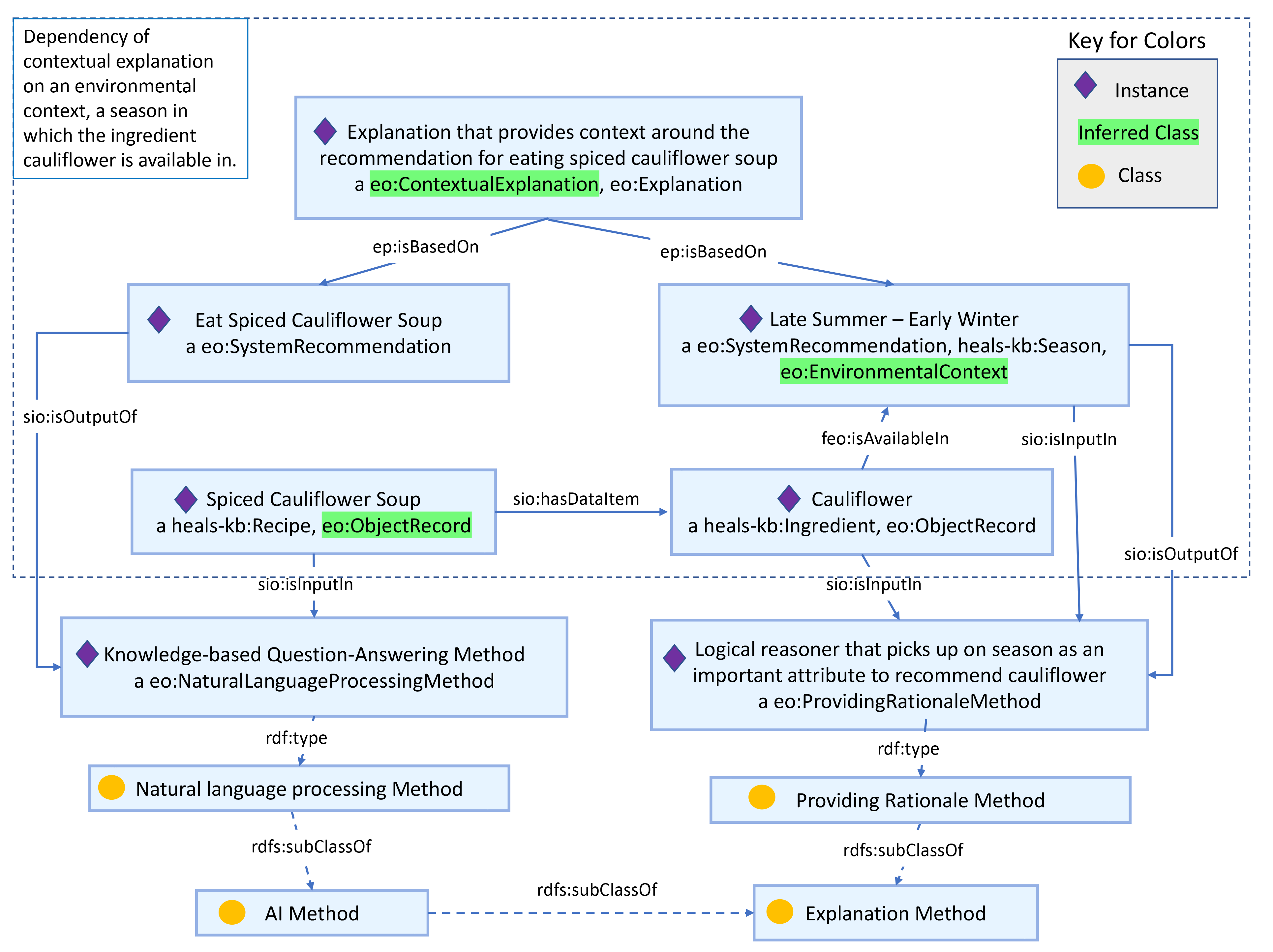}
\caption{A contextual explanation in a food recommender system modelled by the EO. Reproduced from S. Chari, O. Seneviratne, M. Ghalwash, S. Shirai, D.M. Gruen, P. Meyer, P. Chakraborty and D.L. McGuinness, ``Explanation ontology: A general-purpose, semantic representation for supporting user-centered explanations,'' \textit{Semantic Web J.}, vol. Pre-press, pp. 1 - 31, May 2023, doi: 10.3233/SW-233282, with permission from IOS Press. \copyright 2023}
\label{fig:eofoodusecase}  
\end{figure}

In the food recommendation use case, aimed at recommending foods that fit a person's preferences, dietary constraints, and health goals, we have previously published 
a customized version of the EO specifically for the food domain, the Food Explanation Ontology~\cite{padhiar2021semantic}. With the updates to the EO, we are now able to support the modeling of contextual and contrastive examples natively in the EO, whose capabilities were previously only in FEO as depicted in~\cite{padhiar2021semantic}. In the food use case, a knowledge base question-answering (QA) system~\cite{chen2021personalized} has been run and outputs answers to questions like ``What should I eat if I am on a keto diet?'' However, a standard QA system cannot directly address more complex questions that require a reasoner to be run on the underlying knowledge graph to generate inferred content. For example, questions such as whether or not one can eat a particular recipe, like ``spiced cauliflower soup,'' might not be easily addressable by the QA system because it doesn't know to specifically look for inferred information such as the seasonal context and availability of ingredients. 
The EO becomes useful here because the restrictions defined against the `environmental context' class can classify any seasonal characteristics defined against food as environmental context concerning that food `object record.' Hence, when we define an explanation for ``Why one should eat spiced cauliflower soup'' to be based on a seasonal characteristic, our EO reasoner can then classify the season instance to be an `environmental context' and therefore classify the explanation to be a `contextual explanation.' The contextual explanation instance and its dependencies can be viewed in Fig. \ref{fig:eofoodusecase}. Similarly, to address another question, ``Why is creamed broccoli soup recommended over tomato soup,'' we extract `facts' supporting creamed broccoli soup and `foils' or `facts' not in support of tomato soup. Then from this representation, our EO reasoner can infer that an `explanation' depending on the `system recommendation' and that encapsulates reasons in support of creamed broccoli soup over tomato soup, to be a contrastive explanation.  More broadly, if a system designer can define explanations dependent on food system recommendations and link the characteristics related to an `object record' contained in the recommendations. Then, when a reasoner is run on the EO, it could look for patterns that can match the user-centered explanation types we support and populate the explanation types whose patterns match the KG content. 

\subsubsection{Proactive Retention}
\begin{figure}[hbt!]
\centering
\includegraphics[width=1.0\linewidth]{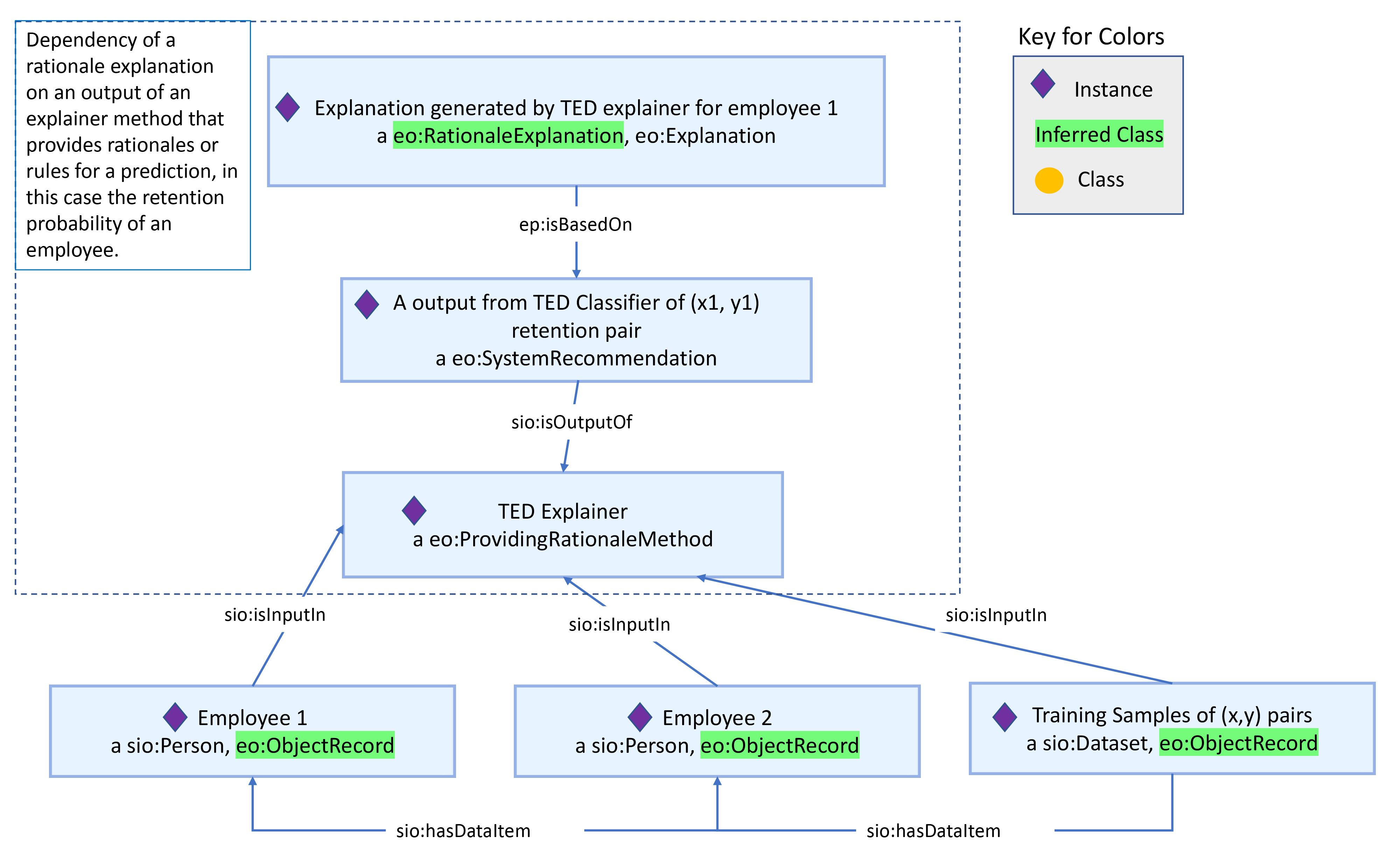}
\caption{A rationale explanation in a proactive retention use case modeled by the EO. Reproduced from S. Chari, O. Seneviratne, M. Ghalwash, S. Shirai, D.M. Gruen, P. Meyer, P. Chakraborty and D.L. McGuinness, ``Explanation ontology: A general-purpose, semantic representation for supporting user-centered explanations,'' \textit{Semantic Web J.}, vol. Pre-press, pp. 1 - 31, May 2023, doi: 10.3233/SW-233282, with permission from IOS Press. \copyright 2023}
\label{fig:eoproactiveretention}  
\end{figure}

In the proactive retention use case\footnote{Proactive retention use case: \url{https://nbviewer.org/github/IBM/AIX360/blob/master/examples/tutorials/retention.ipynb}}, the objective is to learn the rules for employee retention that could signal to an employing organization whether or not employees are likely to have retention potential. Since these rules involve a deep understanding of the contributing attributes of the employee dataset, a domain expert is best adept at providing these rules. Potentially, a supervised ML method could be run to learn the rules for other unlabeled instances. The AIX-360 toolkit supports a TED Cartesian Explainer algorithm~\cite{hind2019ted} that can learn from rules that are defined against a few cases and predict the rules for others. More specifically, in the proactive retention use case KG, the TED Cartesian Explainer method is defined as an instance of `Providing rationale method' in the EO by virtue of the definition for this `explanation method' subclass, and that the TED Explainer provides local explanations for each instance. Additionally, since the TED Explainer provides `system recommendations' for every employee retention output pair, we save this dependency of explanations on local instance-level system recommendations in the KG and define explanations based on these retention rule explanations. As can be seen from the `explanation types supported' column of the proactive retention row, a reasoner infers rationale explanations, see Fig. \ref{fig:eoproactiveretention}, from the proactive retention KG. These rationale explanations are inferred that way because they match the equivalent class restriction of the `rationale explanation' class in the EO, wherein we look for rationales or traces supporting a system recommendation, which in this case are the rules providing rationales for the employee retention output.

\subsubsection{Health Survey Analysis}
The health survey analysis use case\footnote{NHANES use case documentation: \url{https://nbviewer.org/github/IBM/AIX360/blob/master/examples/tutorials/CDC.ipynb}} utilizes the National Health and Nutritional Exam Survey (NHANES) dataset~\cite{nhanes1516}. 
The objective in this use case entails two `explanation tasks': (1) find the most representative patients for the income questionnaire, and (2) find the responses that are most indicative/representative of the income questionnaire. The Protodash method~\cite{gurumoorthy2019efficient}, an `Exemplar explanation method,' that finds representative examples from datasets, is run for both these tasks. However, in the second task, an additional data interpretation or summarization method is run to evaluate the prototype patient cases of questionnaires for how well they correlate to responses in the income questionnaire. Hence, when we instantiate the outputs of these two tasks, we use different chains of representations to indicate the dependencies of the explanations of the two questions that are addressed by these tasks. More specifically, this chaining would mean that we define that the `system recommendation' of the `summarization' method instance to be dependent on or use as input the `system recommendations' of `Protodash' instances. 

Additionally, as can be seen from the `explanation types' supported column of Tab. \ref{tab:usecasequestions} against the health survey analysis, there are two explanation types inferred upon running a reasoner against the NHANES KG: case-based and contextual. The explanation finding for the most representative patients of the income questionnaire is case-based since it contains patient case records. 

\begin{figure}[hbt!]
    \centering
    \lstset{language=Manchester, basicstyle=\ttfamily\fontsize{9}{10}\selectfont, columns=fullflexible, xleftmargin=5mm, framexleftmargin=5mm, numbers=left, stepnumber=1, breaklines=true, breakatwhitespace=false, numberstyle=\ttfamily\fontsize{9}{10}\selectfont, numbersep=5pt, tabsize=2, frame=lines}
\begin{lstlisting} 
Class: eo:ContextualKnowledge
  EquivalentTo:
    (`social entity' or sio:representation or sio:media or sio:Location)
 and (eo:'has attribute' some eo:'Object Record')
  SubClassOf:
    eo:ContextualKnowledge
\end{lstlisting}
 \caption{OWL expression of the \textit{`environmental context'} and its sufficiency conditions in Manchester syntax. 
}
\label{lst:environmentalcontext}
\end{figure}

\begin{figure}[hbt!]
\centering
\includegraphics[width=1.0\linewidth]{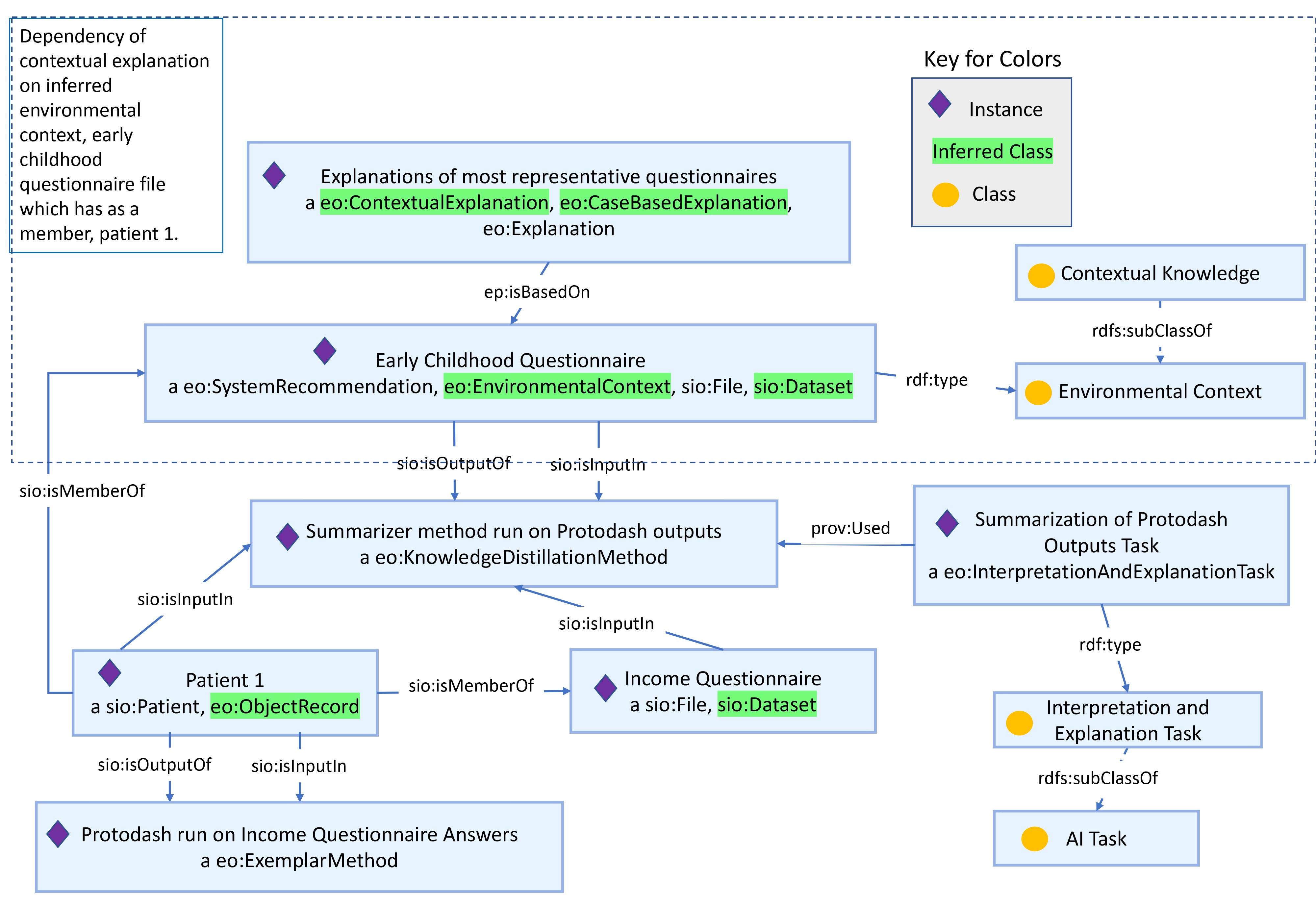}
\caption{A contextual explanation from the health survey analysis use case modeled by the EO. Reproduced from S. Chari, O. Seneviratne, M. Ghalwash, S. Shirai, D.M. Gruen, P. Meyer, P. Chakraborty and D.L. McGuinness, ``Explanation ontology: A general-purpose, semantic representation for supporting user-centered explanations,'' \textit{Semantic Web J.}, vol. Pre-press, pp. 1 - 31, May 2023, doi: 10.3233/SW-233282, with permission from IOS Press. \copyright 2023}
\label{fig:eonhanes}  
\end{figure}

Still, the classification of the explanations of the most representative questionnaire as a contextual explanation is less obvious. However, upon closer investigation of the equivalent condition defined against the `environmental context` class in the EO from Listing \ref{lst:environmentalcontext}, we can see how the patients' questionnaires, a type of `file,' are inferred to be the `environmental context' for the patient, an `object record', since the patients participate in those questionnaires. Hence, an explanation, such as finding the most representative questionnaire that is dependent on both the questionnaire and the representative patient cases, would be classified as a contextual explanation (see Fig. \ref{fig:eonhanes}). Such contextual explanations can help identify which parts of the larger context were impactful in the system recommendation and help system designers and developers better explain their system workings to end-users.

\subsubsection{Medical Expenditure}
\begin{figure}[hbt!]
\centering
\includegraphics[width=0.9\linewidth]{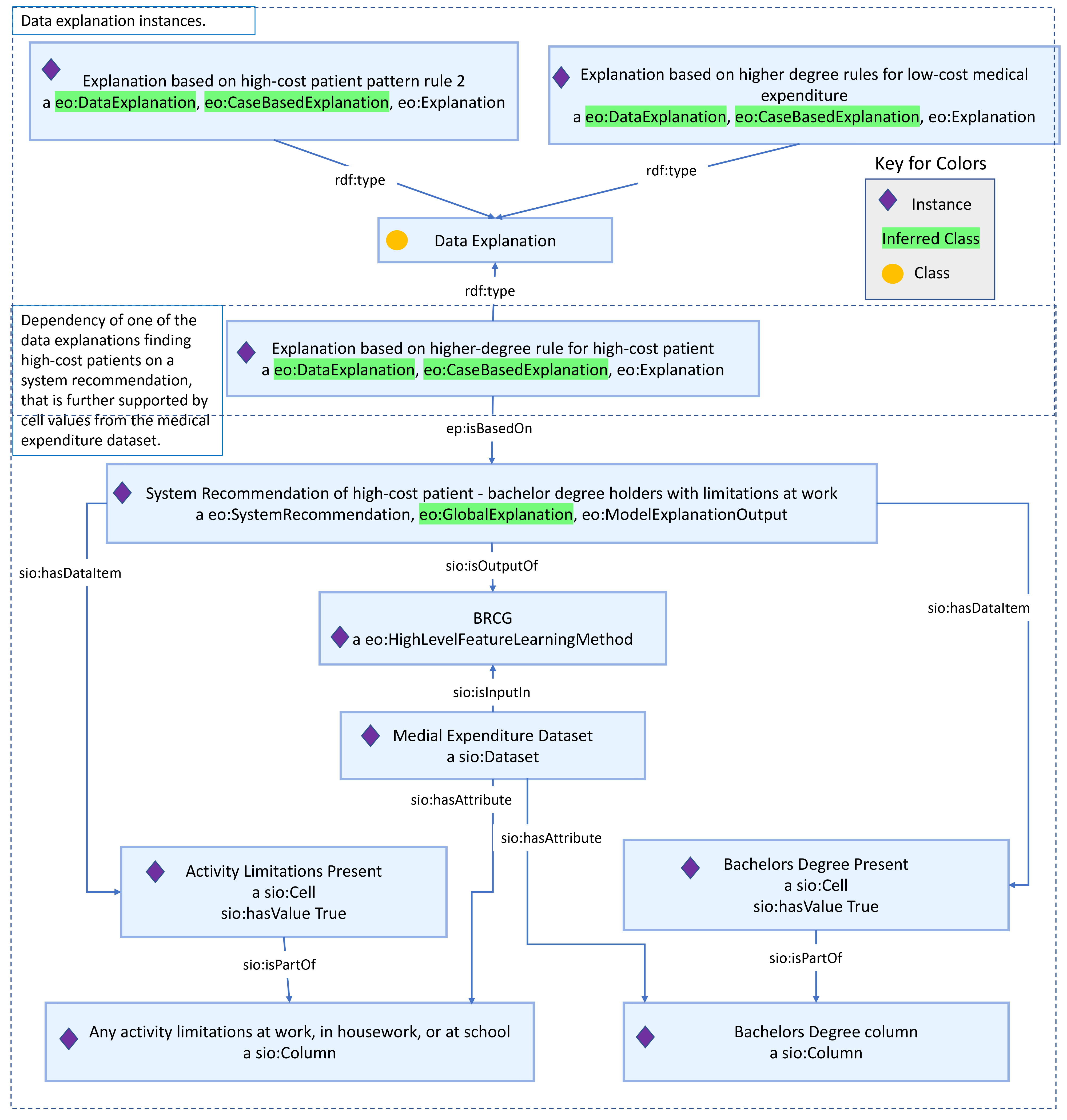}
\caption{A data explanation from the medical expenditure use case modeled by the EO. Reproduced from S. Chari, O. Seneviratne, M. Ghalwash, S. Shirai, D.M. Gruen, P. Meyer, P. Chakraborty and D.L. McGuinness, ``Explanation ontology: A general-purpose, semantic representation for supporting user-centered explanations,'' \textit{Semantic Web J.}, vol. Pre-press, pp. 1 - 31, May 2023, doi: 10.3233/SW-233282, with permission from IOS Press. \copyright 2023}
\label{fig:eomedicalexpenditure}  
\end{figure}

In the medical expenditure use case\footnote{Medical expenditure: \url{https://nbviewer.org/github/IBM/AIX360/blob/master/examples/tutorials/MEPS.ipynb}}, the objective is to learn the rules for the demographic and socio-economic factors that impact the medical expenditure patterns of individuals. Hence, from the description itself, we can infer that in this use case the rules are attempting to understand the patterns in the Medical Expenditure Panel Survey (MEPS) dataset or the general behavior of the prediction models being applied on the dataset, as opposed to attempting to understand why a particular decision was made. This use case involves the use of global explanation methods from the AIX-360 toolkit~\cite{arya2019one}, ~\cite{arya2022ai}, including the Boolean Rule Column Generator and Linear Rule Regression (LRR) methods. The `explanation method' instances, in this case, produce explanations that are dependent on `system recommendations,' which rely on the entire dataset itself. Therefore, system designers should link the `system recommendations' to the dataset. We achieve this association in our MEPS KG by representing the `dataset' instance as an input of the LRR and BRCG methods. Finally, as can be seen from Tab. \ref{tab:usecasequestions}, the reasoner can only infer data explanations (refer to Fig. \ref{fig:eomedicalexpenditure}) from the MEPS KG instances, as the explanations are dependent on the rules identified for patterns in the dataset. Hence, in a use case such as this, wherein the explanations are mainly dependent on the dataset and the patterns within them, if a system designer were to appropriately link the system recommendation that the explanation is based on to the entire dataset or a component of the same (i.e., a column, row or cell), a reasoner run on the use case KG can populate data explanations of the explanation representations. Such data explanations can be helpful to understand aspects of bias and coverage in the data, which can signal to system designers and users whether their dataset is serving its intended purpose.

\subsubsection{Credit Approval}
\begin{figure}[hbt!]
\centering
\includegraphics[width=0.8\linewidth]{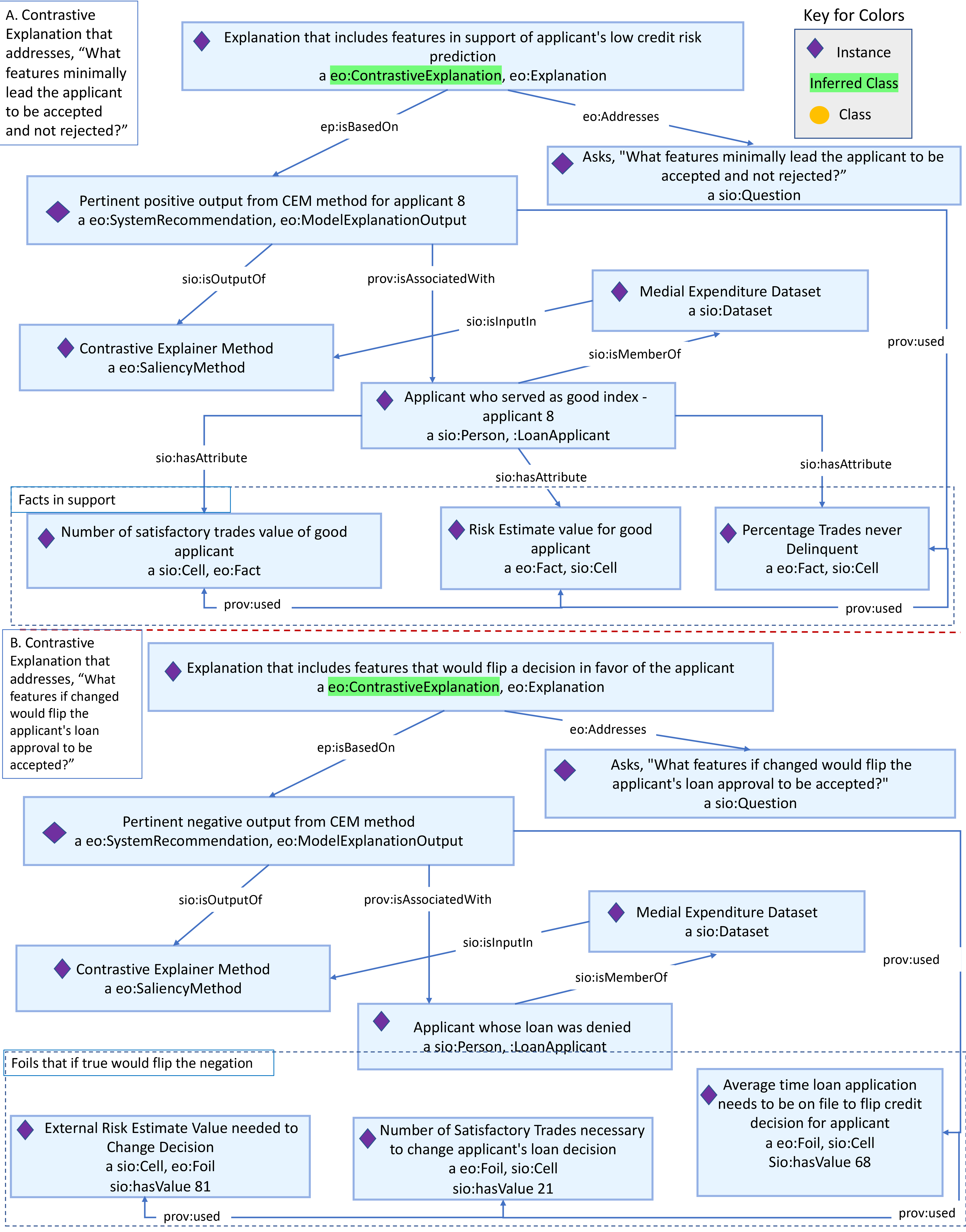}
\caption{A contrastive explanation from the credit approval use case modeled by the EO. Reproduced from S. Chari, O. Seneviratne, M. Ghalwash, S. Shirai, D.M. Gruen, P. Meyer, P. Chakraborty and D.L. McGuinness, ``Explanation ontology: A general-purpose, semantic representation for supporting user-centered explanations,'' \textit{Semantic Web J.}, vol. Pre-press, pp. 1 - 31, May 2023, doi: 10.3233/SW-233282, with permission from IOS Press. \copyright 2023}
\label{fig:eocreditapproval}  
\end{figure}

In the credit approval use case\footnote{Credit approval: \url{https://nbviewer.org/github/IBM/AIX360/blob/master/examples/tutorials/HELOC.ipynb}}, there are several objectives depending on the `user' persona, including to enable data scientists to familiarize themselves with the factors that impact the credit approval outcome, for loan officers to identify prototypical cases of credit approved owners, and for customers to understand what patterns in their profile contribute the most towards their credit approval. The analyses are conducted on the FICO HELOC dataset\footnote{FICO HELOC Dataset: \url{https://aix360.readthedocs.io/en/latest/datasets.html\#id16}}, which contains ``anonymized information about Home Equity Line Of Credit (HELOC) applications made by real homeowners''~\cite{ficoheloc}. We run three explanation methods: (1) BRCG and LRR to provide data scientists with the rules for credit approval ratings, (2) Protodash to provide loan officers prototypical customer cases, and (3) Contrastive Explanation Method (CEM) to provide customers with explanations to what the minimally sufficient factors in achieving good credit (`fact') are and the factors which, if changed, would change their credit (`foil'). In the medical expenditure use case, we have already shown that a system designer dealing with outputs of rule-based methods, such as BRCG and LRR, can represent the explanations dependent on `system recommendations' generated by the methods and, particularly in this use case, define the FICO HELOC dataset as input for these methods. In the case of representing the identified prototypical credit approval customers, system designers can seek inspiration from the health survey analysis case and similarly represent the customer cases as instances of `system recommendations' and as inputs to an `explanation task.' Finally, for the outputs of CEM (see Fig. \ref{fig:eocreditapproval}), we represent the factors that need to be minimally present for credit approval as `facts' in support of an explanation, and the factors which, if present, flip the decision as `foils' of an explanation. Such a representation would align with our definition of restrictions against the contrastive explanation class. In addition, we create \emph{three different user instances} for the data scientist, loan officer, and customer, respectively, and further associate the questions (see credit approval row of Tab. \ref{tab:usecasequestions}) that each of them asks via the properties supported in the EO (Fig. \ref{fig:eomodel}). When a reasoner is run upon the credit approval KG, we can see instances of data explanations for a data science user, case based explanations for a loan officer, and contrastive explanations for a customer.

\section{Resource Contributions}
\label{sec:contributions}
We contribute the following publicly available artifacts: our expanded Explanation Ontology with the logical formalizations of the different explanation types and SPARQL queries to evaluate the competency questions, along with the applicable documentation, all available on our resource website. On our open-source Github repository, we also release our KG files (and the inferred versions too), for the five use cases (Sec. \ref{sec:usecases}) described in this chapter. These resources, listed in Table \ref{tab:resourcetable}, are useful for anyone interested in building explanation facilities into their systems.

The ontology has been made available as an open-source artifact under the Apache 2.0 license \cite{apache2}
and we maintain all our artifacts on our Github repository. We also maintain a persistent URL for our ontology, hosted on the PURL service. All the relevant links are listed below in Tab. \ref{tab:resourcetable}.

\begin{table}[hbt!]
\centering
\caption{Links to resources we have released and refer to in the paper.}
\label{tab:resourcetable}
\begin{tabular}{|l|l|}
\hline
\textbf{Resource}             & \textbf{Link to Resource}                                                  \\ \hline
Resource Website   & \url{http://tetherless-world.github.io/explanation-ontology}   \\ \hline
EO PURL URL        & \url{https://purl.org/heals/eo}                                \\ \hline
Github Repository  & \url{https://github.com/tetherless-world/explanation-ontology} \\ \hline
\end{tabular}%
\end{table}

\section{Evaluation} \label{sec:evaluation}

Our evaluation is inspired by ontology evaluation techniques proposed in Muhammad et al.'s comprehensive ontology evaluation techniques review paper{~\cite{amith2018assessing}}. They introduce an ontology evaluation taxonomy that combines evaluation techniques that each reveal different perspectives of the ontology, such as application-based, metric-based, user-based, and logic/rule-based evaluation techniques. These evaluation techniques in Muhammad \textit{et al.'s} taxonomy are a collection of techniques proposed by four different well-cited papers, including Obrst \textit{et al.}{~\cite{obrst2007evaluation}}, Duque-Ramos \textit{et al.}, Tartir et al.{~\cite{tartir2010ontological}} and Brank \textit{et al.}{~\cite{brank2005survey}}. From this taxonomy, we evaluate our ontology by addressing a representative range of competency questions that illustrate the task-based and application-based capabilities of the EO. We also evaluate the EO by applying the evolution-based technique proposed by Tartir \textit{et al.}{~\cite{tartir2010ontological}} and analyzing the capabilities introduced by version 2.0 of our ontology.

We evaluate the task-based and application-based abilities of the EO to assist system designers in providing support and include user requirements, address explanation dependent questions across the illustrated use cases. In Tab. \ref{tab:competencyquestion} and \ref{tab:competencyquestionsusecase}, we present the competency questions that we have developed to evaluate the task-based and application-based capabilities of the EO, respectively. In each of these tables, we show the setting for the competency question related to the question and its answer. These competency questions are realized via SPARQL queries that are run on the EO or its companion use case KGs (Sec. \ref{sec:usecases}). These SPARQL queries can be browsed through our resource website (Sec. \ref{sec:contributions}). Additionally, as part of the answers, we also include additional metrics in both Tab. {\ref{tab:competencyquestion}} and {\ref{tab:competencyquestionsusecase}}, to help assess the complexity of addressing the competency questions and we borrow these metrics from Kendall and McGuinness's recent Ontology Engineering{~\cite{kendall2019ontology}} book. These metrics include 
the \textit{overall query length for addressing the competency question}, \textit{were any property restrictions accessed in retrieving the answer}, \textit{did a reasoner need to be run for the result} and \textit{finally were there any filter clauses required to narrow down the result}. 
\subsection{Evolution-based Evaluation} \label{sec:evolutionbasedevaluation}
We also evaluate the additions to the EO model since its first iteration described in Chari \textit{et al.}~\cite{chari2020explanation} using the evolution-based evaluation method mentioned in Muhammad \textit{et al.}'s taxonomy~\cite{amith2018assessing}. However, as analyzed by Muhammad \textit{et al.}~\cite{amith2018assessing}, it is hard to quantify the evolution-based evaluation technique of Tartir \textit{et al.}~\cite{tartir2010ontological} for knowledge gain provided by the updates to the ontology model. From a qualitative assessment, we find that the additions to the EO model helped us better represent capabilities including :

\begin{itemize}
\item Capture more granular representations of `AI methods' and their interactions with the explanation types, and support more ways to generate explanations.
\item Introduce characteristics at various strategic attributes that contribute to explanations (e.g., at the system, user, and object classes), which provide the flexibility to define characteristics at multiple levels and allow for better considering explanation types through the restriction of equivalent classes. For example,  we could better represent restrictions against the `contextual knowledge' class with the broader characteristic scope, and therefore, more patterns can be considered as matches for this class, which is a primary contributor to the `contextual explanation' type.
\item Include more of the contributing attributes of the explanation ecosystem itself (such as capturing the `system' in which the `AI methods' are run), which helps maintain better provenance of the infrastructure contributing to the explanations.
\end{itemize}

\subsection{Task-based Evaluation} \label{sec:taskbasedevaluation}
With the increasing demand to support explainability as a feature of user-facing applications, thus improving uptake and usability of AI and ML-enabled applications~\cite{doshi2017towards}, ~\cite{gilpin2018explaining}, ~\cite{chari2020directionschapter}, ~\cite{dey2022xai}, it is crucial for system designers to understand how to support the kinds of explanations needed to address end user needs. An evolving landscape of the explanations, goals, and methods that support them, complicates the task, but querying the EO can help answer such questions in a standalone format. For the task-based abilities, we aim to showcase how the EO can provide ``human ability to formulate queries using the query language provided by the ontology''~\cite{obrst2007evaluation} and ``the degree of explanation capability offered by the system, the coverage of the ontology in terms of the degree of reuse across domains''~\cite{obrst2007evaluation}.

We detail some of the support that the EO can provide to help system designers understand the main entities interacting with explanations in Tab. \ref{tab:competencyquestion}. 
The support that we illustrate includes querying the `AI Method' and `AI Task' that generates the explanation, the example questions that different explanation types can address, and the more nuanced parts of explanation types, such as their components and when they can be generated. The table also shows a set of competency questions and answers that are retrieved from the EO. For answers in this table, we use, for better understanding, simpler descriptions than the results returned by the SPARQL query. The full set of results can be browsed through our resource website\footnote{SPARQL query results: \url{https://tetherless-world.github.io/explanation-ontology/competencyquestions/}}. 


\begin{table}[hbtp!]
\centering
\caption{A catalog of competency questions and candidate answers produced by our EO.}
\label{tab:competencyquestion}
\begin{tabular}{|p{0.1\linewidth}|p{0.2\linewidth} | p{0.2\linewidth} | p{0.08\linewidth}| p{0.08\linewidth}| p{0.08\linewidth}| p{0.08\linewidth}|}
\hline
Setting &
  Competency Question &
  Answer &
  SPARQL Query length &
  Property Restrictions accessed? &
  Inference Required? &
  Filter Statements \\
  \hline
System Design &
  Q1. Which AI model(s) is/are capable of generating this explanation type (e.g. trace-based)? &
  Knowledge-based systems, Machine learning model: decision trees &
  8 &
  Yes &
  No &
  No \\
  \hline
System Design &
  Q2. What example questions have been identified for counterfactual explanations? &
  What other factors about the patient does the system know of? What if the major problem was a fasting plasma  glucose? &
  4 &
  No &
  No &
  No \\
  \hline
System Design &
  Q3. What are the components of a scientific explanation? &
  Generated by an AI Task, Based on recommendation, and based on evidence from study or basis from scientific method &
  2 &
  Yes &
  No &
  No \\
  \hline

\end{tabular}
\end{table}

  \begin{table}[ht!]
  \centering
\caption{A catalog of competency questions and candidate answers produced by our EO.}
\label{tab:competencyquestion1}
\begin{tabular}{|p{0.1\linewidth}|p{0.2\linewidth} | p{0.2\linewidth} | p{0.08\linewidth}| p{0.08\linewidth}| p{0.08\linewidth}| p{0.08\linewidth}|}
\hline
Setting &
  Competency Question &
  Answer &
  SPARQL Query length &
  Property Restrictions accessed? &
  Inference Required? &
  Filter Statements \\
  \hline
  System Analysis &
  Q4. Given the system has ranked specific recommendations by comparing different medications, what explanations can be provided for that recommendation? &
  Contrastive explanation &
  8 &
  Yes &
  No &
  No \\ \hline
System Analysis &
  Q5. Which explanation type best suits the user question asking about numerical evidence, and how does a system generate such an answer? &
  Explanation type: statistical;   System: run `Inductive' AI task with `Clustering' method to generate  numerical evidence &
  18 &
  Yes &
  No &
  No \\
  \hline
System Analysis &
  Q6. What is the context for  data collection and application of the contextual explanation, say for example from the health survey analysis use case? &
  Explanation type: contextual  Environmental context: `Early childhood questionnaire' in a US location &
  5 &
  No &
  Yes &
  No \\
  \hline
\end{tabular}
\end{table}

We split the questions across two settings, including during \textbf{\textit{system design}} (questions 1 - 3 in Tab. \ref{tab:competencyquestion}) when a system designer is planning for what explanation methods and types of support are needed, based on the user and business requirements. The other setting, during \textbf{\textit{system analysis}} (questions 4 - 6 in Tab. {\ref{tab:competencyquestion}} and Tab. \ref{tab:competencyquestion1}), when they are trying to understand what explanation types can be supported given system outputs at their disposal and/or the dependencies of the explanation type instances in their use case KGs on its attributes, such as the system state and context. 

Delving further into answers for system design questions from Tab. \ref{tab:competencyquestion} such as `What particular explanation type addresses a prototypical question?' can signal to system designers what explanation types can best address the user questions in their use case. Additionally, knowledge of what components populate certain explanation types (Q1. and Q3.) can help system designers plan for what inputs they would need to support these explanation types in their use case. 

Similarly, in system analysis settings, which could include, once the system designer knows what system outputs they can build explanations of (Q4 - 5.), or once a reasoner has been run on the use case KG, or once the explanations inferred via the EO have been displayed (Q6.), a system designer might need to be ready for additional questions that users or system developers might have. These include questions such as `What is the context for the data populating the explanation?' or `In what system setting did this explanation get populated?' Upon querying the EO or use case KGs, we can support answers to such system analysis questions, giving system designers more insights around explanations supported in or to support in their use cases. Besides the general questions about the explanations, a system designer might need to assist interface designers or users with more domain-specific questions about the explanation instances. We can support these through the domain-level instantiations that we allow for in the use case KGs. Examples of these application-based evaluations can be seen in Tab. \ref{tab:competencyquestionsusecase} and are described in the next section. 

In summary, the competency questions we address in Tab. {\ref{tab:competencyquestion} 
provide examples for a task-based evaluation of the EO as a model to support user-centered explanations and we checked the utility of these questions against a small expert panel of system designers from our lab. We report their insights in the discussion section (Sec. {\ref{sec:discussion}}). We are 
soliciting additional suggestions 
for more types of questions that showcase a broader range of the EO's capabilities and suggestions can be submitted via recommendations on our website\footnote{Call for participation: \url{https://tetherless-world.github.io/explanation-ontology/competencyquestions/}}.}

\subsection{Application-based Evaluation} \label{sec:appbasedevaluation}
When system designers represent use case specific content, they often need to communicate these representations to other teams, including interface designers who support these use case KGs on UIs, or system developers who want to ensure that they correctly capture system outputs. Hence, in such scenarios, the system designers would need to provide results concerning the domain-specific content in their use case KGs, which could span questions like ``What entities are contained in the contrastive explanation?" to more specific questions about particular objects in the use case, such as ``What is the outcome for employee 1?'' Through the EO, we enable the system designer to provide application-specific details about the explanations supported in their use cases, that improve the presentation of the ``output of the applications''~\cite{brank2005survey}.

\begin{table}[ht!]
\centering
\caption{Example questions that can be asked of our use case knowledge graphs that are modeled using our EO.}
\label{tab:competencyquestionsusecase}
\begin{tabular}{|p{0.1\linewidth}|p{0.2\linewidth} | p{0.2\linewidth} | p{0.08\linewidth}| p{0.08\linewidth}| p{0.08\linewidth}| p{0.08\linewidth}|}
\hline
Use Case &
  Competency Question &
  Answer &
  SPARQL Query Length &
  Property Restrictions Accessed? &
  Inference Required? &
  Filter Statements \\ \hline
Food Recommendation &
  Q1. What explanation types are supported? &
  Contextual and Contrastive &
  2 &
  No &
  Yes &
  No \\ \hline
Food Recommendation &
  Q2. Why should I eat spiced cauliflower soup? &
  Cauliflower is in season. &
  5 &
  No &
  Yes &
  Yes \\ \hline
Proactive Retention &
  Q2. What is the retention action outcome for employee 1? &
  Employee 1 is likely to remain in the same organization. &
  4 &
  No &
  No &
  Yes \\ \hline
Health Survey Analysis &
  Q3. Who are the most representative patients in the income questionnaire? &
  Patient 1 and 2. &
  3 &
  No &
  Yes &
  No \\ \hline
Health Survey Analysis &
  Q5. Which questionnaire did patient 1 answer? &
  Income, early childhood and social determiners. &
  3 &
  No &
  Yes &
  Yes \\ \hline
Medical Expenditure &
  Q6.What are the rules for high-cost expenditure? &
  Individuals are in poor health, have limitations in physical functioning and are on health insurance coverage. &
  4 &
  No &
  Yes &
  Yes \\ \hline
Credit Approval &
  Q7. What factors contribute most to a loan applicants credit approval? &
  Facts: Number of satisfactory trades and risk estimate value &
  7 &
  No &
  Yes &
  No \\ \hline
\end{tabular}
\end{table}

Some examples of questions that we can handle for the use cases we support from Sec. \ref{sec:usecases}, are shown in Tab. \ref{tab:competencyquestionsusecase}. In these instances from Tab. \ref{tab:competencyquestionsusecase}, we query the use case KGs for domain-specific content by leveraging the properties of the EO model, as defined between the entities that contribute to explanation instances in these KGs.

Some examples of domain-specific content that can be queried (see Tab. \ref{tab:competencyquestionsusecase}) include questions like, `what are the facts contributing to a contrastive explanation (Q7)', `what is the `system recommendation' linked to employee 1 (Q3)', and `what are some rules for trace-based explanations in the KG (Q6)?' Answers to such questions can provide insights on the entities contributing to explanations. More specifically, the answers to questions in Tab. \ref{tab:competencyquestionsusecase}, can help system designers convey to interface designers what entities can be shown on the UI concerning these explanations (Q1, Q3 - Q4) or even help application users and system developers navigate the explanation dependencies to understand the interactions between the entities contributing to these user-centered explanations (Q2, Q5 - Q7). An example of a SPARQL query that implements ``Q6. What are the rules for high-cost patient expenditure?'' from Tab. \ref{tab:competencyquestionsusecase}, can be viewed in Listing \ref{lst:sparqlqueryapplication} and the results can be seen in Tab. \ref{tab:resultstableapplication}.

In summary, representing explanations via the EO model allows the content supporting the explanations to be queried easily and through multiple depths of supporting provenance. This is also a step toward supporting an interactive design for explanations, such as the one mentioned in Lakkaraju et al.~\cite{lakkaraju2022rethinking}. We aim for these example questions addressed through our use case KGs, to provide guidance to system designers on the types of questions that they could support in their own use case KGs.

\begin{figure}[hbt!]
    \centering
\lstset{language=SPARQL, basicstyle=\ttfamily\fontsize{9}{10}\selectfont, xleftmargin=5mm, framexleftmargin=5mm, numbers=left, stepnumber=1, breaklines=true, breakatwhitespace=false, numbersep=5pt, tabsize=2, frame=lines}
\begin{lstlisting} 
PREFIX rdf: <http://www.w3.org/1999/02/22-rdf-syntax-ns#>
PREFIX rdfs: <http://www.w3.org/2000/01/rdf-schema#>
PREFIX eo: <https://purl.org/heals/eo#>
PREFIX ep: <http://linkedu.eu/dedalo/explanationPattern.owl#>
PREFIX sio: <http://semanticscience.org/resource/>

SELECT ?subject ?data
	WHERE { 
?subject a eo:DataExplanation . 
?subject rdfs:label ?sl .
?subject ep:isBasedOn ?o .
?o sio:SIO_001277 ?data .
  filter( regex(str(?sl), "high-cost", "i" ))

}
\end{lstlisting}
 \caption{A SPARQL query run on the medical expenditure knowledge graph, that retrieves the rules associated with a data explanation for high-cost expenditure to answer a competency question of the kind, ``What are the rules for high-cost expenditure?''. Reproduced from S. Chari, O. Seneviratne, M. Ghalwash, S. Shirai, D.M. Gruen, P. Meyer, P. Chakraborty and D.L. McGuinness, ``Explanation ontology: A general-purpose, semantic representation for supporting user-centered explanations,'' \textit{Semantic Web J.}, vol. Pre-press, pp. 1 - 31, May 2023, doi: 10.3233/SW-233282, with permission from IOS Press. \copyright 2023}
  \label{lst:sparqlqueryapplication}
\end{figure}

\begin{table}[hbt!] 
\begin{center}
\caption{Results of the SPARQL query to retrieve the higher-degree rules associated with a data explanation instance from the medical expenditure use case.}
\label{tab:resultstableapplication}
\begin{adjustbox}{max width=1.0\textwidth,center}
\begin{tabular}{|l|l|}
\hline
    \multicolumn{1}{|m{0.5\columnwidth}|}{Explanation} & \multicolumn{1}{m{0.5\columnwidth}|}{Rule} \\ \hline
     \multicolumn{1}{|m{0.5\columnwidth}|}{Explanation based on high-cost patient pattern 2.} & \multicolumn{1}{m{0.5\columnwidth}|}{Self-reported poor health - true} \\ \hline
     \multicolumn{1}{|m{0.5\columnwidth}|}{Explanation based on high-cost patient pattern 2.} & \multicolumn{1}{m{0.5\columnwidth}|}{Limitations in physical functioning - present} \\ \hline
      \multicolumn{1}{|m{0.5\columnwidth}|}{Explanation based on high-cost patient pattern 2.} & \multicolumn{1}{m{0.5\columnwidth}|}{Health insurance coverage - present} \\ \hline
\end{tabular}
\end{adjustbox}
\end{center}
\end{table}

\section{Discussion} \label{sec:discussion}
Here we discuss different aspects of the EO, including desired features and design choices, the relevance of results and limitations, and future outlook.

\ul{\textit{How is EO considered to be general-purpose?}}
We have described a general-purpose and mid-level ontology, the EO, that can be leveraged to represent user-centered explanations in domain applications.
In the EO, we encode the attributes that contribute to explanations in a semantic representation as an ontology, and by doing so, the EO can be used to structure explanations based on linkages to the needs the explanations support and the method chains that generate them (Sec. \ref{sec:eomodel}). 
The ability to support different explanation types enables the EO to be a tool that can provide users with different views required to reason over the recommendations provided by AI-enabled systems. In addition to explainability being diverse in terms of the different types of needs that users seek from explanations (Fig. \ref{fig:eoexplanations}), explanations are also domain-specific and are driven by the system outputs and domain knowledge in the application domain or use case~\cite{lakkaraju2022rethinking},~\cite{chari2021leveraging}. Hence, although the EO is a mid-level ontology, it needs to be broad enough to support the representation of explanations across domains. Through our use case KGs (Sec. \ref{sec:usecases}), we have demonstrated how system designers, as our intended users of the EO, can root the domain-specific concepts in our EO model and, upon running a reasoner on their KGs, classify their representations into the user-centered explanation types we support in the EO. 

\ul{\textit{What features were introduced in EO 2.0 and why?}}
This paper is a significantly expanded version of an earlier conference paper on the EO~\cite{chari2020explanation}, and here we also introduce new use cases and explanation types. With the introduction of the six new explanation types from Zhou et al.~\cite{zhou2021evaluating}, we can now infer a broader range of user-centered explanation types in use cases. For example, we describe here how to infer data and rational explanations in the proactive retention, credit approval, and medical expenditure use cases. Also, without expanding the explanation method trees in the EO, we would not have been able to adequately capture the outputs from explanation methods in the AIX-360 use cases. Hence, our additions to the EO model introduced in this paper can provide expanded expressivity, leading to better applicability with the current explainable AI landscape. 

\ul{\textit{What are design choices made during EO development and how do they apply?}}
The EO itself is lightweight, and we have imported two standard scientific ontologies, SIO-O and Prov-O. We reuse a lot of classes and we only introduce classes and properties that do not exist currently to represent explanations. This design choice of reuse is reflected in the ontology metrics reported in Tab. {\ref{tab:ontology-metrics}} and is a feature which can ensure interoperability of the EO with other ontologies that also use the standard scientific ontologies that we reuse. From a modeling perspective, in the EO, we are aiming for expressivity in that we capture multiple paths to compose explanation types which contribute to a somewhat slow reasoner performance. 
Since the reasoner is typically run only once in a use case, or it is a process that is not run in real-time, the speed limitation might not be an issue when displaying such pre-computed results. 

\ul{\textit{What is the coverage of the EO and how can the ontology be adapted?}}
The EO provides an approximate representation of content in the evolving literature surrounding XAI. From a modeling perspective, we have modeled attributes of explanations that we deemed essential to represent explanations that address the standard set of question templates identified in Vera et al.~\cite{liao2020questioning} (e.g., Why, Why not, What, How, etc.), depending on a set of models, knowledge, and data resources, as described in computer science and explanation sciences literature~\cite{gilpin2018explaining}, ~\cite{doshi2017towards}, ~\cite{miller2019explanation}. We also modeled attributes required to represent the system interface user attributes, as described in the human-interaction literature~\cite{wang2019designing}, ~\cite{lim2009and}, ~\cite{dey1998cyberdesk}. 
Interested users can also extend EO with additional classes as they deem necessary, by referring to the descriptions of the EO model available in this paper and on our resource website. Additionally, we update the EO model often, ensuring that it is current. In terms of adding more granular and indirectly connected attributes to explanations, we are also investigating how to represent system attributes, such as error traces, metrics and method parameters, and to link such details in user-centered explanation types that would provide value to a large range of domain and non-domain users. 

\ul{\textit{Discussion of Results:}}
We have evaluated the capabilities of the EO in assisting system designers, our intended users, both from a task perspective of the planning process in supporting explanations in their use cases and addressing domain-specific questions that might arise around their use cases. Hence, we use the task-based and application-based evaluation techniques presented in Muhammad et al.'s well-curated taxonomy of ontology evaluation methods{~\cite{amith2018assessing}}. We crafted these competency questions borrowing from our expertise in the explainable AI domain, i.e., from our previous experiences of interactions with end-users where they have alluded to specific needs for explainability{~\cite{chari2020explanation}}, {~\cite{dan2020designing}}, {~\cite{chari2021leveraging}}, and from literature reviews of 
the kinds of questions that users want addressed from explanations{~\cite{liao2020questioning}} and what explanation taxonomies cover{~\cite{arya2019one}}, {~\cite{arya2022ai}}. We also walked a small expert panel of two system designers in our lab through our evaluation approach and presented them with probing questions like do the competency questions serve their needs when they are trying to use the EO and what other questions they would like to see addressed. They made some suggestions around rewording some task-based questions {\ref{tab:competencyquestion}} to be more clear and expressed interest in seeing the domain capabilities of the EO (Tab. {\ref{tab:competencyquestionsusecase}}). We have made changes to the evaluation to reflect the expert panel suggestions. While our evaluation was not designed to be exhaustive, it is representative of capabilities that the EO can enable around making explanations composable from its dependencies and allowing explanations to be probed to support further user-driven questions around them.

Further, the values for different metrics, including query length, property restriction, and whether or not additional mechanisms like inference are required and filter statements need to be applied, that are reported against queries for each competency question in Tab. {\ref{tab:competencyquestion}} and Tab. {\ref{tab:competencyquestionsusecase}}, reflect the design choices that we have made in the EO. To elaborate, queries in the task-based evaluation have longer query lengths since they access property restrictions defined against explanation types supported in the EO to answer questions around the components of these explanation types. On the other hand, since the EO provides capabilities of explanations to be classified into explanation types, queries around domain-specific content of inferred explanations in the exemplar use case KGs have shorter query lengths. While interested system designers could use the queries we have provided in this paper (Sec. {\ref{sec:evaluation}}) as a reference for other queries they might have around the EO or want addressed by the EO, we also have an active developer community that can assist in crafting these queries. We have a call for participation on our website for interested users\footnote{Competency question support: {\url{https://tetherless-world.github.io/explanation-ontology/competencyquestions/#call}}}.

\ul{\textit{Future Outlook:}} Overall, the EO can be thought of as a semantic representation that allows for easy slot-filling of user-centered explanation types in terms familiar to most system designers. In the future, we hope to develop a natural-language processing method that would interface with the explanation slots in the EO's use case KGs to build natural-language explanations. We continue to provide and update our open-source documentation for using the EO model to support user-centered explanation types in use cases that span various domains. The EO is a solution to combine data, knowledge and model-capabilities to compose user-centered explanations that can address a wide range of user questions and provide multiple views 
and thus support human reasoning of AI outputs~\cite{lakkaraju2022rethinking}, ~\cite{dan2020designing} (as illustrated in the example in Fig. {\ref{fig:eoexplanations}}).

\section{Conclusion} \label{sec:conclusion}
We have presented a significantly expanded explanation ontology that can  serve as a resource for composing explanations from contributing components of the system, interface, and user- attributes. We have modeled the mid-level ontology to be used as a cross-domain resource to represent user-centered explanations in various use cases. In addition, within the ontology, we model fifteen literature-derived, user-centered explanation types and define equivalent class restrictions against these types that allow for explanations to be classified into these patterns. In this paper, we have provided guidance for a system designer, our intended user, to apply our ontology in their use cases. This guidance includes descriptions of five open-source use cases that use our ontology, and answers to competency questions that demonstrate the EO's task and application-based capabilities. We aim for the competency questions to serve as a means for system designers to familiarize themselves with the capabilities of our ontology, to support explanations and understand what types of content-specific questions can be asked around explanations in their use cases. Finally, we hope for this open-sourced ontology to serve as a resource to represent user-centered explanations in various use cases and allow for these explanations to be supported by a broad range of AI methods and knowledge sources while accounting for user requirements.
 
\chapter{CONTEXTUALIZING MODEL EXPLANATIONS VIA A \\ KNOWLEDGE-AUGMENTED QUESTION-ANSWERING METHOD} \label{chapt:qa_contextualization} \blfootnote{This chapter previously appeared as: S. Chari, P. Acharya, D.M. Gruen, O. Zhang, E.K. Eyigoz, M. Ghalwash, O. Seneviratne, F.S. Saiz, P. Meyer, P. Chakraborty, D.L. McGuinnesss, ``Informing clinical assessment by contextualizing post-hoc explanations of risk prediction models in type-2 diabetes,'' \textit{Artif. Intell. in Med.} J., vol. 137, Mar. 2023, Art. no. 102498, doi: 10.1016/j.artmed.2023.102498.}
\section{Overview}
In recent years, efforts to describe and formalize  explanations~\cite{chari2019making},~\cite{chari2020explanation} have identified various dimensions and types for it. Specifically, `contextual explanations'~\cite{challener2019proliferation},~\cite{chari2020explanation} hold great promise to satisfy clinical needs and can improve the adoption of AI methods among clinical workflows. Risk prediction is one of the most important tasks in clinical decision making, and an increasingly important in view of the move toward personalized medicine.~\cite{videha2021adoption},~\cite{banning2008review}. To interpret risk scores, clinicians often consult evidence from different levels of the scientific pyramid~\cite{rosner2012evidence} to lookup associations that might impact the patient's treatment or future trajectory. For example, questions like those in Tab. \ref{tab:samplecontextquestions}, are often asked by clinicians when they are trying to understand or use AI model predictions in their practice. Additional contextual information, such as answers to these questions, can help clinicians interpret and trust predictions to take actions. 
However, current work in risk prediction has often narrowly focused on improving model's accuracy, ignoring the aforementioned needs.
Interestingly, several researchers have posited contextual explanations~\cite{lakkaraju2022rethinking},~\cite{framling2020decision},~\cite{chari2020explanation}, that go beyond post-hoc model explanations to frame the predictions in the context of the applied setting and decisions being made. However, the feasibility of extracting such contextual explanations and the added benefit in an end-to-end setting of clinical relevance has not been studied and forms the focus of this paper. Specifically, we consider how to derive and support contextual explanations from authoritative domain knowledge sources, not already considered by prediction models, that clinicians would typically use to reason through decisions presented to them when dealing with recommendations from learning health systems.

\begin{definition}[Contextual Explanation]
  Explanations that contain context, are often explicit information ~\cite{lieberman2000out} to characterize the situation of (an) entity(ies), wherein ``an entity is a person, place, or object that is considered relevant to the interaction between a user and an application''~\cite{dey1998cyberdesk}. 
\end{definition}

\begin{table}[!htbp]
\centering
\caption{Questions that could be asked in clinical use cases around model explanations / predictions, and which can benefit from contextual explanations in the context of use.} 
\label{tab:samplecontextquestions}
\begin{adjustbox}{max width=1.0\textwidth,center}
   \begin{tabular}{l}
    \toprule
    Sample Question \\
    \midrule
     \multicolumn{1}{m{\columnwidth}}
     {What treatment can be suggested for this patient who has an increased risk of cardiovascular disease?} \\ \midrule
    \multicolumn{1}{m{\columnwidth}}
    {What other conditions does this patient have that might impact this decision?} \\ \midrule
    \multicolumn{1}{m{\columnwidth}}
    {What was the patient's A1C value when this prediction was made?} \\ \midrule
    \multicolumn{1}{m{\columnwidth}}
    {Why are you telling me that this risk is important?} \\
    \bottomrule
\end{tabular}
\end{adjustbox}
\end{table}

\subsection{Use Case} \label{sec:use-case}
\begin{figure*}[!htbp]
    \centering
    \includegraphics[width=\linewidth]{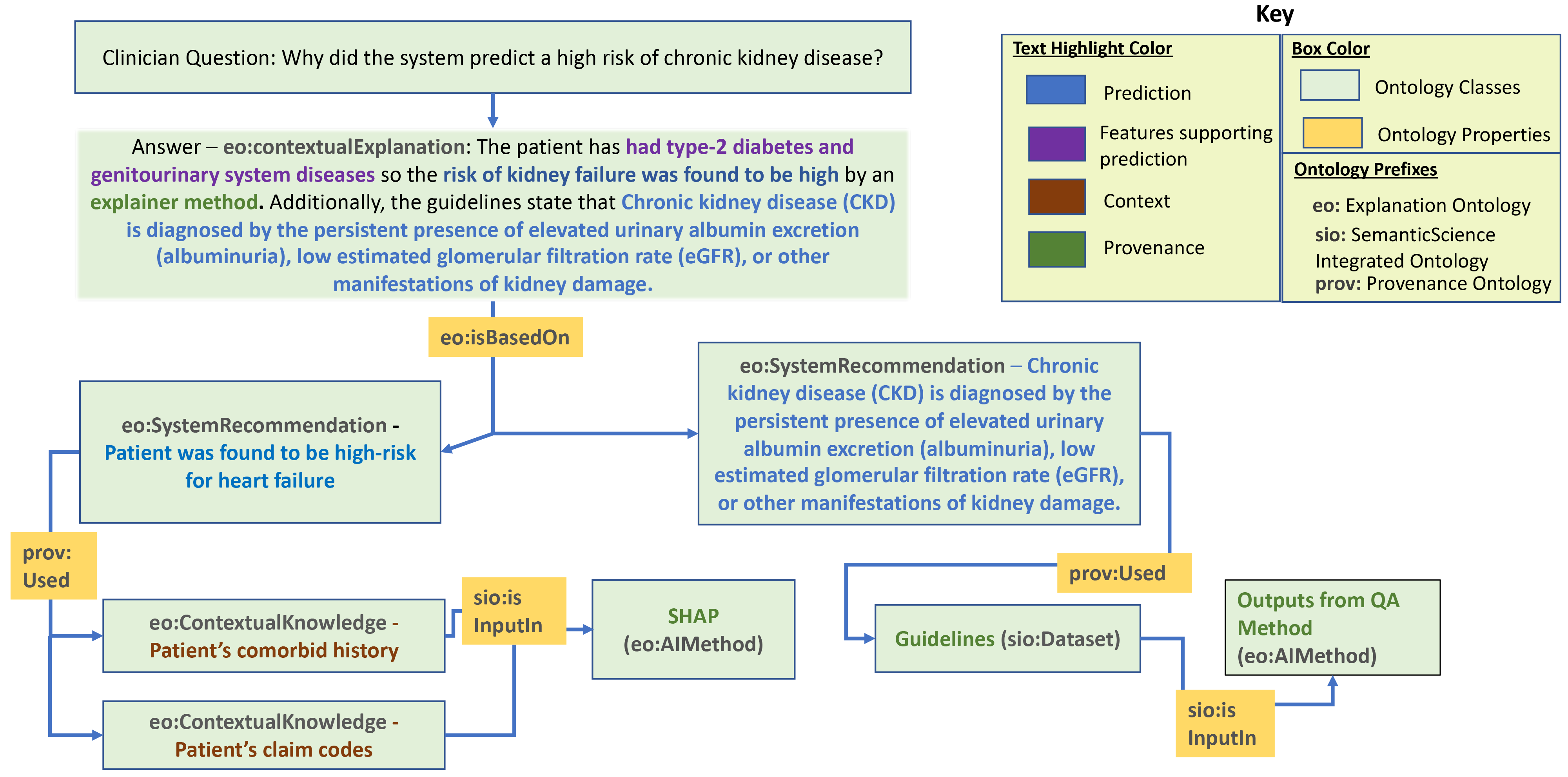}
        \caption{Trace of dependencies of a contextual explanation example on system outputs from post-hoc explainer models and question-answering methods. In this example, we have used ontology classes and properties to annotate the data, but in this paper we mainly focus on a multi-method approach to support such contextual explanations. Reproduced from: S. Chari, P. Acharya, D.M. Gruen, O. Zhang, E.K. Eyigoz, M. Ghalwash, O. Seneviratne, F.S. Saiz, P. Meyer, P. Chakraborty, D.L. McGuinnesss, ``Informing clinical assessment by contextualizing post-hoc explanations of risk prediction models in type-2 diabetes,'' \textit{Artif. Intell. in Med.} J., vol. 137, Mar. 2023, Art. no. 102498, doi: 10.1016/j.artmed.2023.102498.}
        \label{fig:usecaseexample}
\end{figure*}

AI models promise to help clinicians by providing tools for improved decision making. Risk prediction of patients is one of the key steps in a clinical decision scenario and models for such use cases can be consumed by a broad spectrum of clinicians with differing roles and experience, e.g. a specialist vs a primary-care physician. Depending on their roles, the needs, and thereby the desired functionality from a risk prediction contextualization standpoint, can be different. We worked closely with a clinical expert to
to understand the clinical use-case and determine the context of AI tools. 
Crucially, we aimed to form an understanding of the unmet needs and identify relevant contexts that can benefit clinicans. 
We can further motivate this via Fig.~\ref{fig:usecaseexample} which shows an example question posited by a clinician while consuming the outputs of a risk prediction model. 
In this case, the relevant response can be identified via contextual explanations~\cite{chari2020explanation} that is generated from multiple sources and via multiple extracted contexts

Our aim, was to thus scope the relevant contexts that will be the focus of our study. 
We followed a sequence of 
user-centric research principles~\cite{liao2020questioning}, to
(1) define the scope of our tool's capabilities, (2) identify the end-user/target persona who would most benefit from our tool, and (3) scope the most relevant contexts. 
Through our interviews, we identified primary-care physicians (PCP), especially those with lesser years of experience, to be the persona who may most benefit from such contextualizations. We describe the covered context types in Section~\ref{sec:contextualizations}. 
Furthermore, to study our stated problem in a real-world setup, we identified the problem of 
risk prediction of {\ckd} among new {\dm} patients at their first diagnosis.

This is motivated by the fact that diabetes is one of the top five chronic diseases affecting the adult population in the US~\cite{cdcT2D}. 
Diabetes management involves monitoring for and treating 
related comorbid conditions. 
Effective and timely prediction of such conditions can lead to an overall improvement in the quality of care and thus evaluating the impact of AI models in improving clinical decision workflow can have tremendous real-world impact.
Especially, we focus on {\ckd}, a commonly occurring micro-vascular complication of {\dm} and one of the leading causes of death in the US~\cite{cdcCKD}, with an estimated $37$ million cases in the US (who are mostly undiagnosed) and cost medicare in 2018 $81.1B$, and end stage renal disease an additional $36.6B$. Typically, actions to prevent onset of {\ckd} among {\dm} patients revolve around proper disease control,  including close disease monitoring, proper treatment adherence, and patient education. Incorporating accurate risk prediction of {\ckd} in the clinical workflow can lead to more timely actions, potentially delaying the onset of {\ckd}, and in some cases, preventing its progression.
While such predictions could be of use along various time-points of the patients' {\dm} prognosis, in this paper, we predict the risk of developing {\ckd} within 360 days of {\dm} onset. 
Under this use case, we explore strategies to provide context around interventions for particular patients, and explain their {\dm} state and individual risk factors.

\section{Background} \label{sec:background}
\subsection{Selected Contextual Entities of Interest} \label{sec:contextualizations}
To support the goal of providing contextual, clinically relevant, user-centered explanations,
in consultation with a medical expert on our team, we identified three entities of interest to provide contextual explanations about predicting the risk of {\ckd} among {\dm} patients. Fig.~\ref{fig:usecaseexample} shows an example of contextual explanation that can answer a clinician's question around patient management. It can be seen that such explanations are usually composed of multiple entities and from multiple sources. In general, we identified and subsequently focused on extracting the following contexts:

\par \noindent $\bullet$ \ul{Contextualizing the patient} by connecting their clinical history and indicators to treatments typically recommended for such patients, according to CPGs.
\par \noindent $\bullet$ \ul{Contextualizing risk predictions for the patient} in terms of the prediction's impact on decisions, based on general norms of practice concerning potential complications, as evident from guidelines and other domain knowledge, including medical ontologies.
\par \noindent $\bullet$  \ul{Contextualizing details of algorithmic, post-hoc explanations}, such as connecting features that were the most important to other information based on their potential medical significance, such as through connections to physiological pathways and CPGs.

\begin{figure*}[!htbp]
    \centering
    \includegraphics[width=\linewidth]{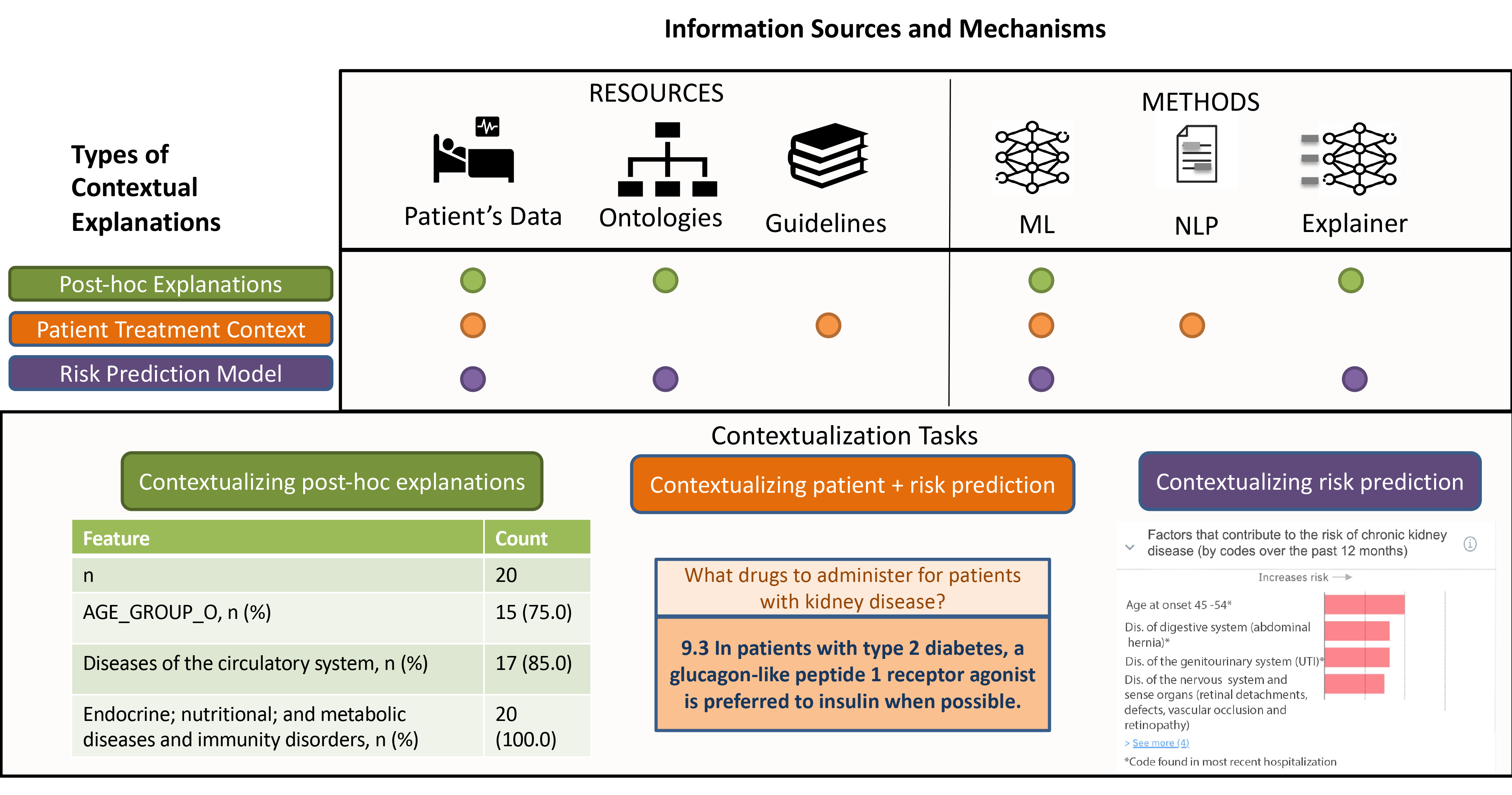}
        \caption{Different types of contextualizations supported by methods, that help provide additional context around patients, their risk predictions and features contributing to risk, via connections to different knowledge sources including patient data, medical ontologies and guidelines. Reproduced from: S. Chari, P. Acharya, D.M. Gruen, O. Zhang, E.K. Eyigoz, M. Ghalwash, O. Seneviratne, F.S. Saiz, P. Meyer, P. Chakraborty, D.L. McGuinnesss, ``Informing clinical assessment by contextualizing post-hoc explanations of risk prediction models in type-2 diabetes,'' \textit{Artif. Intell. in Med.} J., vol. 137, Mar. 2023, Art. no. 102498, doi: 10.1016/j.artmed.2023.102498.}
        \label{fig:contextualizations}
\end{figure*}

In Fig. \ref{fig:contextualizations}, some examples of contexts that we support around the three entities of interest can be seen in the risk prediction setting. Also seen in the figure are the pathways in which responses providing context could borrow from different domain knowledge sources and methods. For example, the answer to the question, ``What drugs to administer for chronic kidney disease?'' provides context around the patient and risk prediction, borrows both from guidelines and patient data, and is supported by the risk prediction and natural language extraction modules which we describe in Sec. \ref{sec:methods}.  

\subsection{Data Sources}
\label{sec:data-sources}
To conduct our real-world study, we focus on two specific sources of data as described below.
\subsubsection{Patient Data} \label{sec:patient-data}
We conduct our analysis on and retrieve patient data from the claims sub-component of the Limited IBM MarketScan Explorys Claims-EMR Data Set (LCED)~\cite{marketscan}, covering both administrative claims and EHR data of over $5$ million commercially insured patients between 2013 and 2017.
Medical diagnoses are encoded 
using International Classification of Diseases (ICD) codes. 
We selected only those {\dm} patients (with ICD9 codes 250.*0, 250.*2, 362.0, and ICD10 code E11)  that satisfied the following criteria as our cohort.
 Only {\dm} patients with the following criteria are included:
\begin{itemize}
  \item have had two or more visits with {\dm} diagnosis,
  \item  were enrolled continuously for $12$ months prior to 
{\dm} diagnosis,
  \item  number of visits for {\dm}  is greater than those for other forms of diabetes such as \texttt{T1D}, and
  \item  age at the initial {\dm} diagnosis is between 19-64 years.
\end{itemize}

Among {\dm} patients, we use the first diagnosis of chronic kidney disease ({\ckd}) (ICD10 N18 or ICD9 585.*, 403.*) after the initial diagnosis of {\dm} as the outcome to predict. 
At the time of the first {\dm} diagnosis, we predict the risk that the patient develops {\ckd} within $1$ year using Clinical Classifications Software (CCS) codes, age group and sex as features for the predictive model.

\subsubsection{Clinical Practice Guidelines} \label{sec:cpg}
\begin{figure*}[!htbp]
    \centering
    \includegraphics[width=\linewidth]{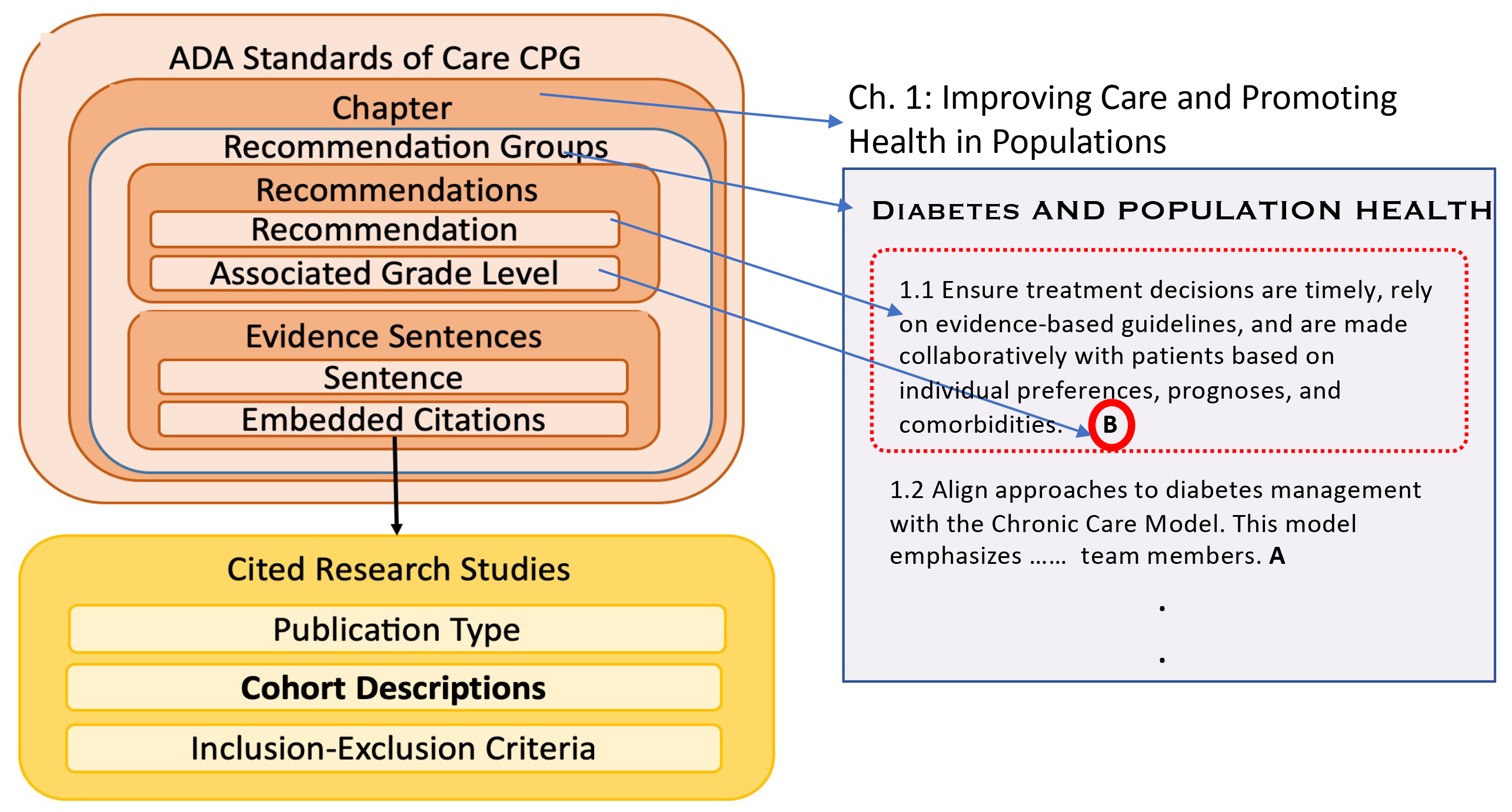}
        \caption{ADA Guideline structure wherein recommendations are associated with a recommendation group and evidence grade which if high enough is supported by citations. Reproduced from: S. Chari, P. Acharya, D.M. Gruen, O. Zhang, E.K. Eyigoz, M. Ghalwash, O. Seneviratne, F.S. Saiz, P. Meyer, P. Chakraborty, D.L. McGuinnesss, ``Informing clinical assessment by contextualizing post-hoc explanations of risk prediction models in type-2 diabetes,'' \textit{Artif. Intell. in Med.} J., vol. 137, Mar. 2023, Art. no. 102498, doi: 10.1016/j.artmed.2023.102498.}
        \label{fig:qaguidelinestructure}
\end{figure*}

Clinical Practice Guidelines are position statements published by a board of experts in different disease areas~\cite{murad2017clinical}. These guidelines are updated often, latest summaries of updated evidence in the disease areas, and follow the highest standards of evidence appraisal (e.g., Grading of Recommendations, Assessment, Development and Evaluations (GRADE) evidence schemes~\footnote{https://www.gradeworkinggroup.org}). Further, the guidelines are written to be comprehensive sources covering different aspects of treatment, management, and assessment of the disease and are often regarded as first-line lookup sources for clinicians and primary care physicians~\cite{graham2011trustworthy, murad2017clinical}. Given their comprehensive and updated nature, CPGs provide a great resource for providing clinical contexts in various clinical settings. We utilize the 2021 edition of the American Diabetes Association (ADA) Standards of Care guidelines (Fig. \ref{fig:qaguidelinestructure}) for our experiments.

\section{Methods} \label{sec:methods}
\begin{figure}[hbt!]
\centering
\includegraphics[width=1.0\linewidth]{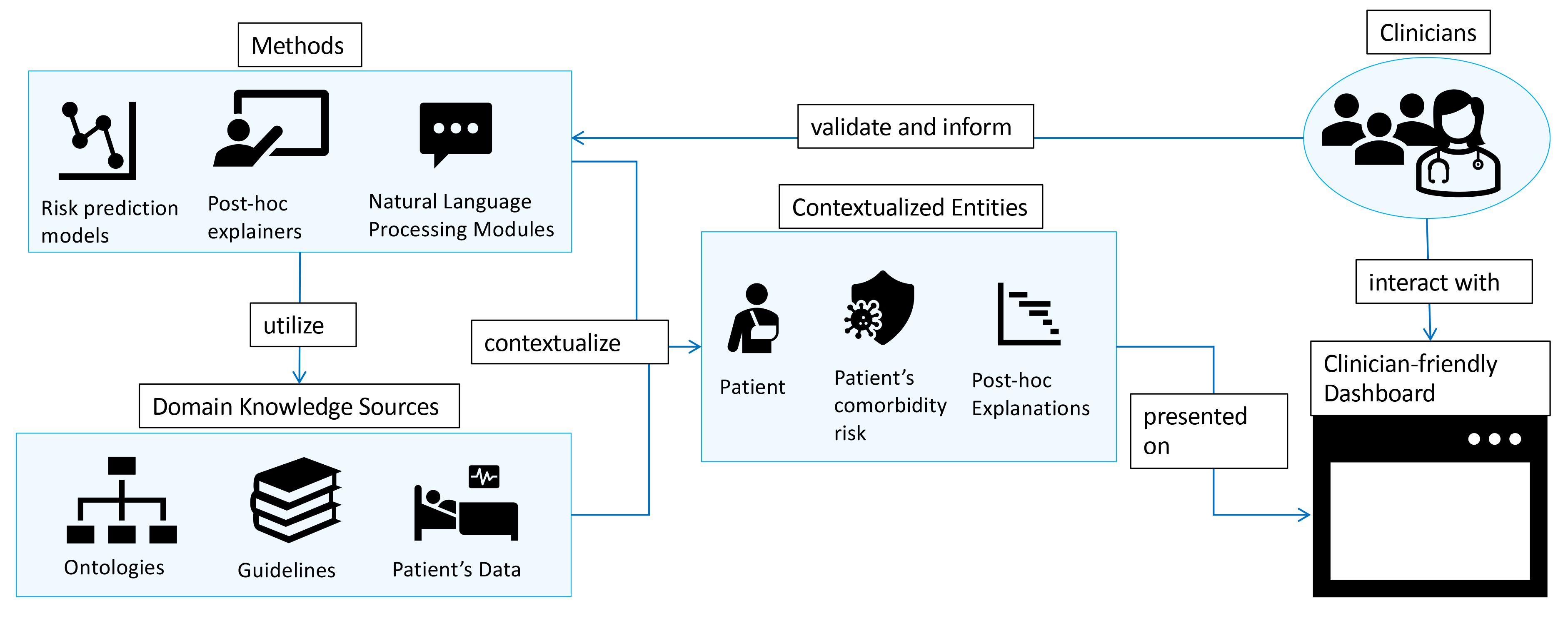}
\caption{A framework built to contextualize entities of interest from authoritative domain knowledge sources in a comorbodity risk prediction setting. Reproduced from: S. Chari, P. Acharya, D.M. Gruen, O. Zhang, E.K. Eyigoz, M. Ghalwash, O. Seneviratne, F.S. Saiz, P. Meyer, P. Chakraborty, D.L. McGuinnesss, ``Informing clinical assessment by contextualizing post-hoc explanations of risk prediction models in type-2 diabetes,'' \textit{Artif. Intell. in Med.} J., vol. 137, Mar. 2023, Art. no. 102498, doi: 10.1016/j.artmed.2023.102498.}
\label{fig:overallMethodSummary}  
\end{figure}

To study the problem of risk prediction of {\ckd} among {\dm} patients, we created an end-to-end AI enabled system to provide relevant contextual explanations from authoritative literature sources such as guidelines. Fig. \ref{fig:overallMethodSummary} shows a conceptual overview of the components of this system. In general, to extract contextual explanations around our three identified entities of interest, we used a number of components including risk prediction models, post-hoc explanation models, and our multi-method, question-answering approach to provide context. 
Crucially, to analyze the importance of the supported contextual explanations, we prototyped and implemented a risk prediction dashboard (Sec. \ref{sec:dashboard} to display the extracted contextual explanations from our QA pipeline (Sec. \ref{sec:qa}) alongside patient predictions and data, and ran qualitative analysis with an expert panel of clinicians using this dashboard. 
%
In this section, we provide high-level details of some of the key components involved in the process.

\subsection{Risk Prediction Models}\label{sec:risk_model}
In the first step of our pipeline, we build risk prediction models from the constructed cohort and the aforementioned use-case (Sec.~\ref{sec:use-case}). In particular, we train a suite of machine learning models (ML), including both classical and deep-learning models, and select the best performing one based on the highest predictive accuracy and other appropriate metrics for the use case, such as favoring models with a higher recall. We used DPM 360~\cite{dpm360}, an open-source, reusable, disease progression model training package, we compared a suite of classification models on the patients' demographic and diagnosis history to predict future complications. In this contribution, we only used the demographic and diagnostic features to model risk. Furthermore, to handle the temporal features, for some of our models, such as Logistic Regression (LR) and Multi-layer perceptron (MLP), we used temporally aggregated features (summation). 
We also compared two state-of-the art Recurrent Neural Networks (RNN) where temporal history can be handled in a more natural manner, viz., Long-Short Term Memory (LSTM) and Gated Recurrent Units (GRU). All model implementations are available via DPM360 including classical ML models (backed by scikit-learn) and deep learning models (custom built for DPM).
In this contribution, we split the data according to a train-validation-test split (70-10-20). Using the best performing models on the validation set, we present our results on the hold-out test set.
 Since the data is imbalanced, we selected the models based on the best AUC-ROC and AUC-PRC from the validation set. We also evaluate the models based on precision, recall, and brier score~\cite{rufibach2010use}. Deep learning networks are known to be under-calibrated and the last metric measures how well the model is calibrated, i.e., it measures the probabilistic interpretation of risk prediction. In other words, if a model predicts a $0.7$ risk for a patient, brier score measures how well that translates to a $70\%$ chance of the patient developing the complication.
The hyper-parameters for the deep-learning models were selected using a grid search strategy varying batch sizes  $\lbrace 8, 16, 32, 128 \rbrace$, number of layers $\lbrace 1, 2, 3 \rbrace$, and dropout $\lbrace 0.0, 0.1, 0.2 \rbrace$ along with standard initialization and using ADAM as the optimizer of choice.

\subsection{Post-hoc Explainer Models}

While some of the classical algorithms considered in Section~\ref{sec:risk_model} are inherently interpretable with easy access to the features deemed important for the model (such as LR), several of the deep learning models are black-box models. 
To extract feature importances from such models, we used post-hoc explainers which have been found to be favored by clinicians in past studies~\cite{tonekaboni2019clinicians}.
In particular, we used the well accepted SHAP algorithm~\cite{lundberg2018explainable} to find feature importance~\footnote{In this contribution, our primary goal is study the importance of contextual explanations and thus we chose SHAP as a well-known SOTA post-hoc explainer. However, we caution the readers about known criticism of SHAP, and in general explainability methods, that are still active area of research without a common consensus}. The algorithm uses game-theoretic principles to identify importance of features by ascertaining the dip in performance of the model with and without access to the feature at the personalized level. 
Such personalized feature importance is key so that our overall risk prediction presentations are more actionable for the clinicians by allowing them to focus on the particular attributes of the patients that are driving their risk.

Typically, clinician time is costly and hard to obtain. Thus to conduct the expert panel sessions and let them focus on some of the most `interesting' patients, we apply Protodash~\cite{gurumoorthy2019efficient} to select a subset of patients. 
Protodash is a post-hoc sample selection method used to obtain a set of prototypical or representative patients from the high risk category that naturally spans the varied set of patient characteristics for the selected sub-group. This also allows the clinician to build trust in using the AI models by inspecting the different patient modalities of the dataset without having to inspect the entire dataset.
\subsection{Information Extraction}
\begin{figure}[hbt!]
\centering
\includegraphics[width=1.0\linewidth]{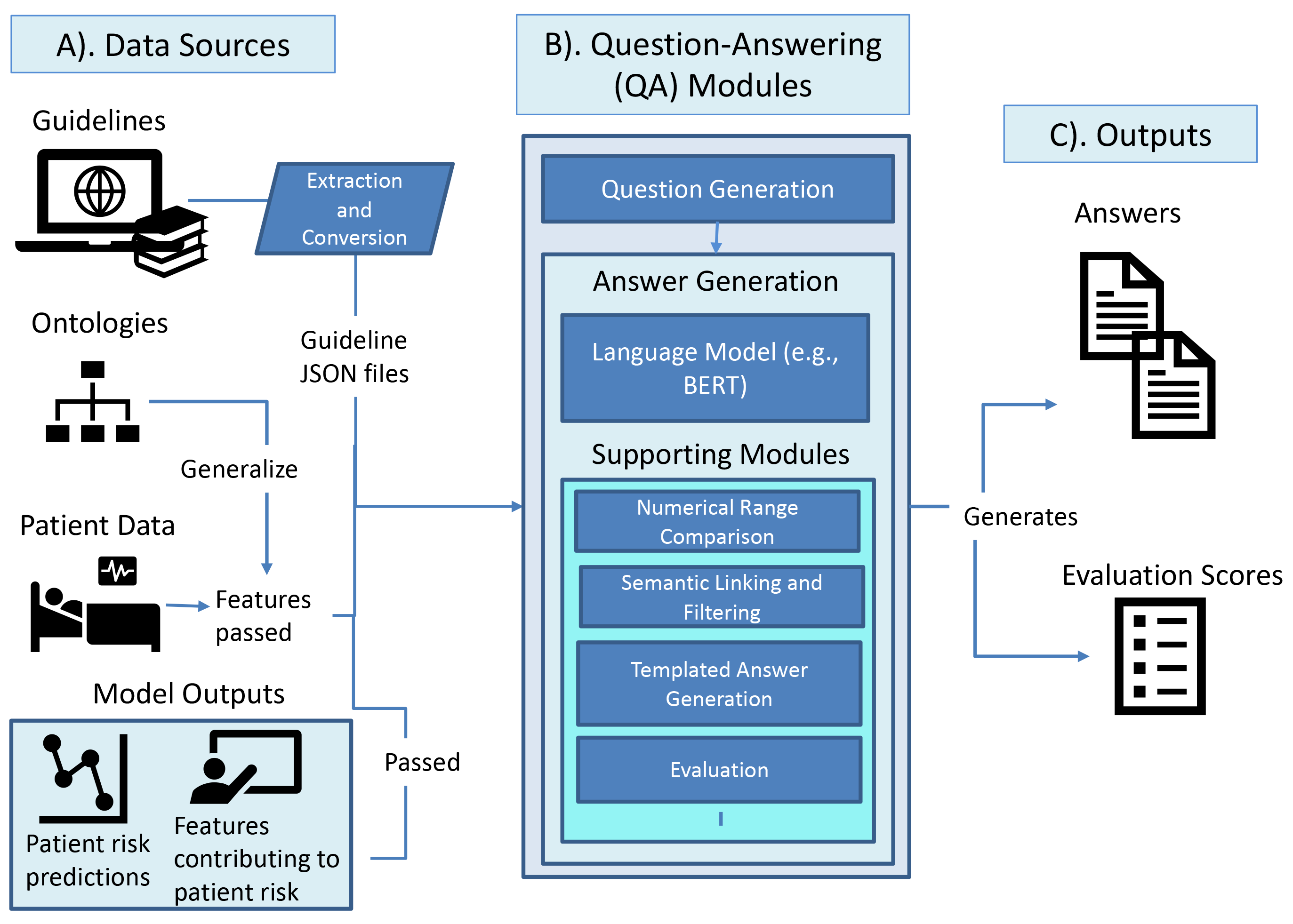}
\caption{QA Pipeline developed to address patient-related questions via clinical practice guidelines (CPG). Seen are A) data sources used within the QA pipeline, B) the various methods implemented in the pipeline and finally C) the overall outputs that include the extracted answers from the CPG and the  evaluation scores. Reproduced from: S. Chari, P. Acharya, D.M. Gruen, O. Zhang, E.K. Eyigoz, M. Ghalwash, O. Seneviratne, F.S. Saiz, P. Meyer, P. Chakraborty, D.L. McGuinnesss, ``Informing clinical assessment by contextualizing post-hoc explanations of risk prediction models in type-2 diabetes,'' \textit{Artif. Intell. in Med.} J., vol. 137, Mar. 2023, Art. no. 102498, doi: 10.1016/j.artmed.2023.102498.}
\label{fig:guidelineqa}  
\end{figure}

We support the extraction of context from three domain sources in our QA approach, including patient data, medical ontologies like Clinical Classification Software (CCS) codes~\footnote{\url{https://www.hcup-us.ahrq.gov/toolssoftware/ccs10/ccs10.jsp}} and medical guidelines from ADA Standards of Care 2021 (as introduced in Sec. \ref{sec:background}). We query patient data from Limited Claims Explorys Dataset (LCED) claim records (see Sec. \ref{sec:use-case}) on-demand, either when we need to create questions based on patient parameters or when we need to include these patient values in answers to questions about the patient.  We extract content from the HTML or web version of the `Standards of Medical Care in Diabetes'~\cite{care2021standards} guidelines, published by the American Diabetes Association (ADA)~\footnote{\url{https://care.diabetesjournals.org/content/44/Supplement_1}} using a Python library, BeautifulSoup~\cite{richardson2007beautiful}. We also query patient risk predictions and feature importances using a unique identifier, the patient ID. Some of the extracted contexts are used for generating questions such as patient data, risk predictions and feature importances, and others are used to query against such as the extracted guidelines.
\subsection{Question-Answering (QA) Modules} \label{sec:qa}

\begin{table*}[!htbp]
 \centering
  \caption{Question types currently supported by our QA module, we also indicate the data sources used to address questions of the type. } 
  \label{tab:questiontypes}
  \small
  \begin{tabular}{lllc}
    \toprule
    {} & Question Type & \multicolumn{1}{m{3cm}}{Contextualized Entity} & \multicolumn{1}{m{5cm}}{Domain Knowledge Source} \\
    \midrule
    1 & \multicolumn{1}{m{4cm}}{Patient's {\dm} summary} & \multicolumn{1}{m{3cm}}{Patient} & \multicolumn{1}{m{5cm}}{Patient data}\\
    2 & \multicolumn{1}{m{4cm}}{Patient's risk summary} & \multicolumn{1}{m{3cm}}{Risk Prediction} & \multicolumn{1}{m{5cm}}{Risk Prediction and population data}\\
    3 & \multicolumn{1}{m{4cm}}{Features contributing to patient's {\ckd} risk} & \multicolumn{1}{m{3cm}}{Post-hoc Explanation} & \multicolumn{1}{m{5cm}}{Feature importances and ADA guidelines}\\
    4 & \multicolumn{1}{m{4cm}}{Patient's medication list} & \multicolumn{1}{m{3cm}}{Patient and Risk Prediction} & \multicolumn{1}{m{5cm}}{Patient Data and guidelines}\\
    5 & \multicolumn{1}{m{4cm}}{Patient's lab values} & \multicolumn{1}{m{3cm}}{Patient} & \multicolumn{1}{m{5cm}}{Patient Data and guidelines}\\
  \bottomrule
\end{tabular}
\end{table*}

Here we describe part B). of our QA architecture (Fig. \ref{fig:guidelineqa}), including the question and answer generation modules and their supporting submodules. In our QA setup we leverage SOTA LLMs and introduce knowledge augmentations to improve their performance on the ADA 2021 medical guidelines. Additionally, we introduce sub-modules to enhance the LLMs' capabilities to address question types 3 - 5 from Tab. \ref{tab:questiontypes}, i.e., diagnosis codes, drugs and clinical indicators, which are run against our extracted guideline content. Below are the submodules in our QA setup:
\par \textit{Question Generation:} The \textit{question generation module} almost always creates templated questions using Python's native support for String Templates,~\footnote{\url{https://docs.python.org/3/library/string.html}} and does so based on patient data, more specifically from the patient's diagnoses codes, lab values, and medication list. We also support the creation of two standard, non-variant questions for each patient, i.e., whose values don't change from patient data, that can help clinicians easily interpret their predicted risk (question type 1) and their {\dm} state (question type 2). Moreover, as can be seen from Tab. \ref{tab:questiontypes}, each of the question types that we support on a per patient basis is populated from different data sources. Hence, we have developed different answering methods for each, including simple lookups and knowledge augmented language model capabilities, including combinations of either a LM + value range comparison or LM + semantic filtering. We provide examples of questions and answers for each question type in Tab. \ref{tab:appendix:questiontypeexamples} and Tab. \ref{tab:appendix:questiontypeexamples2}.
\par \textit{Answer Generation:} For questions types 1 and 2 from our supported question types, whose context does not depend on guidelines as shown in Tab. \ref{tab:questiontypes}, we query patient data and feature importances and use a similar templating approach for question generation to populate answer templates with the retrieved query results. For other question types 3 - 5 that are answered by guideline content, we pass them through our LLM and knowledge-augmented LLM setup that we describe next.
\par \textit{Language Models for Generating Answers:} We use a LLM approach in order to find answers to our questions with the unstructured and natural language discussion and recommendation sentences of the ADA 2021 guidelines. We have applied the original Bidirectional Encoder Representations from Transformers (BERT) model or BERT~\cite{devlin2018bert} and other variants of the same retrained on clinical datasets, including SciBERT~\cite{otegi-etal-2020-automatic}, BioBERT~\cite{lee2020biobert}, BioBERT-ASQ~\cite{Yoon2019PretrainedLM} and BioClincalBERT-ADR~\cite{huggingfaceBioClinicalBERT-ADR}. All of the models we utilize are available on the HuggingFace~\cite{wolf2019huggingface} model repository, and we choose BERT models that were made available specifically for clinical question-answering. See below (Tab. \ref{tab:bertmodeldetails} for more details on the corpuses that the BERT models we selected were trained on. 

\begin{longtable}{| m{5cm} | m{10cm} |}
\caption{Training details of BERT-variants of Language Models (LLMs) used in our QA pipeline.} \label{tab:bertmodeldetails} \\ \hline
\textbf{BERT Model} & \textbf{Data Trained On} \\ \hline
BERT~\cite{devlin2018bert} & Stanford Question Answering Dataset (SQUAD) \\ \hline
BioBERT~\cite{lee2020biobert} & English Wikipedia, BooksCorpus, PubMed Abstracts, PMC Full-text articles \\ \hline
SciBERT~\cite{otegi-etal-2020-automatic} & 1.14M papers from Semantic Scholar (18\% - Computer Science, 82\% - Biomedical Domain) \\ \hline
BioBERT-BioASQ~\cite{Yoon2019PretrainedLM} & BioBERT data + BioASQ Challenge Dataset \\ \hline
BioClinicalBERT-ADR~\cite{huggingfaceBioClinicalBERT-ADR} & ADE-Corpus-V2 Dataset: Adverse Drug Reaction Data \\ \hline
\end{longtable}

We built two other submodules to enhance the capabilities of the LLM approach, specifically to address questions with numerical comparisons of question type 5, and to improve the semantic match between the question and the answers returned by a LLM (question type 3 and 4).

\par \textit{\textbf{Numerical Range Comparison}:} LLMs cannot currently determine if a question that has a numerical value comparison, e.g., ``What can be done for patients, whose Hemoglobin A1C $>$ 10?,'' falls in the range of the answer returned. However, clinicians often look for recommendations that match patients' lab values in clinical settings such as ours. Further, within the ADA 2021 guidelines, there are mentions for suggestions based on different ranges for lab values, e.g., a recommendation from the Pharmacological Chapter of these guidelines has a recommendation with the mention of ``when A1C levels ($>10\%$ $[$ 86 mmol/mol $]$." Hence, for question type 3 from Tab. \ref{tab:questiontypes}, we need to determine if the patient's lab values are in the range of the answer returned by the LLM module. To address this requirement of performing numerical range comparison between the question and answer produced by LLM, we leverage syntactic parsing capabilities (similar to ~\cite{chen2021personalized}) to identify numerical phrases in the question and answer and then determine if each numerical phrase from the question is in range of the same in the answer. 

We use Natural Language Toolkit (NLTK) chunking and parsing functionalities to identify noun phrases, comparatives, and numerical mentions within both the question and answer. We then write regular expression (regex) rules to identify the patterns of the positional tags returned by NLTK that can constitute numerical phrases. For each of the numerical phrases, we convert them into a tuple of ``(noun phrase, [upper bound, lower bound]).'' This tuple representation allows us to go through the phrases between the question and answer iteratively, and for those that match on the noun phrase dimension, identify if the ranges are in agreement. With these steps, we can then populate an answer using the Template-based Answer generation module, which says if the answer outputted by LLM is in/out of the range of the question. Hence, in this manner, we enhance the capabilities of BERT LLMs for numerical range comparisons via rule-based syntactic methods, which is also why we consider this step a rule augmentation of LLMs' capabilities.

\par \textit{\textbf{Augmenting Knowledge to LLM}:}  Transformer based LLM approaches like BERT and its variants, work on sequences of words that are often seen together and their surrounding words, but don't leverage the semantics of whether these words are diseases, medications, or biological processes. We found that in the absence of this semantic knowledge, we would often get answers from BERT that don't correlate on a semantic level with the question.  To eliminate such answers, we explored options for a biomedical semantic mapper and zeroed in on the National Library of Medicine (NLLM)'s Metamap tool~\cite{aronson2010overview}. We choose Metamap because of its extensive coverage of biomedical semantic types and its ability to capture entity mentions within the ADA 2021 CPG. Within our pipeline, we have integrated a Python wrapper for Metamap\footnote{PyMetamap: \url{https://github.com/AnthonyMRios/pymetamap}} that can recognize biological entities within the guideline text and their semantic types (e.g., dsyn: disease or syndrome, phsu: pharmacolgic substance, etc. for a complete list of types returned by Metamap see: \footnote{\url{https://lhncbc.nLLM.nih.gov/ii/tools/MetaMap/Docs/SemanticTypes_2018AB.txt}}). Additionally, given this ability to filter based on semantic types, we want to allow additional answers with mentions of related diseases. To provide more broad answers, we use the UMLS Concept Unique Identifier (CUI) codes from the Metamap returned outputs to map to SNOMED-CTdisease codes~\cite{donnelly2006snomed}. From the mapped SNOMED-CT disease codes, we can traverse the SNOMED-CT disease tree to identify how many hops apart question and answer disease codes are and if the answer codes are an ancestor of those in the question. We operate on the idea that answers about the parent disease code apply to children nodes. We use the outputs of these knowledge augmentation modules to both pre-filter and post-sort LLM model answers for question types 3 and 4 from Tab. \ref{tab:questiontypes}. We report the accuracies for answers that use these knowledge augmentations in the results section (Sec. \ref{sec:guidelineqaresults}). 

We use the outputs of these knowledge augmentation modules to both pre-filter and post-sort the LLMs answers. The LLM and LLM + post sorting settings 1, 2 and 5, were run against $410$ passage chunks of guideline text, of average length $267$ tokens, since BERT has a 512 token limit for an answer passage. The LLMs on the pre-filtering settings 3 and 5 were run on passage chunks of variable length, depending on the number of filtered sentences to be passed to the LLM model. In the pre-filtering settings, we varied the values of the features that we were filtering by to understand which feature values improve accuracy. In essence, the pre-filtering settings can be thought of as algorithmic knobs to control the set of answers that the LLM has to process. In contrast, in the post-filtering settings we sorted the LLM's answers by feature values, and here we could control the ordering of answers to be outputted. In the pre-filtering setting 2, we filter the guideline sentences by length of disease overlap with the question. In pre-filtering setting 4, we have more possibilities in the feature column because the number of hops in the SNOMED-CT KG between a question and answer can range between a continuous range of integer values. We report if restricting the number of hops to allow for more general yet precise answers improves accuracy. Similarly, in the post-sorting settings, 2 and 4, we use the feature values from the knowledge augmentation modules in addition to the LLM's own confidence scores to rank answers. Specifically, in setting 2, we sort the LLM's answerset on variations to a combination of length of disease overlap between question and answer Metamap phrases and the LLM confidence scores. In setting 5, we sort the LLMs answerset based on variations to a combination of sum of hops between question and answer SNOMED-CT disease codes, number of SNOMED-CT ancestors in the answer and LLM confidence scores. 

Below in Tab. \ref{tab:appendix:questiontypeexamples} and \ref{tab:appendix:questiontypeexamples2}, we present sample questions and answers for each question type to provide examples of questions and extracted answers supported by our QA approach. We intentionally don't show patient values in these examples to be compliant with HIPAA restrictions.

\begin{table*}[!htbp]
  \centering
  \caption{Sample questions and answers for each question type supported by our question-answering approach. Answers such as these serve as contextual explanations that provide information to interpret risk predictions better. We don't provide patient values here due to HIPAA restrictions.} 
  \small
  \label{tab:appendix:questiontypeexamples}
  \begin{tabular}{lllc}
    \toprule
   \multicolumn{1}{m{3cm}}{Question Type} & \multicolumn{1}{m{4cm}}{Sample Question} & \multicolumn{1}{m{6cm}}{Answer} \\
    \midrule
     \multicolumn{1}{m{3cm}}{1. Patient's {\dm} summary} &  \multicolumn{1}{m{4cm}}{What is the patient's A1C value? What are their most frequent diagnoses codes?} & \multicolumn{1}{m{6cm}}{Patient's A1C is A. Their most frequent diagnosis codes are essential hypertension, septicemia, etc.}\\
     \midrule
    \multicolumn{1}{m{3cm}}{2. Patient's risk summary} & \multicolumn{1}{m{4cm}}{How does the predicted risk of the patient compare against the population?} & \multicolumn{1}{m{6cm}}{The predicted risk of chronic kidney disease the patient is X \%. The population averages for the same condition are as follows: For Medicare patients: Y \%  For patients with Charlson Comorbidity Index (CCI) score of 3 : Z \%}\\
    \midrule
    \multicolumn{1}{m{3cm}}{3. Features contributing to patient's {\ckd} risk} & \multicolumn{1}{m{4cm}}{What can be done for Essential Hypertension?} & \multicolumn{1}{m{6cm}}{10.3 For patients with diabetes and hypertension, blood pressure targets should be individualized through a shared decision-making process that addresses cardiovascular risk, potential adverse effects of antihypertensive medications, and patient preferences. C}\\
  \bottomrule
\end{tabular}
\end{table*}

\begin{table*}[!htbp]
  \centering
  \caption{Sample questions and answers for each question type supported by our question-answering approach. Answers such as these serve as contextual explanations that provide information to interpret risk predictions better. We don't provide patient values here due to HIPAA restrictions.} 
  \label{tab:appendix:questiontypeexamples2}
  \small
  \begin{tabular}{lll}
    \toprule
   \multicolumn{1}{m{3cm}}{Question Type} & \multicolumn{1}{m{4cm}}{Sample Question} & \multicolumn{1}{m{6cm}}{Answer} \\
    \midrule
    \multicolumn{1}{m{3cm}}{4. Patient's lab values} & \multicolumn{1}{m{4cm}}{What should be done for this patient, whose A1C levels are greater than 10 ?} & \multicolumn{1}{m{6cm}}{The early introduction of insulin should be considered if there is evidence of ongoing catabolism (weight loss), if symptoms of hyperglycemia are present, or when A1C levels are greater than 10\% [86 mmol/mol] or blood glucose levels greater than or equal to 300 mg/dL [16.7 mmol/L] are very high.}\\
    \midrule
    \multicolumn{1}{m{3cm}}{5. Patient's medication list} & \multicolumn{1}{m{4cm}}{What do the guidelines state about the GLP-1 RA drug the patient is taking?} & \multicolumn{1}{m{6cm}}{ Meta-analyses of the trials reported to date suggest that GLP-1 receptor agonists and SGLT2 inhibitors reduce risk of atherosclerotic major adverse cardiovascular events to a comparable degree in patients with type 2 diabetes and established ASCVD (185).}\\
  \bottomrule
\end{tabular}
\end{table*}

\section{Results} \label{sec:results}
In this section, we present quantitative results for the guideline question-answering methods (Sec. \ref{sec:guidelinecoverage} and Sec. \ref{sec:guidelineqaresults}). As a qualitative analysis, we also discuss themes and subthemes that we found from an analysis of discussions from our expert panel sessions (Sec. \ref{sec:studyresults}). For our risk prediction model we choose MLP, and derive important features for the prediction using the SHAP model. 
\subsection{Auxiliary Supporting Results}
\subsubsection{Risk Prediction and Model Explainer Results} \label{sec:riskpredictionresults}
We present the performance of the risk prediction models in Tab. ~\ref{tab:resultsCKD}. As the table shows, while GRU performs the best overall, depending on the use case, we may want to prefer other models. For the purposes of this paper, we chose MLP as our risk prediction model to benefit from the higher recall (such that the probability of false negatives is low) and high brier-score (to allow a more natural interpretation of our model outputs for clinicians), while still achieving an acceptable level of overall performance (AUC-ROC = $0.59$).

\begin{table}[ht!]
  \centering
  \caption{Results of CKD risk from different prediction models.}
  \label{tab:resultsCKD}
  \begin{tabular}{lccccc}
    \toprule
    Method & Precision & Recall & AUC-ROC & AUC-PRC & Brier\\
    \midrule
    LR    &  0.333 & 0.023  & 0.582 & 0.215 &  0.127 \\
    MLP   &  0.139 &  {\bf 0.977} & 0.587 & 0.224 & {\bf 0.621}  \\
    LSTM  &  {\bf 0.242} &  0.442 & {\bf 0.678} & 0.263 & 0.208  \\
    GRU   &  0.240  &  {\bf 0.605} & 0.677 & {\bf 0.311} & 0.220   \\
  \bottomrule
\end{tabular}
\end{table}

\begin{figure*}[!htbp]
    \centering
    \includegraphics[width=\linewidth]{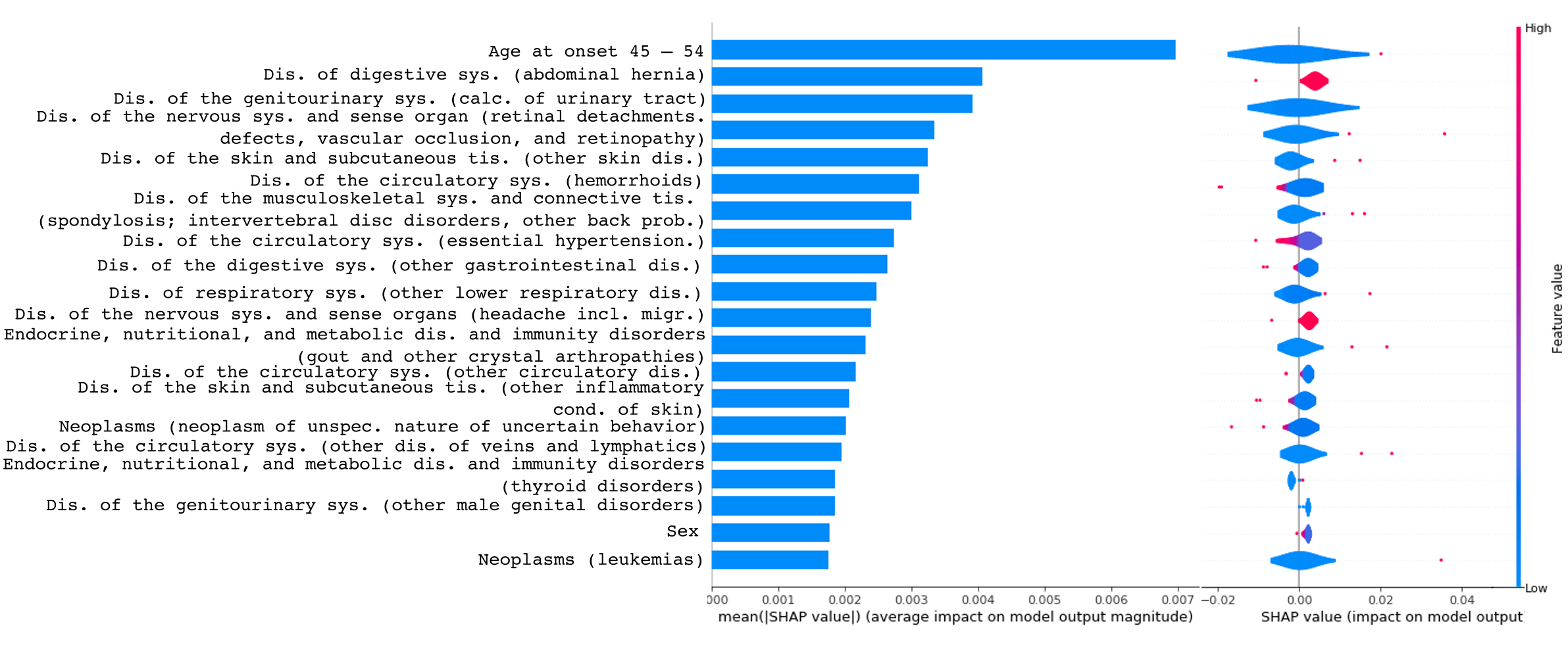}
    \vspace{-2em}
        \caption{Feature importance for {\ckd} prediction among {\numproto} prototypical patients using SHAP (left), showing absolute importance, and (right) showing feature impact on model prediction w.r.t. presence/absence of features. Reproduced from: S. Chari, P. Acharya, D.M. Gruen, O. Zhang, E.K. Eyigoz, M. Ghalwash, O. Seneviratne, F.S. Saiz, P. Meyer, P. Chakraborty, D.L. McGuinnesss, ``Informing clinical assessment by contextualizing post-hoc explanations of risk prediction models in type-2 diabetes,'' \textit{Artif. Intell. in Med.} J., vol. 137, Mar. 2023, Art. no. 102498, doi: 10.1016/j.artmed.2023.102498.}
        \label{fig:shap:summary_proto}
\end{figure*}

Feature importance for risk factors found by algorithmic explainers can be further contextualized, as mentioned previously.
The left hand column of Fig.~\ref{fig:shap:summary_proto} shows the top $20$ features for the set of {\numproto} prototypical patients under investigation. These prototypical patients are all found to be at high-risk for {\ckd} and hence, would be interesting to clinicians within the scope of this {\dm} and {\ckd} use case. 
For these prototypical patients, we present aggregated feature importance, as seen in Tab. \ref{tab:protosummary_appendix}, to account for HIPAA restrictions.  We can see that demographic features, such as age and the presence of other disorders, such as `other skin disorders,' were found to be important for the {\ckd} risk prediction. 
The right side of Fig. ~\ref{fig:shap:summary_proto} shows an alternate view of the same, providing a view into the spread of individual importance. From this deeper view, we can see that features such as `calculus of urinary tract' could be the most important drivers of risk for some patients. Such results further support our need to personalize features found to be important for the risk predictions. 
While insights about the importance of such features are helpful, such clinical and patho-physiological features may need further contextualization for clinicians. Our structured feedback sessions found that clinicians found the contextual explanations we support around these features helpful, and we cover some of this next.

\begin{table}
\centering
\caption{Summary (generated using Tableone library~\cite{pollard2018tableone}) of  $20$ prototypical patients highlighting the demographic and diagnoses counts. We report the disease diagnoses by their higher-level disease groupings (e.g., for {\dm} the higher-level code is endocrine, nutritional and metabolic disorders). We highlight in bold the conditions that are most prevalent amongst the patients ($> 50\%$).}
\small
\label{tab:protosummary_appendix}
 \begin{tabular}{lc}
\toprule
                                                                       Feature                  &     Overall counts ($\%$) \\
\midrule
Age at onset 45-54 &    4 (20.0) \\
\textbf{Age at onset} $\geq$ 55 &   \textbf{15 (75.0)} \\
Age at onset $\leq$ 44 &     1 (5.0) \\
SEX - FEMALE &    7 (35.0) \\
\midrule
Mood disorders  &    3 (15.0) \\
Diseases of the blood and blood-forming organs  &    3 (15.0) \\
\textbf{Diseases of the circulatory system} &   \textbf{17 (85.0)} \\
Diseases of the digestive system &    6 (30.0) \\
Diseases of the genitourinary system &    9 (45.0) \\
\begin{tabular}[c]{@{}l@{}} \textbf{Diseases of the musculoskeletal system and} \\ \textbf{ connective tissue} \end{tabular}  &   \textbf{12 (60.0)} \\
Diseases of the nervous system and sense organs  &    9 (45.0) \\
\textbf{Diseases of the respiratory system} &   \textbf{11 (55.0)} \\
Diseases of the skin and subcutaneous tissue &    7 (35.0) \\
\begin{tabular}[c]{@{}l@{}} \textbf{Endocrine; nutritional; and metabolic diseases and} \\ \textbf{ immunity disorders} \end{tabular} &  \textbf{20 (100.0)} \\
Infectious and parasitic diseases &   10 (50.0) \\
Injury and poisoning &    4 (20.0) \\
Mental Illness &    3 (15.0) \\
Neoplasms  &    6 (30.0) \\
\begin{tabular}[c]{@{}l@{}} Symptoms; signs; and ill-defined conditions \\ and factors influencing health status \end{tabular} &   10 (50.0) \\
\bottomrule
\end{tabular}
\end{table}

\subsubsection{Guideline Coverage} \label{sec:guidelinecoverage}
We extracted the recommendations and discussion sentences across the $16$ chapters of the current ADA 2021 CPGs. These recommendation and discussion sentences are expressed in natural language (See Fig. \ref{fig:qaguidelinestructure}). Table~\ref{tab:coverageguidelines} shows a high-level overview of the coverage statistics. Specifically, the extracted sentence corpus can hence be analyzed for the total number of tokens, average token length per sentence, and their composition of Metamap semantic types, to understand the coverage of the guideline text in terms of volume and semantic diversity. For tokens, we report words recognized by both BERT's tokenizer model to be consistent with our QA approach.
~\cite{hematialam2021identifying} report similar statistics for three other CPGs, neither of which are Diabetes focused, but it can be seen that the total number of tokens and sentences in the ADA 2021 CPG are more than the three guidelines reported in this paper. Hence, pointing to the comprehensiveness of our approach. Also, note that some of the recommendations were not captured by our guideline extraction script, and hence our statistics might be lesser than the actual count.~\footnote{In future, we aim to expand our coverage and update our methods to better capture these recommendation groups}.

\begin{table}[!htbp]
\centering
  \caption{Coverage statistics from extracted content from the ADA Standards of Care - Diabetes Guidelines 2021. We report these statistics on the recommendations and discussion sentences we extracted across chapters.} 
  \label{tab:coverageguidelines}
  \small
  \begin{tabular}{lc}
    \toprule
    Field & Count \\
    \midrule
    Chapters & 16 \\
    No. of sentences & 2379\\
   Tokens from BERT & 118350 \\
   Avg. BERT tokens per sentence &  49\\
    Metamap Semantic types covered & 116 / 126\\
  \bottomrule
\end{tabular}
\end{table}

\begin{figure*}[!htbp]
    \centering
    \includegraphics[width=\linewidth]{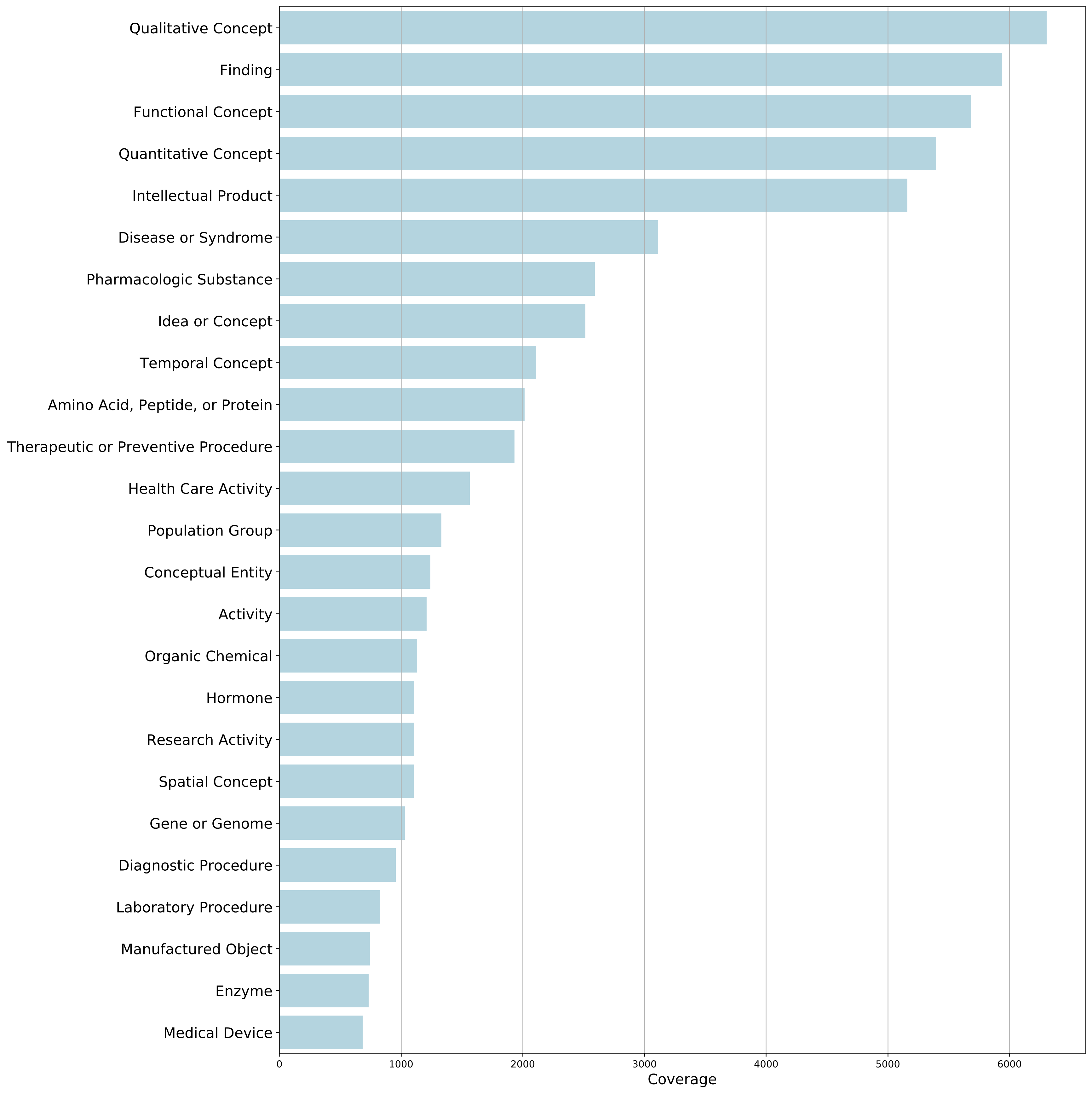}
    \vspace{-2em}
        \caption{Frequency distribution for $25$ / $116$ of the top semantic types that were found in the extracted guideline text. Reproduced from: S. Chari, P. Acharya, D.M. Gruen, O. Zhang, E.K. Eyigoz, M. Ghalwash, O. Seneviratne, F.S. Saiz, P. Meyer, P. Chakraborty, D.L. McGuinnesss, ``Informing clinical assessment by contextualizing post-hoc explanations of risk prediction models in type-2 diabetes,'' \textit{Artif. Intell. in Med.} J., vol. 137, Mar. 2023, Art. no. 102498, doi: 10.1016/j.artmed.2023.102498.}
        \label{fig:semantic_type_dist}
\end{figure*}

The many semantic types (see Fig. \ref{fig:semantic_type_dist} for 25 most populous semantic types) covered by the ADA 2021 CPG reaffirms that guidelines are a comprehensive source of evidence-based information in the clinical domain~\cite{murad2017clinical}. Further, in Sec. \ref{sec:discussion}, we also discuss how well the ADA 2021 CPG alone can support the themes we analyzed from conversations with clinicians during our structured feedback sessions and hence discuss the CPG's ability to serve as a source of context in our risk prediction setting.
\subsection{Quantitative Results - QA Performance} \label{sec:guidelineqaresults}
One of our key aim is to study whether SOTA LLM methods can be used to extract high quality contextual explanations. 
Thus, while we can address different question types in our QA approach, as seen in Tab. \ref{tab:questiontypes}, we only evaluate those question types that are addressed by ML methods, i.e., through our knowledge and knowledge augmented LLM modules (as described in Sec. \ref{sec:methods} and seen in Fig. \ref{fig:guidelineqa}). We evaluate feature importance questions of types 3 (of diagnostic importance), type 4 (of treatment importance), and  type 5 (asking about clinical indicators and important for both diagnostic and management purposes). Additionally, we evaluate answers to questions of types 3 and 4 
differently from question type 5 since questions of types 3 and 4 are served by the knowledge augmented LLM modules and questions of type 5 are addressed by the LLM with the numerical range comparison module. For answers to questions of type 3 and 4, we report standard and frequently used NLP QA metrics including mean average precision (MAP)~\cite{teufel2007overview}, F1-score their contributors of precision and recall and BLEU scores. For answers to questions of type 5, we report the number of times the combination of the LLM and numerical range comparison module could correctly predict whether the answer outputted was in / out range for the numerical value being asked about in the question. 
\subsubsection{Results on Questions with Disease, HbA1C Values and Drugs}
\par \textbf{\textit{Results on Disease Questions:}} We first evaluated the quality of extracted contextual explanations for feature importance questions, type 3. As outlined in Sec.{~\ref{sec:methods}}, our aim was to evaluate the readiness of SOTA LLM models for this task.
Here we report the results for $71$ questions covering relevant feature importances for patients' risk predictions and these questions cover $14$ CCS LVL 1 diagnosis code types. As mentioned previously, we report the performance in terms of a number of standard metrics. However,  given our information extraction setting, among these we are especially interested in the precision metrics. These metrics measure how many documents, among the ones retrieved, were relevant. Furthermore, `MAP' and `precision@k' measures the same while considering the order of retrieval. This aligns closely with how clinicians evaluate presented informations where the presented answers are expected to accurate in order of most acceptable to least.  
It is to be noted, that we evaluated the predicted results to disease feature importance questions against candidate answers by manually inspecting the ADA 2021 CPG. The annotations were done by an author and some of these annotations were verified by a clinical expert on the team who is also a co-author in this paper. We report results on the expert validated subset in \ref{sec:result_settings:appendix}. Table~\ref{tab:nativemodelresults} reports results in comparison to the entire annotated dataset of $85$ feature questions and $654$ candidate answers for the out-of-the shelf LLM models under consideration. The results show that in terms of MAP as well as precision at k metrics, vanilla BERT outperforms the other LLM models. SciBERT is a close second with an improved recall and F1 score. We analyze the importance of these results further in Sec.~\ref{sec:qadiscussion}.

Additionally, we evaluated the results by augmenting the base LLM models with different strategies. 
Tab.~\ref{tab:kmdiseaseresults} reports the result scores for answers to question type 3 using the best knowledge augmentation strategy (across the five different settings) for each LLM models. Overall, we can see a significant improvement in MAP and precision at 5 for knowledge augmented BERT model (BERT-KA) over BERT, the best performing base language model. In terms of other metrics, knowledge augmented SciBERT-KA shows a consistent improvement for all metrics while being best/second-best overall.

While these evaluation numbers are reported from a small evaluation set of $71$ questions and $654$ candidate actual answers, we consider this as a somewhat comprehensive evaluation due to the diversity of diseases covered ($1844$ diseases) and in total semantic types covered within the answers ($116$ semantic types, see Tab. \ref{tab:coverageguidelines}), both in the candidate and predicted sets. Overall, from these results, we see that  knowledge augmentation can improve the base language model performance. 

\begin{table}[!htbp]
\caption{Performance of Guideline QA on different language model approaches reported at mean average precision (MAP), F1 and recall at top-10 answers and precision at top-1 and top-5 for $71$ disease feature importance questions. The models are sorted by MAP values, to indicate an ordering of the best models. }

\label{tab:nativemodelresults}
\small
\centering
\begin{tabular}{lrrrrrr}
\toprule
{} & {bleu} & {P@1} & {P@5} & {map} & {f1} & {recall} \\
{model} & {} & {} & {} & {} & {} & {} \\
\midrule
BERT & 0.117 & \color{green} 0.468 & \color{green} 0.382 & \color{green} 0.390 & 0.213 & 0.241 \\
BioBERT & 0.116 & 0.431 & 0.339 & 0.346 & 0.200 & 0.238 \\
BioBERT-BioASQ & \color{blue} 0.132 & 0.383 & 0.329 & 0.332 & \color{blue} 0.217 & \color{blue} 0.281 \\
BioClinicalBERT-ADR & 0.125 & 0.368 & 0.317 & 0.316 & 0.205 & 0.259 \\
SciBERT & \color{green} 0.165 & \color{blue} 0.461 & \color{blue} 0.349 & \color{blue} 0.364 & \color{green} 0.261 & \color{green} 0.354 \\
\bottomrule
\end{tabular}
\end{table}

\begin{table}[!htbp]
\centering
\caption{Performance of Guideline QA of knowledge augmented language models reported at mean average precision (MAP), F1 and recall at top-10 answers and precision at top-1 and top-5 for $71$ disease feature importance questions. Here we show the best knowledge augmentation approach per model to indicate highest gains over baseline performance for the native language model approaches we tried. Best and second-best values for each column are highlighted in green and blue color, respectively. Language model (e.g. BERT) suffixed with KA represents the corresponding knowledge augmented model (e.g. BERT-KA).}
\label{tab:kmdiseaseresults}
\small
\begin{tabular}{lrrrrrr}
\toprule
{} & {bleu} & {P@1} & {P@5} & {map} & {f1} & {recall} \\
{model} & {} & {} & {} & {} & {} & {} \\
\midrule
BERT-KA & 0.075 & \color{blue} 0.467 & \color{green} 0.419 & \color{green} 0.438 & 0.169 & 0.186 \\
BioBERT-KA & 0.127 & 0.434 & 0.348 & 0.353 & 0.215 & 0.254 \\
BioBERT-BioASQ-KA & \color{blue} 0.141 & 0.458 & \color{blue} 0.362 & 0.369 & \color{blue} 0.237 & \color{blue} 0.280 \\
BioClinicalBERT-ADR-KA & 0.121 & 0.406 & 0.321 & 0.330 & 0.202 & 0.242 \\
SciBERT-KA & \color{green} 0.192 & \color{green} 0.473 & 0.341 & \color{blue} 0.375 & \color{green} 0.291 & \color{green} 0.405 \\
\bottomrule
\end{tabular}
\end{table}

\par \textbf{\textit{QA Results on HbA1C Questions:}}

\begin{table}[!htbp]
\centering
  \caption{Results of Guideline QA with rule augmentation of language model approaches for numerical comparisons reported for $9$ questions across the $20$ prototypical patients identified from our predicted high-risk chronic kidney disease cohort. The split of question variations is equal across the different numerical range comparison operators of lesser than, equal to and greater than.} 
  \label{tab:resultsguidelines_numerical}
  \small
  \begin{tabular}{lllllll}
    \toprule
    Comparison & Accuracy & TP & TN & FP & FN & Total  \\
    \midrule
     Overall & 0.78 & 7 & 7 & 3 & 0 & 18 \\
     Lesser Than & 0.84 & 2 & 3 & 1 & 0 & 6  \\
    Equal To & 0.67 & 1 & 3 & 2 & 0 & 6  \\
     Greater Than & 100 & 4 & 2 & 0 & 0 & 6  \\
  \bottomrule
  \multicolumn{7}{m{10cm}}{TP - True Positives, TN - True Negatives, FP - False Positives, FN - False Negatives. Accuracy computed as accuracy = (TP + TN)/ Total} \\
\end{tabular}
\end{table}

In Tab. \ref{tab:resultsguidelines_numerical}, we report the accuracy statistics for the performance of our rule augmentation / numerical range comparison module of our QA approach. We show granular breakdowns based on different numerical operators, greater than, lesser than, and equal to, to indicate which settings are being picked up the best by the rule augmentation and which others are harder. In this evaluation, we manually went through the outputted answers to ensure that they were within range of the numerical values in questions. The reason for this annotation approach is that the guidelines have few sentences for actions to be taken on clinical indicators.
Hence, there is not much diversity in the answers that a LLM like BERT can output before passing the answer to our range comparison module for validation. These accuracies highlight the value in combining syntactic parsing output with a LLM to improve its capabilities. 

\par \textbf{\textit{QA Results on Drug Questions:}} To demonstrate the flexibility of the LLM setup on various settings that match the coverage of the T2D guideline data (see Fig. \ref{fig:semantic_type_dist}) as well as to test the generalizability of our results. We also report results for $6$ anti-diabetic drug questions of question type 4. 
Table~\ref{tab:drugresults} show the results for out-of-the shelf language models as well as the knowledge augmented language models for the metrics of interest.
Overall, we once again found the knowledge-augmented language models to be the best performing ones. In terms of `MAP' and `precision at 5', BERT-KA comes out as the best performing model, where as SCIBERT-KA comes out as either the best or the second-best model for all metrics. It is to be noted that the overall results are significantly better than the ones for disease questions. One possible reason behind this effect may be related to the fact that drugs are referred directly in guidelines and thus QA models are able to pick these sentences with greater efficacy.

\begin{table}[!htbp]
\caption{Performance of Guideline QA with different knowledge augmentations of language model approaches reported at mean average precision (MAP), F1 and recall at top-10 answers and precision at top-1 and top-5 for $6$ anti-diabetic drug feature questions. Best and second-best values for each column are highlighted in green and blue color, respectively. Language model (e.g. BERT) suffixed with KA represents the corresponding knowledge augmented model (e.g. BERT-KA).}
\label{tab:drugresults}
\small
\centering
\small
\begin{tabular}{lrrrrrr}
\toprule
{} & {bleu} & {P@1} & {P@5} & {map} & {f1} & {recall} \\
{model} & {} & {} & {} & {} & {} & {} \\
\midrule
BERT & 0.100 & 0.910 & 0.751 & 0.757 & 0.254 & 0.206 \\
BioBERT & 0.100 & 0.726 & 0.643 & 0.635 & 0.231 & 0.192 \\
BioBERT-BioASQ & 0.081 & 0.708 & 0.694 & 0.704 & 0.222 & 0.162 \\
BioClinicalBERT-ADR & 0.075 & 0.593 & 0.614 & 0.597 & 0.192 & 0.146 \\
SciBERT & \color{blue} 0.121 & \color{green} 0.947 & 0.757 & 0.772 & \color{blue} 0.281 & \color{blue} 0.228 \\
\midrule
BERT-KA & 0.099 & 0.900 & \color{green} 0.863 & \color{green} 0.821 & \color{blue} 0.281 & 0.213 \\
BioBERT-KA & 0.083 & 0.802 & 0.704 & 0.720 & 0.234 & 0.170 \\
BioBERT-BioASQ-KA & 0.117 & 0.711 & 0.725 & 0.716 & 0.272 & 0.221 \\
BioClinicalBERT-ADR-KA & 0.085 & 0.598 & 0.595 & 0.587 & 0.199 & 0.152 \\
SciBERT-KA & \color{green} 0.128 & \color{blue} 0.912 & \color{blue} 0.823 & \color{blue} 0.794 & \color{green} 0.298 & \color{green} 0.232 \\
\bottomrule
\end{tabular}

\end{table}

Overall,  these quantitative evaluations points to the potential in applying scalable, augmented LLM-based approaches to extract content from authoritative guideline literature that can then be used to provide context to interpret model predictions, such as in our setting, risk prediction scores and their model explanations.

\subsubsection{Disease Subgroup and LLM Variants Analysis}
For our feasibility analysis, we further analyze the results for type 3 questions with respect to the disease groups to gain a deeper understanding of the QA performance. 
Specifically, we want to understand whether SOTA LLM backed QA methods, potentially augmented with knowledge, are ready for real-world use for our states use-case as well as identify patterns that might apply to other CPGs in different disease areas. We pose a number of questions around this idea as follows:

\noindent \ul{Is CPG guideline suitable for {\dm} contexts}: \textit{How well are the disease subgroups among {\dm} patients covered in the guidelines?}
\begin{figure*}[!htbp]
    \centering
    \includegraphics[width=\linewidth]{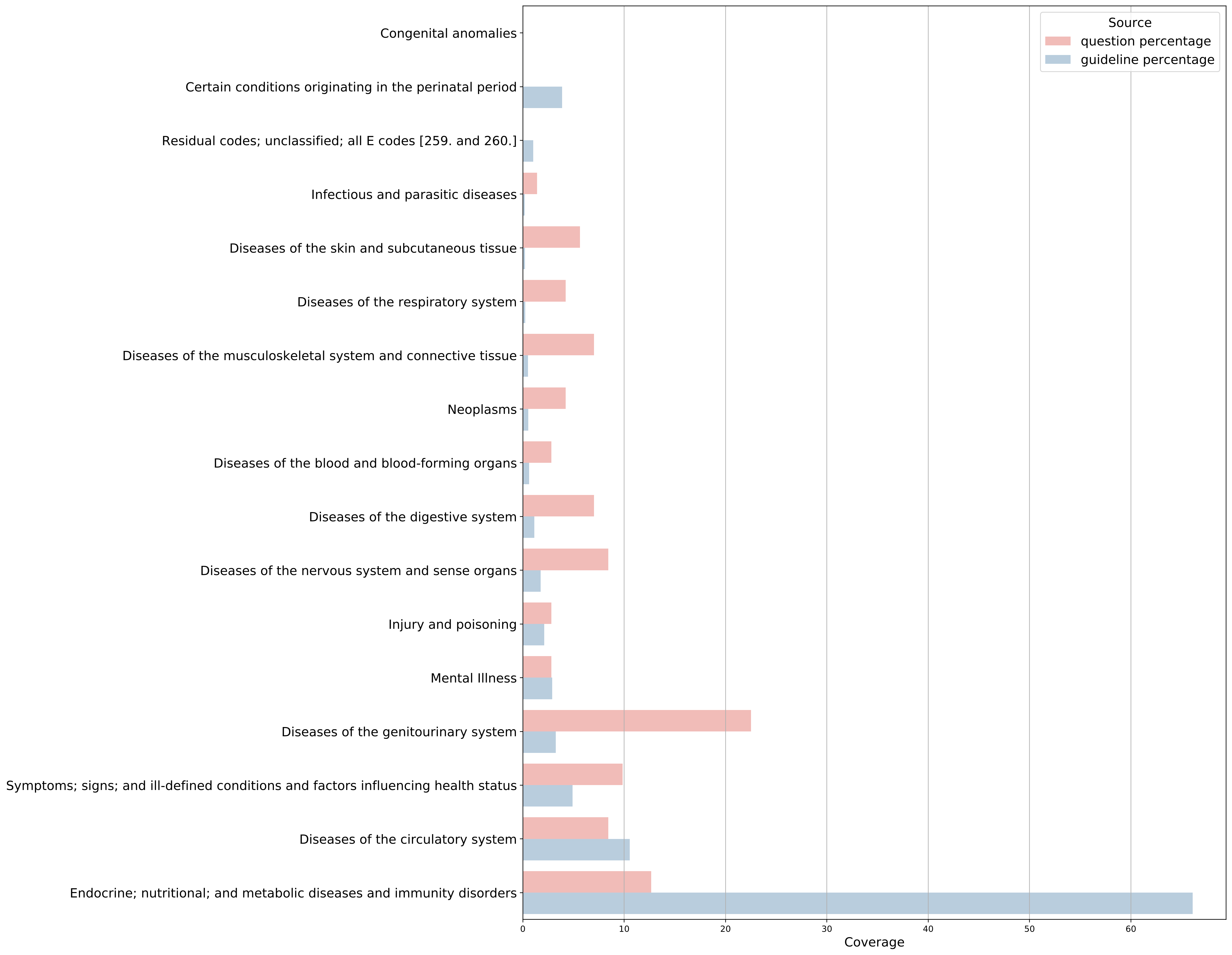}
    \vspace{-2em}
        \caption{Comparing disease group occurrences in the ADA CPG 2021 versus those in the feature importance questions from our chosen prototypical patients. Reproduced from: S. Chari, P. Acharya, D.M. Gruen, O. Zhang, E.K. Eyigoz, M. Ghalwash, O. Seneviratne, F.S. Saiz, P. Meyer, P. Chakraborty, D.L. McGuinnesss, ``Informing clinical assessment by contextualizing post-hoc explanations of risk prediction models in type-2 diabetes,'' \textit{Artif. Intell. in Med.} J., vol. 137, Mar. 2023, Art. no. 102498, doi: 10.1016/j.artmed.2023.102498.}
        \label{fig:guidelines_patient_diseases}
\end{figure*}

Here, we attempt to understand the applicability of ADA 2021 CPG in our use-case. From Fig.~\ref{fig:guidelines_patient_diseases}, we can interpret that the guidelines cover a smaller number of disease groups~\footnote{Disease groups are derived by rolling up disease codes both in the patient data and guidelines to their higher-level CCS LVL 1 groups.} than the patient data. Since CPGs are authoritative literature in their disease fields, their coverage is mainly limited to the primary disease area. Thus ADA 2021 CPG focuses on Diabetes, an Endocrine, Nutritional and Metabolic Disorder, and its comorbid conditions (mainly spanning diseases of the circulatory and genitourinary systems). 
Unsurprisingly, these patterns are seen in the Fig~\ref{fig:guidelines_patient_diseases} as well where the Endocrine, Nutritional and Metabolic Disorder have the largest coverage in the guideline data, a $66\%$.
In contrast, patients might have other conditions that do not arise from the {\dm} diagnosis alone, and hence we can deduce that we see more diversity in disease groups in the patient data.

\begin{figure}[hbt!]
\centering
\includegraphics[width=1.0\linewidth]{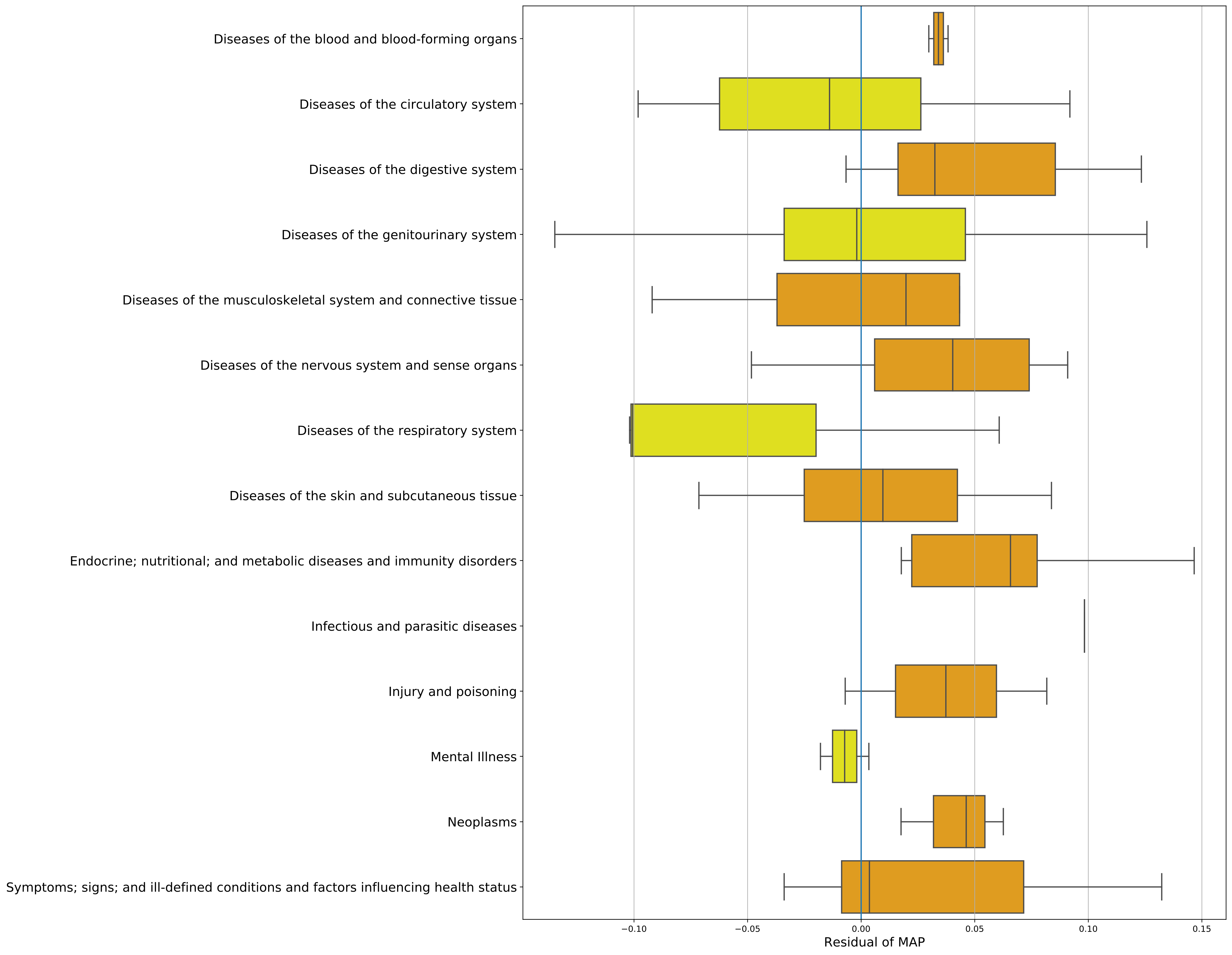}
\caption{Differences between two of the best performing LLMs, SciBERT and BERT, in our QA setup against disease subgroups in the patient questions. Reproduced from: S. Chari, P. Acharya, D.M. Gruen, O. Zhang, E.K. Eyigoz, M. Ghalwash, O. Seneviratne, F.S. Saiz, P. Meyer, P. Chakraborty, D.L. McGuinnesss, ``Informing clinical assessment by contextualizing post-hoc explanations of risk prediction models in type-2 diabetes,'' \textit{Artif. Intell. in Med.} J., vol. 137, Mar. 2023, Art. no. 102498, doi: 10.1016/j.artmed.2023.102498.}
\label{fig:bestmodelresults} 
\end{figure}

\noindent \ul{Can a single SOTA LLM method be used to extract the contexts?}
We attempt to understand if the LLMs are inherently better at certain disease groups over others. Fig~\ref{fig:bestmodelresults} shows the distribution of base LLM models over the disease groups. We can see that there are a few disease groups which have a higher MAP performance than others (towards the right end of the plot), some have their the box centers in the middle of the plot and others who are not doing as well since they are in the first quadrant of the plot. While the results are not strikingly decisive and statistically significant everywhere,  in concordance with our overall results, we note that SciBERT and BERT models have  better performance over most of the disease groups.
Thus to further discern between these top $2$ performing models, we conducted a point-wise analysis of relative performance difference between BERT and SciBERT (distribution of residual values between the MAP performances of BERT and Scibert under equal performance hypothesis). Fig. \ref{fig:bestmodelresults} shows the outcomes of the analysis where orange box indicates that BERT performs better on average while yellow indicates the same for SciBERT. 
We see that BERT is better for most disease groups, especially for `Disease of the blood and bone forming organs', `Diseases of the digestive system', `Diseases of the nervous system and sense organs', `Endocrine, nutritional, and metabolic diseases and immunity disorders', `Injury and poisoning', and `Neoplasms' ($0$ not contained in the inter-quartile range).   SciBERT is only doing better on `Diseases of the respiratory system' (and marginally better for `Mental Illness'). 
These results, in addition to the quantitative results, indicate that LLM models are better at addressing some disease groups than others. While vanilla BERT is a defensible choice, the results point to the need for domain adaptation for LLM for this problem. However, considering the limited availability of data, novel ML methods such as one-short learning and as weak supervised techniques may be 
required to improve the performances of these LLMs reliably across multiple disease groups. 

\begin{figure}[hbt!]
\centering
\includegraphics[width=1.0\linewidth]{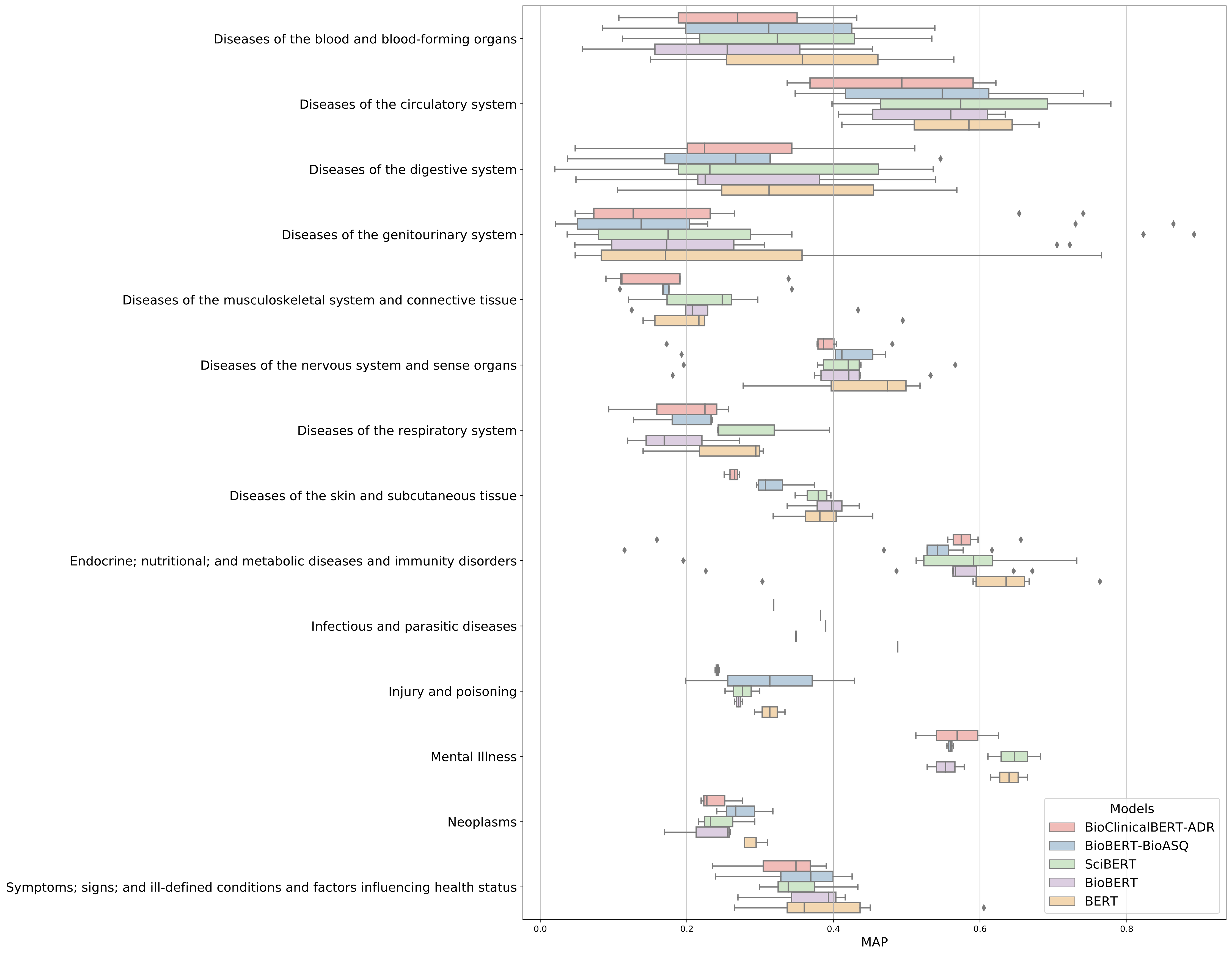}
\caption{Differences in performance on various disease groups in the patient questions by the different LLM models we tried in the guideline QA. Reproduced from: S. Chari, P. Acharya, D.M. Gruen, O. Zhang, E.K. Eyigoz, M. Ghalwash, O. Seneviratne, F.S. Saiz, P. Meyer, P. Chakraborty, D.L. McGuinnesss, ``Informing clinical assessment by contextualizing post-hoc explanations of risk prediction models in type-2 diabetes,'' \textit{Artif. Intell. in Med.} J., vol. 137, Mar. 2023, Art. no. 102498, doi: 10.1016/j.artmed.2023.102498.}
\label{fig:diseasemodelresults}  
\end{figure}

\vspace{1em}

\begin{figure}[hbt!]
\centering
\includegraphics[width=1.0\linewidth]{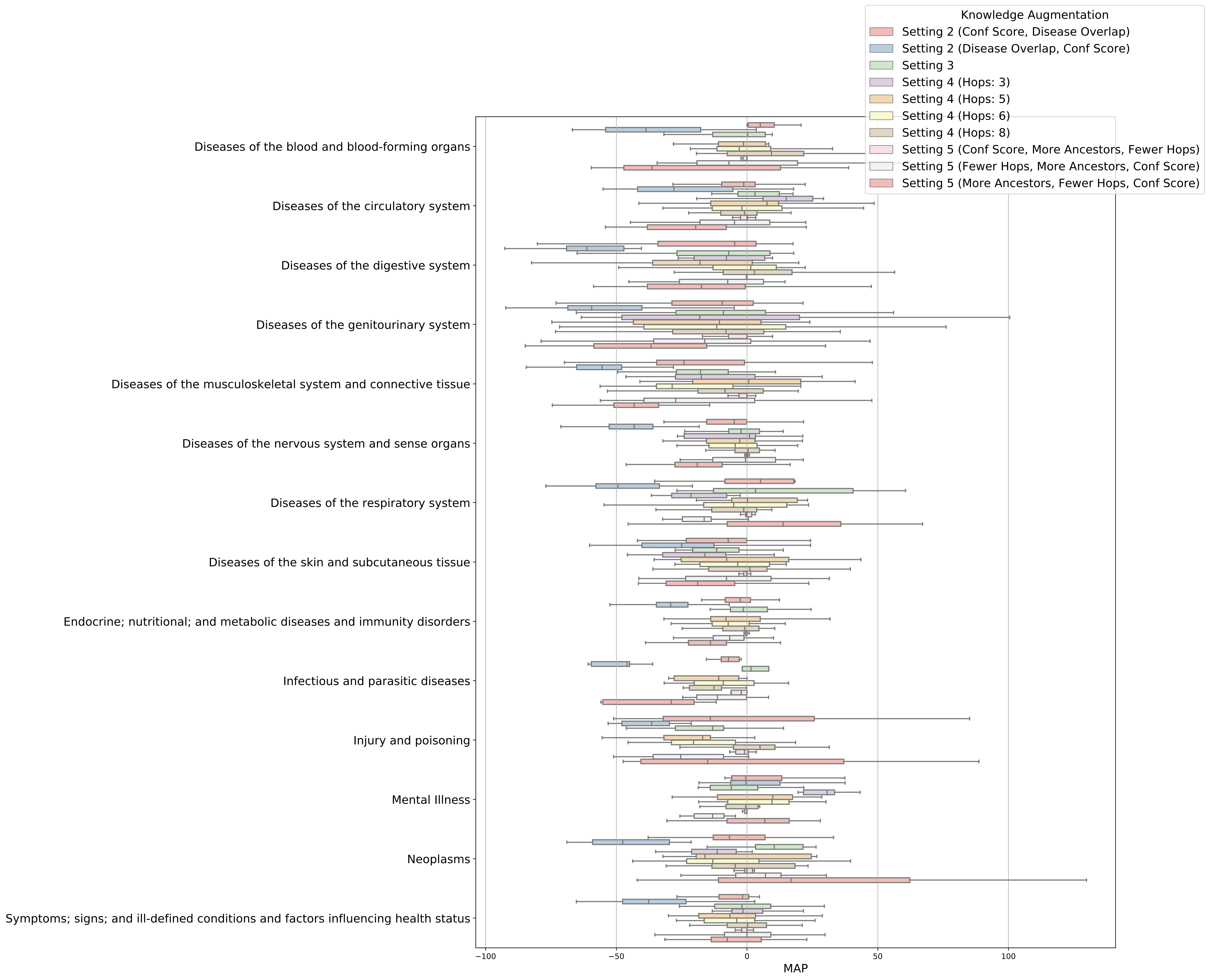}
\caption{Differences between knowledge augmentation strategies in our QA setup against disease subgroups in the patient questions. Reproduced from S. Chari, P. Acharya, D.M. Gruen, O. Zhang, E.K. Eyigoz, M. Ghalwash, O. Seneviratne, F.S. Saiz, P. Meyer, P. Chakraborty, D.L. McGuinnesss, ``Informing clinical assessment by contextualizing post-hoc explanations of risk prediction models in type-2 diabetes,'' \textit{Artif. Intell. in Med.} J., vol. 137, Mar. 2023, Art. no. 102498, doi: 10.1016/j.artmed.2023.102498.}
\label{fig:bestkaresults} 
\end{figure}

\noindent \ul{Does knowledge augmentation reliably improve QA methods?}
We proposed $4$ possible strategies for Knowledge-augmentation (See Appendix~\ref{sec:qamethod:appendix}).  Among these settings, the best knowledge augmentation strategies reflected in Table.~\ref{tab:kmdiseaseresults} originated from a composite of strategies. As seen from Tab.~\ref{tab:nativemodelresults}, ~\ref{tab:kmdiseaseresults} and ~\ref{tab:drugresults}, the guideline QA's \textit{best MAP score of 0.82} is obtained in a BERT + knowledge augmented setting on drug questions and among disease features, the guideline QA's \textit{best MAP score is 0.438}. The \textit{recall with its highest value of 0.405} is obtained in a post-filtering knowledge augmentation setting 5 of SciBERT (Tab. \ref{tab:kmdiseaseresults}), where we sort answers by disease overlap between question and answer and we also see the best \textit{BLEU score of 0.19} in this setting.
These points to the fact that there is value in either filtering the answers to be passed to a LLM  or sorting the answers from it, using aids from known domain knowledge sources that data sources like guidelines are expected to adhere to.
We further analyzes these at disease sub-group level in Fig.~\ref{fig:bestkaresults} where we plot the lift in performance over corresponding base LLM using any particular strategy. Although any one strategy is not found to dominate the others, for most disease groups, we can find one or more strategies that improve performance (median lift greater than $0$). These results support our previous insight that there is a value in augmenting domain knowledge. However, finding a single universal strategy is difficult and may need further research.

\subsection{Dashboard} \label{sec:dashboard}
\begin{figure*}[!htbp]
    \centering
    \includegraphics[width=0.7\linewidth]{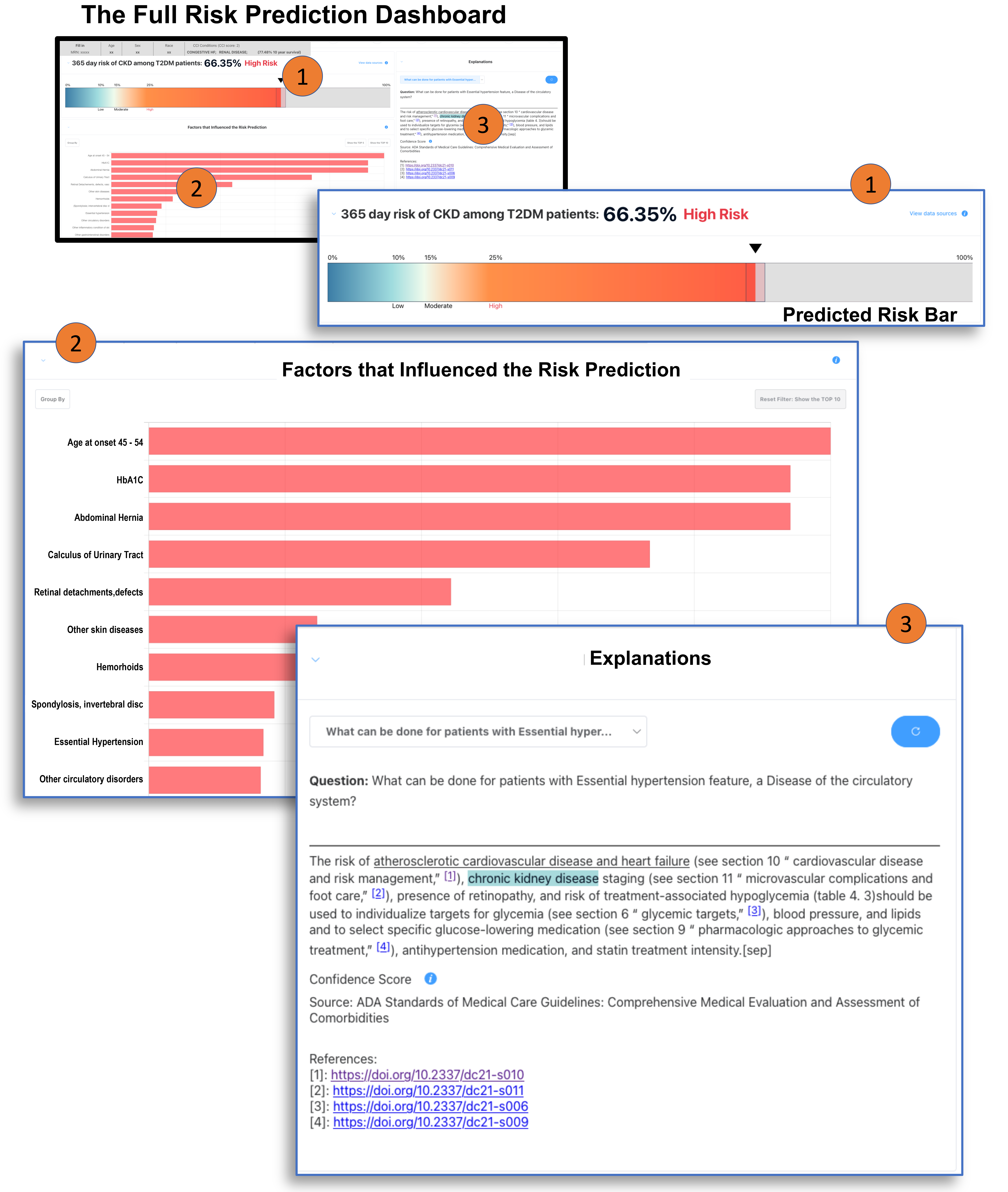}
        \caption{A screenshot of a running prototype of our risk prediction dashboard which includes: 1) the risk prediction score, 2) the features contributing to the predicted risk with the size of their impact on the model results, and 3) a "questions in context pane", in which the user can select and see answers to questions that provide additional contextual information about the patient, the predicted risk calculation, and individual features contributing to the risk. Reproduced from: S. Chari, P. Acharya, D.M. Gruen, O. Zhang, E.K. Eyigoz, M. Ghalwash, O. Seneviratne, F.S. Saiz, P. Meyer, P. Chakraborty, D.L. McGuinnesss, ``Informing clinical assessment by contextualizing post-hoc explanations of risk prediction models in type-2 diabetes,'' \textit{Artif. Intell. in Med.} J., vol. 137, Mar. 2023, Art. no. 102498, doi: 10.1016/j.artmed.2023.102498.}
        \label{fig:prototypedashboard}
\end{figure*}

To present the supported contextual explanations, 
we have adapted a question-driven design~\cite{liao2020questioning} for user-interface (UI) development and built a running prototype of a risk prediction dashboard (as seen in Fig.~\ref{fig:prototypedashboard}). The content we show on it is rendered on a per-patient basis chosen from a landing page not shown here. For each patient, we show multiple panes (or UI sections) at a high level, each of which displays content under a particular grouping. These panes include groupings of \textit{patient details, history timeline of claim incidences, risk prediction scores, features contributing to risk, and questions in context}. In Fig. \ref{fig:prototypedashboard}, we highlight the risk prediction, feature importance, and questions in context panes. The explanations pane serve as a section where our contextual explanations, that provide context around our identified entities of interest in the risk prediction setting - patients, their predicted risk, and the features contributing to risk - can be selected and browsed. Additionally, as we have described in Sec. \ref{sec:background}, risk scores can be interpreted better in the context of use, i.e., by enabling connections to patient data, feature importance, and domain knowledge, hence, we had to support interactions between these panes, which would make it easier for clinicians to establish the connections.

\subsubsection{Expert Panel Sessions Using Prototype Dashboard as an Aid}
We used our risk prediction dashboard as an aid during our structured feedback sessions, where we walked clinicians through a live demonstration of our dashboard for a set of prototypical patients (see Tab. \ref{tab:protosummary_appendix}). We conducted sessions individually with four clinicians in our expert panel to understand whether the contextual explanations provided, patient predicted risk, and risk explanations, were helpful for clinical practice. We \textit{explained that this dashboard would be available in addition to the clinician's regular EHR tools and the patient information they provide}, and is meant specifically to provide additional information related to the CKD Risk Prediction. To strike a balance between limited clinician time and the need for diverse feedback, we generated such reports from among $20$ prototypical {\ckd} high-risk patients from our {\dm} cohort, identified by the Protodash algorithm~\cite{gurumoorthy2019efficient}. 

During the sessions, we first familiarized each clinician with the different sections of the risk-prediction dashboard. We asked them to imagine that they would be meeting with the patient and had seen the CKD prediction, and stated we wanted to understand what information would be useful to them in understanding the prediction and its impact on their treatment decisions. We then presented the dashboard as it would appear for $3$ randomly selected prototypical patients. We asked the panel members to imagine that they were preparing to treat a patient that was new to them.  We navigated through the dashboard as instructed by the subjects, opening sections or clicking on items as they requested.  We asked the clinicians to speak aloud as they were working with the dashboard.  We also probed the relevance and usefulness of the different sections of the dashboard and the specific content shown in them.  We asked if there was other information they would have liked to have been provided, or questions they would want answered. Sessions were recorded and transcribed, similar to the approach mentioned in ~\cite{knoll2022user}. 

Through these sessions, we wanted to understand the usefulness of our supported patient contextualizations and the features contributing to their risk. Specifically, we showed clinicians the content on different panes of this dashboard and the supported interactions to understand  what features were most important, both from a UI and informational perspective. We report the results of these interactions in Sec. \ref{sec:studyresults}. 

\subsection{Qualitative Results - Themes from Expert Panel Interviews} \label{sec:studyresults} 
We conducted a thematic analyses on the responses and feedback received during the expert panel sessions (Sec. \ref{sec:dashboard}), as follows.  Three independent researchers, who are coauthors on this paper, reviewed the transcripts, flagged significant utterances, and characterized these utterances in terms of the major points and themes they expressed.  The researchers then reviewed their sets of identified themes and utterances together, and grouped and combined them into a single agreed-upon set of overarching themes.   We report this combined list of themes below (Tab. \ref{tab:thematicanalysis1}, \ref{tab:thematicanalysis2}, \ref{tab:thematicanalysis3}, \ref{tab:thematicanalysis4}). These themes reflect areas that clinicians prioritize and where the support of explanation-driven, AI risk prediction tools would be appreciated. 

As seen in Tab. \ref{tab:thematicanalysis1}, \ref{tab:thematicanalysis2}, \ref{tab:thematicanalysis3} and \ref{tab:thematicanalysis4}, we have grouped the discussions from the expert panel sessions into \textit{four high-level themes} spanning different areas where clinicians would benefit the most in a chronic disease, comorbidity risk-prediction setting such as ours. The high-level themes we found include `Theme 1: Clinical Value of Explanations and Contextualizations', `Theme 2: Highlighting Actionability', `Theme 3: Connections to Patient Data' and `Theme 4: Connections to External Knowledge'. We were further able to create sub-themes for more granular topics that came up during the discussions under each of these themes, bringing the theme and sub-theme total to {\textit{four high-level themes and twelve sub-themes}}. 

\begin{table}[!htbp]
\begin{center}
\caption{\textit{Clinical Value of Explanations and Contextualizations} - $1^{\text{st}}$ theme that emerged during our expert panel interviews with clinicians where we walked through the risk prediction dashboard and the contextual explanations that we support. We attach a description for each sub-theme that we found and we also provide examples in quotes.}
\label{tab:thematicanalysis1}
\scriptsize
\begin{tabular}{|l|l|}
      \toprule
      \multicolumn{1}{m{4cm}}{Sub-theme} & \multicolumn{1}{m{4cm}}{Description} \\
      \midrule
      \multicolumn{1}{m{4cm}}{Value of Contextual Information around {\ckd} risk} & \multicolumn{1}{m{10cm}}{All clinicians saw value in connecting the {\dm} patient's {\ckd} risks to \textit{data on their other conditions}, and to, \textit{relevant recommendations from the {\dm} guidelines}. For example, some clinicians reasoned about ``how the patient's {\ckd} risk changes their dosage / treatment'' and some others were interested about ``connections to other conditions that patient has.''} \\ \midrule
      \multicolumn{1}{m{4cm}}{Value of Contextual Information around Individual Features} & \multicolumn{1}{m{10cm}}{Clinicians found that information from {\dm} guidelines and cited literature relevant to factors that contributed to the system's predicted {\ckd} risk were helpful to understand how the \textit{factors could be related to {\ckd} or {\dm}}, and how they might interact with \textit{other factors shown}.  For example, ``how does a skull fracture elevate CKD risk?''  This was particularly valuable when not previously known by the clinician, for example:  ``it is surprising, and I have learned something about celiac disease and abdominal pain connection'' in patients with diabetes.}  \\ 
      \midrule
       \multicolumn{1}{m{4cm}}{Value of Contextual Information around patient's {\dm}} & \multicolumn{1}{m{10cm}}{Besides the patient's {\ckd} risk and its implications, the clinicians were interested to know about the \textit{patient's {\dm} state, their comorbid conditions and other parameters in relation to their {\dm} diagnosis}. For example, the clinicians wanted to know ``how long has the patient had their {\dm}'' or ``what is their A1C progression?''} \\ \bottomrule
\end{tabular}
\end{center}
\end{table}

\begin{table}[!htbp]
\begin{center}
\caption{\textit{Highlighting Actionability} - $2^{\text{nd}}$ theme that emerged during our expert panel interviews with clinicians, where we walked through the risk prediction dashboard and the contextual explanations that we support. We attach a description for each sub-theme that we found and we also provide examples in quotes.}
\label{tab:thematicanalysis2}
\scriptsize
\begin{tabular}{|l|l|}
      \toprule
      \multicolumn{1}{m{4cm}}{Sub-theme} & \multicolumn{1}{m{4cm}}{Description} \\
      \midrule
     \multicolumn{1}{m{4cm}}{Highlight Actionable and Modifiable Factors} & \multicolumn{1}{m{10cm}}{Most of the clinicians were interested in highlighting patient risk factors that could be controlled or acted upon, vs. those (such as age) that could not be influenced: ``what factors can be changed?''} \\ \midrule
           \multicolumn{1}{m{4cm}}{Highlight the Impact of CKD risk prediction on Treatment Decisions for Diabetes and other conditions} & \multicolumn{1}{m{10cm}}{When shown information from treatment guidelines, the clinicians wanted to understand how they \textit{reflect} \textit{the patient's CKD risk and T2D diagnosis}:``are any of these proposed medications contraindicated?''} \\ \midrule
      \multicolumn{1}{m{4cm}}{Suggest Specific Actions to Reduce CKD risk } & \multicolumn{1}{m{10cm}}{Clinicians wanted to understand ways to reduce the CKD risk including ways of addressing risk factors and changes to medications: ``do any of the patient's current medications increase risk of renal toxicity?'} \\ \bottomrule
\end{tabular}
\end{center}
\end{table}

\begin{table}[!htbp]
\begin{center}
\caption{\textit{Connection to Patient Data} - $3^{\text{rd}}$ theme that emerged during our expert panel interviews with clinicians, where we walked through the risk prediction dashboard and the contextual explanations we support. We attach a description for each sub-theme that we found and we also provide examples in quotes.}
\label{tab:thematicanalysis3}

\scriptsize
\begin{tabular}{|l|l|}
      \toprule
      \multicolumn{1}{m{4cm}}{Sub-theme} & \multicolumn{1}{m{4cm}}{Description} \\
      \midrule
   \multicolumn{1}{m{4cm}}{Connections to Patient's Clinical Indicators} & \multicolumn{1}{m{10cm}}{The clinicians indicated that they want to see clinical indicators for diagnoses ( if available ), when interpreting the \textit{factors} \textit{that led to the risk} or the \textit{patient's CKD risk score}: ``Given the patient has essential hypertension, what was their lab systolic blood pressure reading,'' or ``The patient's eGFR value will be important to show for CKD,''  or  ``what do the guidelines say about this patient's systolic and diastolic blood pressure readings.''} \\ \midrule
    \multicolumn{1}{m{4cm}}{Need for Information on Related Diagnoses} & \multicolumn{1}{m{10cm}}{When shown factors which were diagnosis codes that influenced the risk prediction, clinicians wanted to \textit{see what other diagnoses} \textit{ that the patients had, that might align or contribute}: ``show COVID-19 answers for lower respiratory disorders?'' or ``what episodes of abdominal pain did the patient have?''}\\ \midrule
    \multicolumn{1}{m{4cm}}{Connections to Patient's History} &  \multicolumn{1}{m{10cm}}{When shown certain factors, the clinicians wanted to know the when the patient had the diagnosis,  if the condition was a current one, and about changes over time. for example, ``when did the patient have a genitourinary diagnosis?,'' or``what does their eGFR progression look like?' } \\ \bottomrule
 \end{tabular}
\end{center}
\end{table}   
    
\begin{table}[!htbp]
\begin{center}
\caption{\textit{Connection to External Medical Domain Knowledge} - $4^{\text{th}}$ theme that emerged during our expert panel interviews with clinicians, where we walked through the risk prediction dashboard and the contextualization we support. We attach a description for each sub-theme that we found and we also provide examples in quotes.}
\label{tab:thematicanalysis4}
\scriptsize
\begin{tabular}{|l|l|}
      \toprule
      \multicolumn{1}{m{4cm}}{Sub-theme} & \multicolumn{1}{m{4cm}}{Description} \\
      \midrule
     \multicolumn{1}{m{4cm}}{Links to Medication Databases} &  \multicolumn{1}{m{10cm}}{When deciding upon \textit{treatment suggestions} for patients given the \textit{knowledge of their {\ckd} risk and {\dm} diagnosis},the clinicians wanted to understand how their medications interact : ``what drugs they are on currently have a bad renal impact?''  or ``how does their current anti-diabetic drug interact with a {\ckd} drug?''} \\ \midrule
     \multicolumn{1}{m{4cm}}{Links to Published Articles} & \multicolumn{1}{m{10cm}}{When connections between the \textit{{\ckd} risk prediction} and \textit{the factors contributing to the risk} were unclear, clinicians mentioned they would look for published references: ``what is the connection between {\ckd} and respiratory disorders?'' or ''how does celiac disease mention from the guideline answer, affect {\ckd}?''} \\ \midrule
      \multicolumn{1}{m{4cm}}{Support familiar categorizations} & \multicolumn{1}{m{10cm}}{Some clinicians were looking for more \textit{provenance around the categorization schemes} we were utilizing to show higher-level physiological pathways for diagnoses codes, and were also hoping for connections to familiar schemes like ``ICD-10'': ``how are hemorrhoids linked to the circulatory system?''  } \\ \bottomrule
\end{tabular}
\end{center}
\end{table}

More specifically, under `Theme 1: Clinical Value of Explanations and Contexts', we group instances where clinicians could make sense of the risk predictions and post-hoc explanations by the additional context provided, or instances where clinicians would appreciate more context. Within `Theme 2: Highlight Actionability,' we discuss instances where clinicians mentioned a need to depict actionable features and indicate actions for them concerning the patient's {\dm} diagnoses or their elevated {\ckd} risk. Under `Theme 3: Connections to Patient Data,' we cover instances where clinicians looked for connections to patient history or their lab results while reasoning about the patient case. Finally, under `Theme 4: Connections to External Knowledge,' we describe instances where clinicians mentioned a need to make connections to the latest literature, other medication databases, or other clinical schemes they utilize. In addition, these themes have an order among them, to address the clinical value of explanations from Theme 1 and the actionability aspects from Theme 2, the content requests from Themes 3 and 4 would be contributing features, which are the connections to patient data from Theme 3 and the links to external knowledge from Theme 4. We present deeper breakdowns in the form of sub-themes and descriptions for each of these four themes in Tab. \ref{tab:thematicanalysis1}, \ref{tab:thematicanalysis2}, \ref{tab:thematicanalysis3} and \ref{tab:thematicanalysis4}.

\begin{figure}[hbt!]
\centering
\includegraphics[width=1.0\linewidth]{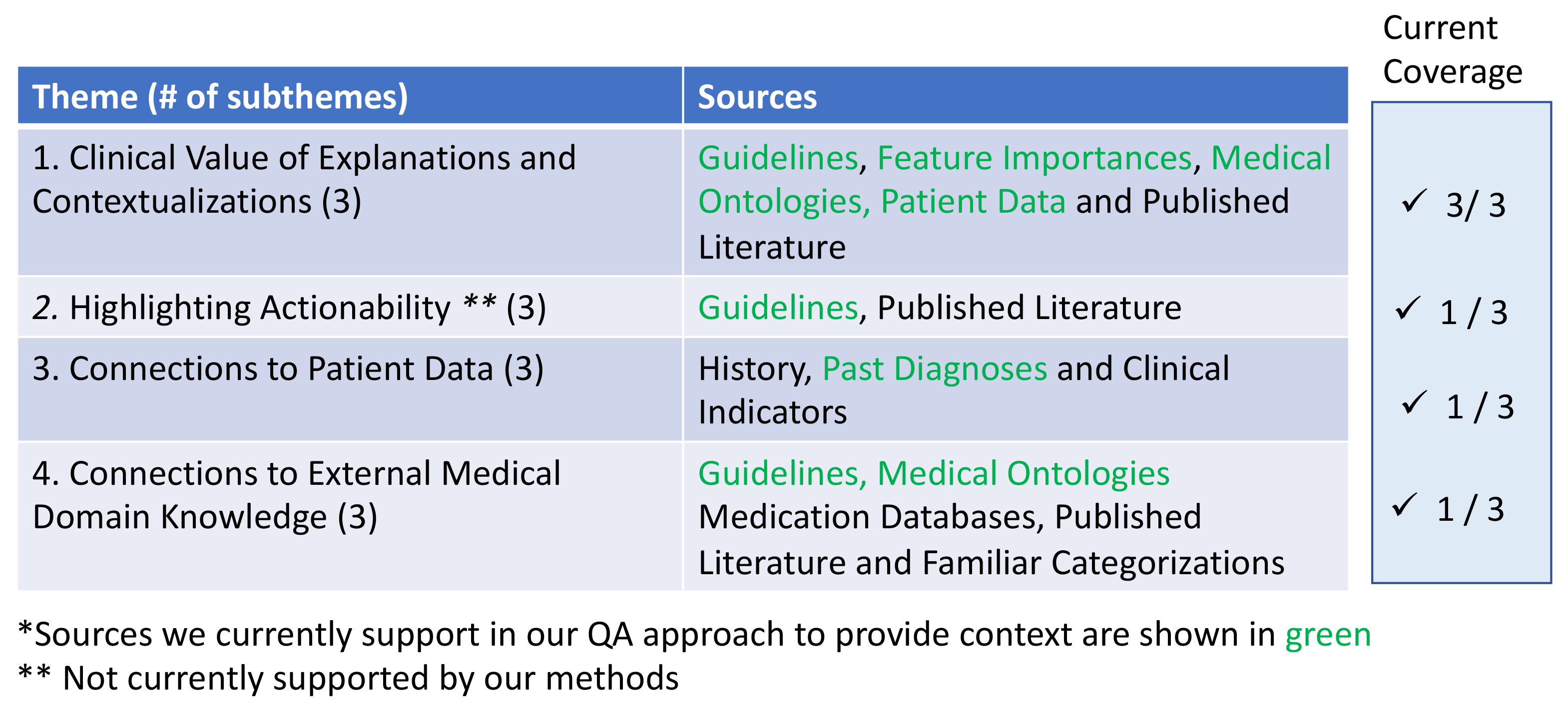}
\caption{Broad themes that emerged from conversations with clinicians while walking them through the results of our QA contextualization framework. Reproduced from: S. Chari, P. Acharya, D.M. Gruen, O. Zhang, E.K. Eyigoz, M. Ghalwash, O. Seneviratne, F.S. Saiz, P. Meyer, P. Chakraborty, D.L. McGuinnesss, ``Informing clinical assessment by contextualizing post-hoc explanations of risk prediction models in type-2 diabetes,'' \textit{Artif. Intell. in Med.} J., vol. 137, Mar. 2023, Art. no. 102498, doi: 10.1016/j.artmed.2023.102498.}
\label{fig:expertpanelthemes} 
\end{figure}

As the expert panel sessions were conducted mid-way through our current implementation, some of these themes served as a means for further refinements, such as features to support a `need for more related diagnoses,'  and `connections to patient's clinical indicators,' 
To address these themes, we refined different modules of our pipeline, including the user interface, the question-answering, and post-explanation modules. 
In future, to address the requests under `Theme 4: Connections to External Knowledge,' we plan to support connections to external resources, which clinicians may find valuable. The coverage of current supported sources to support the themes are reported in Tab. \ref{fig:expertpanelthemes}.

In summary, these themes and sub-themes from the expert panel sessions validate our hypothesis about the need for additional clinical context to situate risk predictions and span requests for better connections and presentations of domain knowledge that clinicians are familiar with in these settings.

\section{Discussion} \label{sec:qadiscussion}
In this section we analyze both the quantitative and qualitative results presented in Sec.~\ref{sec:results}. We analyze both the feasibility of supporting contextual explanations from authoritative sources such as CPGs, and the usefulness of providing contextual explanations from an analysis of the themes derived from our expert panel sessions. 

\subsection{Feasibility of Supporting Contextual Explanations}
\noindent \ul{Overall, how feasible it is to extracting contexts from guidelines? Which strategies are beneficial?} 
We have addressed $175$ clinically relevant questions that provide context around $20$ prototypical patients, their predicted risk, and the factors influencing their risk. We have implemented logical adaptations given what we know about the guideline data to improve the LLM model's capabilities and performance. These adaptations include knowledge augmentation from well-used medical ontologies like Metamap and SNOMED-CT to improve semantic overlap and rule augmentation to address numerical range questions. Our baseline LLM, BERT itself, has a variable performance that does well on some questions and not on others. 
Similarly, the best performances on the LLM + knowledge augmentation approaches varies across pre-filtering settings 3 and 4, that filter by Metamap disease codes and SNOMED-CT disease hops and post filtering-setting 5 that sorts by SNOMED-CT disease hops. 
 From our result evaluations, we see that the order of introducing the knowledge augmentation outputs impacts the accuracy scores, namely the MAP and recall. Mainly, pre-filtering the answer set before passing to a LLM can help it output more precise answers. In the best case, pre-filtering settings provide a gain of $4\%$ over the baseline LLMs both for disease and drug questions (BERT-KA from Tab. \ref{tab:nativemodelresults} and BERT-KA from Tab. \ref{tab:drugresults}). Similarly, post sorting the answers from a LLM can improve the recall, and in the best case (SciBERT-KA from Tab. \ref{tab:kmdiseaseresults}), we see a gain of $5\%$ from the baseline LLM. 

\vspace{1em}
 \noindent \ul{Can the extraction methods be improved?} 
Our result numbers also indicate that unsupervised adaptations can only reach a certain accuracy and point to the need for domain adaptations~\cite{singhal2023domain} to medical guidelines. With the continuous improvements in research in adapting LLMs to new domains, there are far more techniques for domain adaptation now, than when we implemented the contextualization pipeline. There can be further accuracy gains in applying fine-tuning techniques~\cite{xie2020distant} to train the newer generation of LLMs such as MedPALM~\cite{singhal2023towards} on our annotated QA corpus. We demonstrate how it is easier to use LLMs for a broader range of tasks in our MetaExplainer chapter (Chapt. \ref{chapt:Metaexplainer}).

\vspace{1em}
 \noindent \ul{Can the extraction methods be scaled?}
Since we were dealing with a setting with little or no annotations on the ADA 2021 CPG, we had to create our own annotations. Currently, we are dealing with a relatively small annotated corpora ($85$ questions and $654$ candidate sentences), and we consulted with a medical expert on our team to review these annotations. Even for these small corpora, we find that it is time consuming for a clinical expert (s) to review the annotations or create them. Hence, another interesting research direction, is on exploring techniques like weak supervision to scale and improve the coverage of the annotations. 

In summary, our guideline QA results depict incremental gains in adding knowledge and rule augmentations to enhance LLMs' performance and capabilities in domain applications and point to the need for supervised and semi-supervised approaches to improve these gains. We are, to the best of our knowledge, the first to report any QA performance numbers on the ADA CPG 2021 dataset. Additionally, we are the first few who have tried a LLM approach for more scalable upstream tasks on medical guidelines like QA than the current more time-consuming and dataset-dependent task of converting guidelines to rules and applying logical reasoning techniques over these rules~\cite{gatta2019clinical},~\cite{riano2019ten}.
Our approach to guideline extraction and QA (Fig. \ref{fig:guidelineqa}) is a step towards providing a more flexible way (~\cite{hematialam2021identifying},~\cite{hussain2021text}) to swap in guideline text from different diseases as needed. Our enhanced LLM approach (Fig.~\ref{fig:guidelineqa}) can be applied to any medical text corpus like medical guidelines extracted into a machine-comprehensible format and can address different question types (as seen in Tab.~\ref{tab:questiontypes}) relevant in risk prediction settings.

\subsection{Understanding the Added Benefit of the Derived Contexts}

\noindent \ul{What were the takeaways and feedback from clinicians about the supported contextual explanations?} The four major themes - \textit{Clinical Value of Explanations and Contextualizations, Highlighting Actionability, Connections to Patient Data, and Connections to External Knowledge Sources} - that we found during the expert panel interviews to evaluate our contextualization approach, mainly point to the overall value of supporting different types of contexts, both from literature and patient data, and the need to better present connections between these contexts. Many of the contexts the clinicians on our expert panel were looking for were around the post-hoc explanations of the factors contributing to the risk. This finding corroborates a recent study that reports that post-hoc explanations themselves are insufficient to provide reasoning that clinicians can interpret and act upon~\cite{GHASSEMI2021e745}, and also add to the well-accepted belief that risk scores are insufficient. 

\vspace{1em}
\begin{figure*}[!htbp]
\centering
\includegraphics[trim=0 0 0 30,width=\linewidth]{images/ThemeCoverage.png}
\caption{Summarizing the coverage of our current data sources to support the themes we found from our expert panel discussions. Reproduced from: S. Chari, P. Acharya, D.M. Gruen, O. Zhang, E.K. Eyigoz, M. Ghalwash, O. Seneviratne, F.S. Saiz, P. Meyer, P. Chakraborty, D.L. McGuinnesss, ``Informing clinical assessment by contextualizing post-hoc explanations of risk prediction models in type-2 diabetes,'' \textit{Artif. Intell. in Med.} J., vol. 137, Mar. 2023, Art. no. 102498, doi: 10.1016/j.artmed.2023.102498.}

\label{fig:themecoverage}
\end{figure*}
\noindent \ul{Do the supported data sources address the clinicians needs?} 
Through further analysis, we find that the contexts the clinicians were looking for and discussing can be addressed either by connections to patient history and data - patients' diagnoses, medications, and lab values - or through published literature. Specifically, we find that the different questions and question types (Tab. \ref{tab:questiontypes})
that we support from the {\dm} guidelines can address \textbf{$6$ of the $12$ sub-themes} (Fig. \ref{fig:themecoverage}), i.e., providing contextual information around patient's {\dm} state, their {\ckd} risk and the individual features (Theme 1), highlighting the impact of {\ckd} risk on treatment decisions for {\dm} (Theme 2), providing links to published articles (Theme 4), and showcasing connections to patient clinical indicators (Theme 3) where mentioned. Some other themes can be easily addressed by enabling connections from the {\ckd} risk scores and the features contributing to them, to patient timelines for diagnoses and lab values. Other themes - support for familiar categorizations (Theme 4) and the need for information on related diagnoses (Theme 3) - benefit from connections to medical ontologies that support either drill-downs to more specific diagnoses or abstracting up to higher-level pathways. We currently only support abstractions to higher level physiological pathways on the prototype dashboard (e.g., all disease of the circulatory system can be filtered from the patient's feature importances) and are investigating how to support drill-downs based off of these pathways more broadly.

\begin{figure*}[!htbp]
\centering
\includegraphics[trim=0 0 0 30,width=\linewidth]{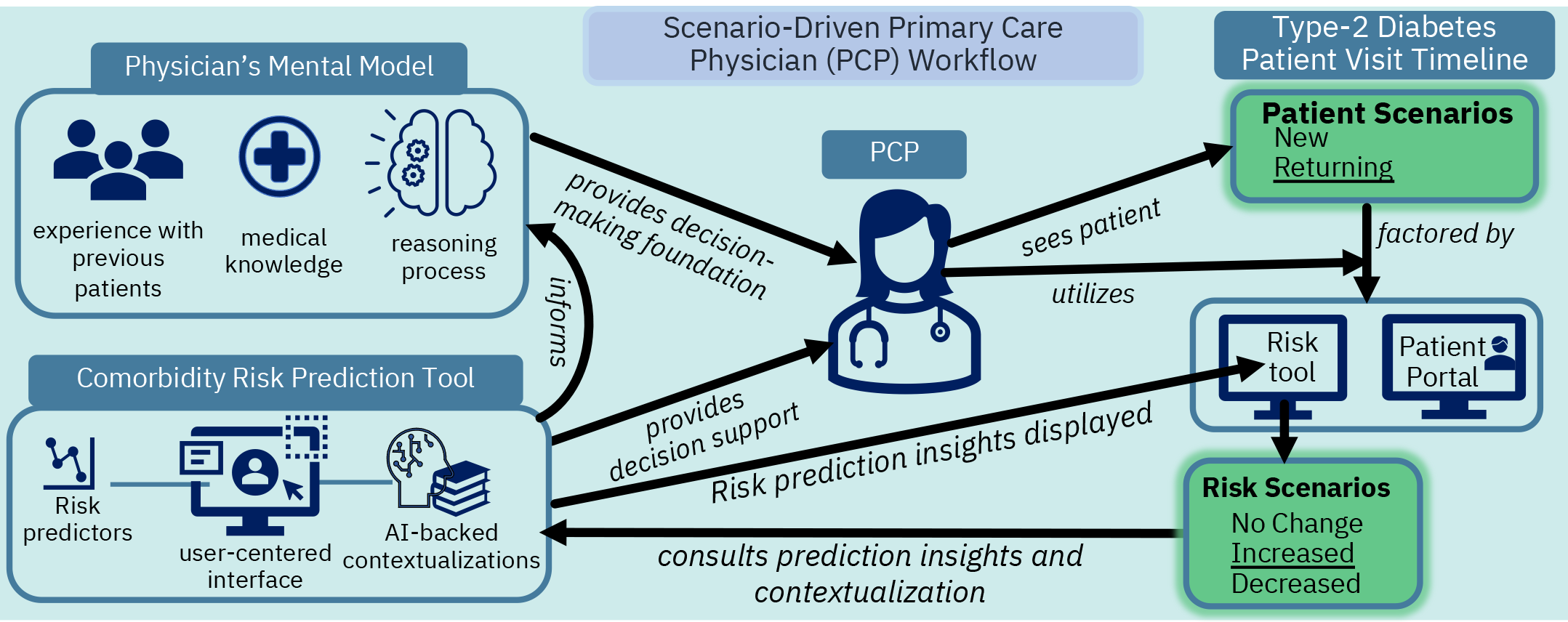}
\caption{Illustration of findings for a Primary Care Physician (PCP) - one target persona among clinicians - whose workflow is dependent on the patient context and the clinician-patient history. We show an example scenario where the predicted risk for a returning patient has increased since the last visit. This context and the PCP's mental model drive the PCP's following actions, such as differential treatment decisions made by probing reasons for the increased risk. Reproduced from: S. Chari, P. Acharya, D.M. Gruen, O. Zhang, E.K. Eyigoz, M. Ghalwash, O. Seneviratne, F.S. Saiz, P. Meyer, P. Chakraborty, D.L. McGuinnesss, ``Informing clinical assessment by contextualizing post-hoc explanations of risk prediction models in type-2 diabetes,'' \textit{Artif. Intell. in Med.} J., vol. 137, Mar. 2023, Art. no. 102498, doi: 10.1016/j.artmed.2023.102498.}

\label{fig:PCP-workflow-diagram-condensed}
\end{figure*}
\noindent \ul{What can be the impact of our contextual explanations beyond the comorbidity risk prediction setting?} 
To identify specific scenarios in which our contextualizations might be most impactful, we discussed with a clinical expert typical situations of {\dm} patient care by clinician type and patient characteristics  to understand where risk predictions 
could help clinicians improve patient care planning. 
While current literature~\cite{wang2019designing} on designing healthcare models points to a user-centered approach, from this understanding of clinician workflows, our discussion showed the importance of grounding such user centered work in specific clinical scenarios (see Fig~\ref{fig:PCP-workflow-diagram-condensed}). 
For example, it became clear that dashboards containing contextual explanations around risk prediction could be used by clinicians in different ways. For example, a clinician seeing a patient for the first time and/or for the first diagnosis would be interested in creating their mental model of the patient's diagnosis and understanding the causes for the risk, whereas a clinician seeing a returning patient with a previously established diagnosis (where clinicians might be more interested in understanding any changes to the  risk prediction over time and the effectiveness of various interventions). 
Hence, based on these understandings, we formalized our use case to provide contextual explanations to a PCP around the predicted risk of {\ckd} among new {\dm} patients at their first diagnosis.

Additionally, while we focus our approach in the risk prediction of {\ckd} among {\dm} patients, the contextualization approach can be applied to other comorbid risk prediction settings given access to authoritative guidelines in the disease area, and likewise, the themes that we analyze from our expert panel interviews are also general enough to be considered applicable in other disease settings. These themes indicate a larger need for AI systems to support insights from multiple data and knowledge sources and present them as actionable and contextual explanations~\cite{chari2020explanation} (also pointed out in the self-explanation scorecard from~\cite{mueller2021principles}). In summary, our approach is a step towards extracting clinically relevant context from different data and knowledge sources, including guidelines, patient data, and medical ontologies, and using these contexts to augment and explain answers to a list of clinically relevant questions that can help clinicians reason and interpret the risk predictions for patients.



\vspace{1em}
\noindent \ul{What are some future directions that emerge from the clinician discussions?} 
Some subthemes under \textit{Theme 2: Highlighting Actionability} provide future directions, highlighting actionable factors and suggesting specific actions to reduce {\ckd} risk are not currently addressed by our contextualization approach. These sub-themes require more investigation and development of methods to identify actionable, most relevant factors to {\ckd} risk. Another point which we observed is that some of the factors that the model picked up on are not covered by the {\dm} guidelines, and could either be factors only relevant to {\ckd}, or are those that are not considered to be well-known enough to be covered in position statements like CPGs. We are also investigating how to combine insights from multiple guidelines (also mentioned in~\cite{sittig2008grand}) and if that would be useful. In summary, while some of these themes provide validation for the modules we currently support in our multi-method approach to provide context, others offer directions for us to build towards, such as enabling connections to external medication databases, supporting temporality in post-hoc explanations of risk,
and efforts to better present answers in terms of relevance and actionability. We are also considering interviewing more user groups within the clinical domain to strengthen an understanding of where such a risk prediction tool would be most impactful.

Future steps would also include a practical study in a clinical setting to further assess the utility of our method. More work is needed to leverage our contextualization pipeline for a clinical study, including, as stated earlier in this section, implementing efforts to understand better the bounds of model performances~\cite{ghosh2022uncertainty} in our approach, including running experiments with more recent LLMs and adapting these LLMs to the ADA CPGs. In summary, several steps for improvement should be undertaken before our approach can be deployed on a broader scale. Furthermore, an assessment of the Food and Drug Administration's guidelines for LLM-based applications~\cite{fdaLLM} can be valuable. 

\section{Conclusion} \label{sec:qaconclusion}

Contextual explanations have been posited to be useful for clinicians for real-world usage of AI models. In this contribution, we have developed an end-to-end AI systems and studied the feasibility and usability of extracted contextual explanations from medical guidelines using state-of-the art QA methods. 
We have focused our study in a risk prediction use-case for {\ckd} among {\dm} patients and have conducted both quantitative and qualitative analysis.
Upon conversations with clinicians, we have selected three entities of interest in the risk prediction setting to provide contextual explanations along - the patient, their predicted risk, and the model explanations of their risk. 
Crucially, we have identified several themes covering the explainability needs of clinicians. The supported contextual explanations support some of these themes and thus improves clinician's confidence is using AI supported systems. We also found state-of-the art large language models to be effective in extracting such contexts, especially for certain disease sub-groups. While our results support the hypothesis that presenting contextual explanations to clinicians is both feasible and usable, the performance and requirement gaps points to the need for further research in this field.
For example, while we have considered three domain sources for the contexts in this paper, the themes from the expert panel interviews also indicate that there may be value to connecting to other sources, including extracting additional guideline details from tables and flowcharts, and potentially
involving multiple layers  of the evidence pyramid to include such sources as systematic reviews, randomized clinical trials, cohort studies, and expert opinions. 
Similarly, novel machine learning techniques such as weak-supervision or one-shot learning may be need to improve the quality of extracted contextual explanations. A combination of both may also enable other approaches such as `prompt engineering' whereby patient data can be used to seed the QA model questions and get richer response.
Future research could be directed at overcoming the aforementioned opportunities. Overall, by closely working with clinical experts and adopting inter-disciplinary approaches, from the use case crystallization and methods development, to the evaluation stages, we have shown the promise in supporting clinically relevant contexts to help clinicians better interpret risk prediction scores and their model explanations. 
 
\chapter{METAEXPLAINER: A METHOD TO COMBINE MULTIPLE EXPLANATIONS} \label{chapt:Metaexplainer}
In the pursuit of supporting conversational~\cite{lakkaraju2022rethinking}, diverse~\cite{mittelstadt2019explaining}, ~\cite{miller2019explanation} and knowledge-enabled user-centered~\cite{dey2022human} explainability, we propose the development of a MetaExplainer, an explanation framework capable of generating explanations for end-user questions. Specifically, we explore a research contribution in providing end-user explainability emerging from user-driven reasoning and conversations with the design and implementation of MetaExplainer. We address the following research issues, (1) How to build an extensible explainability framework that provides appropriate and useful explanations drawing from different data modalities, knowledge sources, and explainer methods? and (2) How can a MetaExplainer be expanded and evaluated to support evolving user-centered needs across representative use cases? 

\begin{figure}[hbt!]
\centering
\includegraphics[width=1.0\linewidth]{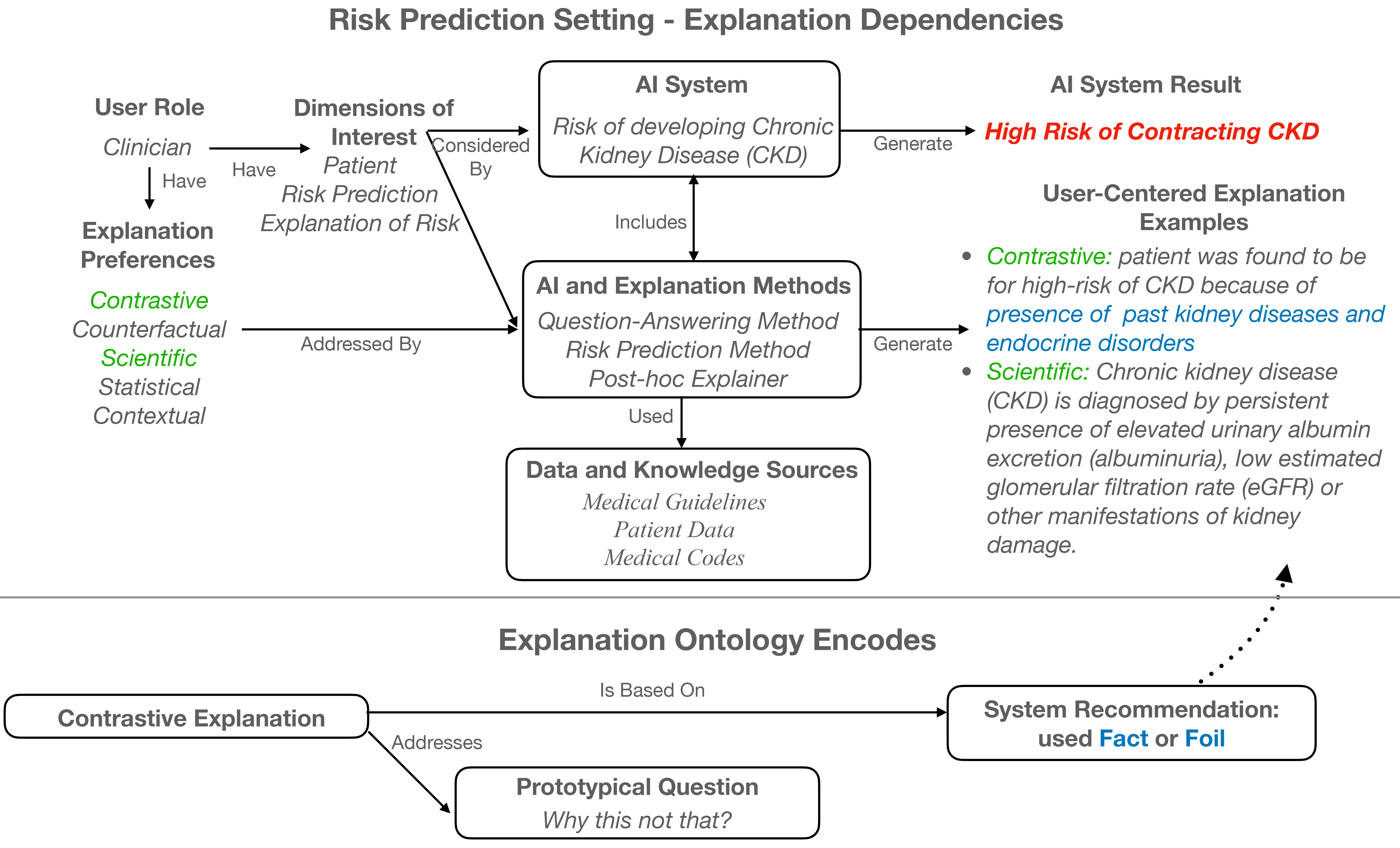}
\caption{Clinicians require different explanation types that are each supported by content from different data sources and AI methods. Also seen is how the EO encodes explanation dependencies that these user-centered explanations are built on. Reproduced from: S. Chari, P. Acharya, D.M. Gruen, O. Zhang, E.K. Eyigoz, M. Ghalwash, O. Seneviratne, F.S. Saiz, P. Meyer, P. Chakraborty, D.L. McGuinnesss, ``Informing clinical assessment by contextualizing post-hoc explanations of risk prediction models in type-2 diabetes,'' \textit{Artif. Intell. in Med. J.}, vol. 137, Mar. 2023, Art. no. 102498, doi: 10.1016/j.artmed.2023.102498.}
\label{fig:needforMetaexplainer}  
\end{figure}

In the previous two contributions, we demonstrated our ability to design a general-purpose representation for explanations and an implementation to support contextual explanations that provide context to help domain experts, such as clinicians, interpret model explanations. However, we don't yet have an implementation that would generate explanations along the various explanation types supported in the EO in response to user questions. Having the ability to see explanations along various types, can beneficial to end-users to help understand the AI system and its decisions from multiple perspectives~\cite{dey2022human},  ~\cite{wachter2017counterfactual}. For example, in the case of use of comorbidity risk prediction (Fig. \ref{fig:needforMetaexplainer}), both contrastive and scientific explanations could help clinicians reason through different dimensions when deciding what treatment option to prescribe to a patient. Hence, in the remainder of the thesis, we design and implement the MetaExplainer that would make these capabilities possible. 

\section{Overview} \label{sec:MetaExplainer-overview}
The design for the MetaExplainer is three-fold, i.e., we break it down into three different stages - Decompose, which converts the user question into an actionable machine interpretation that model explainers can address; Delegate, which runs the explainers identified to be capable of addressing the user question; and Synthesis, which populates natural-language explanations from the explainer outputs. Each stage addresses a specific goal towards generating natural-language explanations along one of EO's explanation types and builds on the previous stage's outputs as seen in the pseudocode \ref{lst:MetaExplainer}. We leverage and adapt a suite of AI methods to implement the MetaExplainer, and they are as follows: we elaborate on the methods in Sec. \ref{MetaExplainer:methods}.
\begin{itemize}
    \item In the Decompose stage, we leverage SOTA generational LLMs in conjunction with extracted prototypical questions from the explanation ontology (EO) for each explanation type. In this stage, the user question is augmented with parsed details as a reframed question, $rq$.
   \item  In the Delegate stage, we use relevant explainer methods from the IBM AIX-360 explainer toolkit~\cite{arya2019one} and other open-source repositories~\cite{macha2022rulexai},  ~\cite{mothilal2020explaining}. Here, the reframed question, $rq$, is further passed to identify filters on dataset features, if any, and the identified explanation type, $t$, is used to find suitable explainers $em$ to run to generate explainer outputs, $eq$. 
   \item In the Synthesis stage, we use LLMs' Retrieval Augmented Generation (RAG) capabilities to generate natural-language explanations by using EO-supported explanation templates as retrieval prompts against explainer outputs $eq$. The final output from this stage and the MetaExplainer is a natural-language explanation, $E$. 
\end{itemize}

We run evaluations for each stage using typically-used community metrics for the tasks; i.e., for the Decompose stage, we report F1 and exact match ratios of how the models were able to convert from a user question to the corresponding machine interpretation; in the Delegate stage, we report scores for the model explainers - i.e., faithfulness, monotonicity, etc. (see Tab. \ref{tab:modality-metrics}); and finally, for the Synthesis stage, we report the relevance of the generated explanations. We evaluate our MetaExplainer approach on two open-source tabular datasets, Diabetes (PIMA Indian), with plans to expand to another tabular dataset and potential datasets along the text and image modalities. 
Through our MetaExplainer approach, we demonstrate the utility of using an ontology to bootstrap prompts and templates for guiding LLMs to generate explanations along various types to a broad range of user questions about predictions, data, and knowledge used in application domains. 

\begin{algorithm} [hbt!]
    \caption{Pseudocode for MetaExplainer indicating the three stages - Decompose, Delegate and Synthesis, and their inputs and outputs.}
    \label{lst:MetaExplainer}
    \begin{algorithmic}[l]
		\Require Explanation Ontology (EO), Data Store (DS)
        \State Explanation type - explainer graph, $G^1 = \{(t_i, em_j),  \forall i \in N \text{ and } \forall j \in M\}$  
        \State Data type - explainer graph, $G^2 = \{(d_k, em_j),  \forall k \in K \text{ and } \forall j \in M\}$  \newline
        
        \Input
          \Desc{$uq$}{User Question}
        \EndInput  \newline
          
          \State $rq$ = \texttt{Decompose}($uq$) 
          \State $eq$ = \texttt{Delegate}($rq$) 
          \State $E$ = \texttt{Synthesis}($eq$) 
              
        \Output
          \Desc{$rq$}{list of questions reframed from $uq$}
          \Desc{$E$}{list of explanations $\{E_e\}$ that answer $uq$ where 
          $E_e = \{ {\explanation}_e, t_e, em_e, rq_e, uq_e\}$ } 
          \EndOutput
    \end{algorithmic}
\end{algorithm}

\section{Background} \label{MetaExplainer:background}
\subsection{Need for an Explanation Framework that Leverages the EO to Generate Natural-language Multi-type Explanations to User Question (s)}
From our previous work on contextual explanations for a comorbidity risk-prediction use case (Chapter \ref{chapt:qa_contextualization}) and our multiple conversations with clinicians to understand the need and utility of explanations~\cite{gruen2021designing}, we learned that explanations are composed of different knowledge sources and explainer methods. Additionally, model explainers or explainers that explain AI decisions typically provide scores, features, rules, or instances, which on their own can be overwhelming for end-users to interpret, lack the grounding in the domain knowledge or context and are often misleading on their own~\cite{ghassemi2021false}. Hence, end-users, specifically domain experts, prefer explanations that provide answers to a wide-range of user questions such as the Why, Why not, What ifs, What cases, etc~\cite{liao2020questioning},  ~\cite{liao2022connecting}. In the EO (Chapter \ref{chapt:explanation_ontology}), we capture the compositional needs and dependencies of fifteen user-centered explanation types that address these different set of user question types; however the EO implementation is more suited for system developers who want to represent various explanations in their use cases and are looking to structure their data and method outputs to enable explainability. Conversely, domain experts have a deep knowledge of the use cases but less so of the AI system, and either want to probe the decisions made by the system to understand its reasoning or learn more about the accuracy and trustworthiness of the decisions or globally learn about the data distributions the system operated on, or about global behavior. Hence, a system such as the MetaExplainer that can provide answers in the form of explanations for model decisions in various use cases is useful. The development of such a system introduces challenges of scalability in terms of minimal effort to load for new use cases, generalizability across wide-range of use cases and interoperability with existing explainers. In our current implementation of the MetaExplainer, we prioritize these attributes and develop a modular architecture that can be spun up with minimum human intervention, improved, and adapted at each stage.
\subsection{Explanation Ontology}
In the Explanation Ontology (EO), we model the system-, user- and interface- dependencies of explanations of AI systems. The EO model helps represent fifteen user-centered explanations, which were introduced, when explainability was identified as an essential prong of trustworthy AI. During this time, several researchers studied end-users and surveyed literature, finding that explanations serve different purposes for end-users and address various questions~\cite{wachter2017counterfactual},  ~\cite{miller2019explanation}, ~\cite{doshi2017accountability}. Hence, suggesting that there is an inherent diversity in the types of explanations. However, the descriptions of various explanation types were mainly contained in several papers and were difficult to support without a unified representation such as the EO. We designed the EO with the primary goal of representing explanations such that they can be constructed from the dependencies. Since, the first version of the EO was released in 2020, we have made updates in version 2 to support model explainers and their outputs such that the explanations can be built from explainer methods that generate them. While in both our previous EO papers~\cite{chari2023explanation}, ~\cite{chari2020explanation}, we have demonstrated how explanations can be inferred to be of particular types based on running logical reasoners over EO-represented knowledge graphs, in neuro-symbolic systems, it would be ideal also to have the EO integrate with other neural and ML methods directly to generate explanations. In the MetaExplainer, we explore how the EO can serve multiple purposes to support the generation of natural-language explanations, including identifying what explanation type would best address a user question, what explainer methods are capable of providing explanations for the identified type, and finally, what templates are suggested to structure explanations of the identified type. We use the EO as a knowledge base to provide explanation related content within the MetaExplainer. Further, we have made changes to make the EO more code-ready while developing the MetaExplainer, and plan to release a version three of the EO along with the MetaExplainer.

\subsection{Model Explainers}
Model explainers or explainer methods provide model explanations of AI systems, and they typically explain model behavior on a per prediction level (LIME~\cite{ribeiro2016should}, SHAP~\cite{lundberg2017unified}, etc.) or globally about the model's functioning (Boolean Rule Column Generator~\cite{dash2018boolean}). Several model explainers have been released, and they provide explanations in different forms ranging from representative instances on the dataset that models run, rules describing the patterns that the model focused, feature importances of features that influenced decisions, and less frequent descriptions of data distributions, preprocessing steps, and relevant literature. The model explainers are typically applied post-hoc after the training models have been run, or the training models are inherently explainable (Decision trees) or as models that work with intermediate representations of the models~\cite{arya2019one}. 

There are several challenges with using these model explainers directly in use cases to support end-user explainability. The outputs of these model explainers have often been found to be sufficient or lacking in domain context for end-users such as domain experts to take action on~\cite{ghassemi2021false, tonekaboni2019clinicians}. Further, running the model explainers is better suited for system developers who understand the functioning of AI systems. Additionally, it is hard to decide on what model explainer to run and how to present the results to domain experts; hence, encoding or registering the model explainers in EO by what explanations they best support and explaining the model explanation outputs in natural language via EO explanation types can be beneficial. Further, another benefit of presenting model explanation outputs as natural-language~\cite{slack2023explaining} is that they can be adapted to different user groups within the same domain~\cite{dey2022human} (e.g., clinical researchers prefer data explanations over clinicians who prefer scientific explanations). Additionally, combining the outputs of multiple model explainers~\cite{krishna2022disagreement} can provide better views of the AI system, which can help end-users learn different perspectives of the system and decide if they should trust or take action on a decision or not. In the MetaExplainer, we include a well-cited set of explainers mainly from IBM's AIX-360 toolkit~\cite{arya2019one} and other sources, with provisions and directions on how to support more. 

\section{Methods} \label{MetaExplainer:methods}
\begin{figure}
    \centering
    \includegraphics[width=1.0\linewidth]{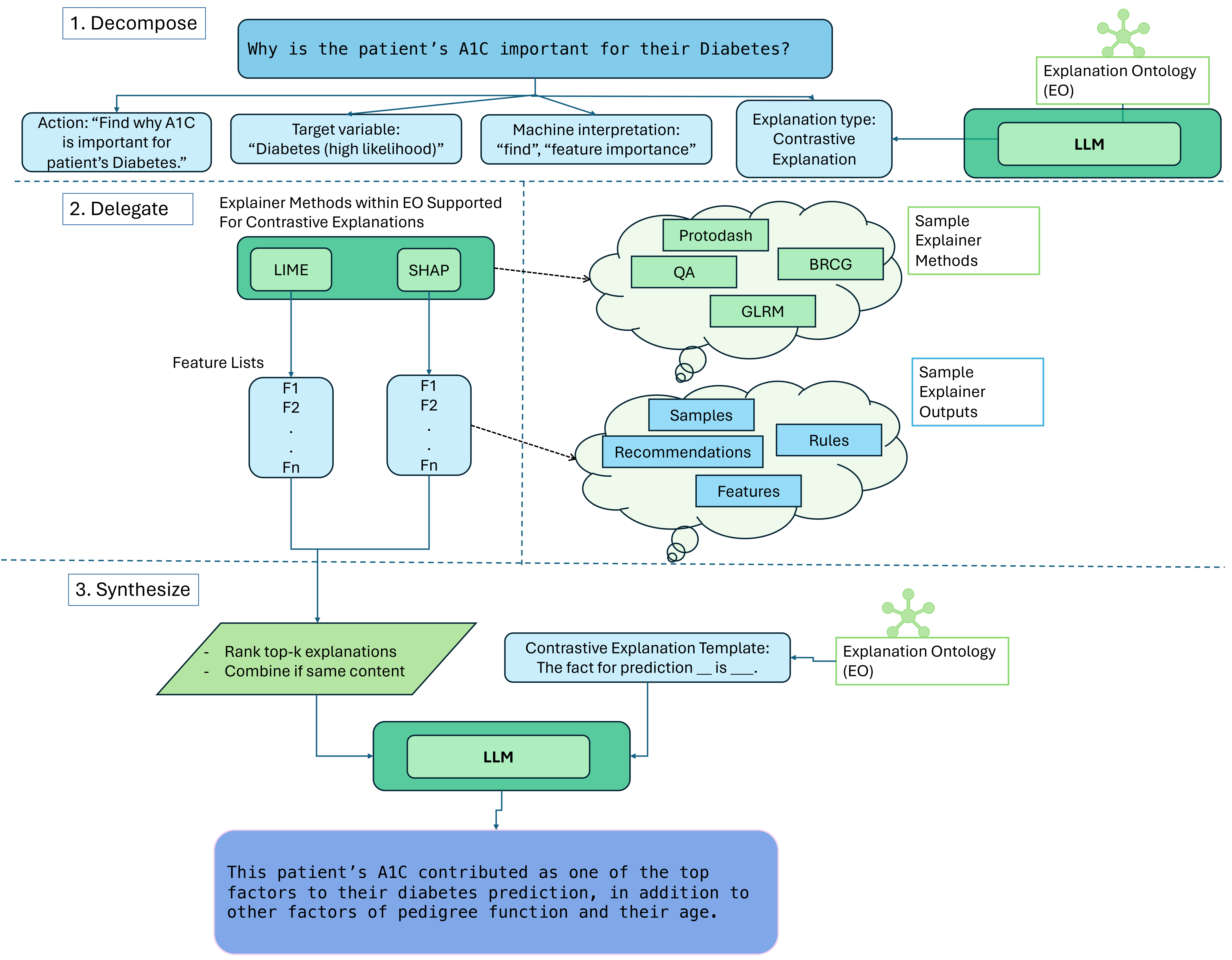}
    \caption{Workflow diagram of MetaExplainer, highlighting the different input and output streams. Methods are indicated by green boxes and all other data components are shown in blue.}
    \label{fig:MetaExplainer-workflow}
\end{figure}

As discussed in Sec. \ref{sec:MetaExplainer-overview} and \ref{MetaExplainer:background}, the MetaExplainer addresses a gap in user-centered explainability where users can flexibly ask questions for which they seek explanations in use cases where AI systems are utilized. The framework provides explanations along one of the EO-represented explanation types. We do so by adopting a modular approach to \textit{decompose} the user question into a machine interpretation that can then be \textit{delegated} to explainer methods registered against explanation types in EO. Finally, the explainer outputs are \textit{synthesized} into natural-language explanations presented to the end-users. We implement the MetaExplainer framework as a Python package, such that each of the stages can call methods and classes from one another, and all of the stages write outputs to the same data folder, hence enabling the stages to build off of the intermediatory stage output as seen in Fig. \ref{fig:MetaExplainer-workflow}. For example, the question in Fig. \ref{fig:MetaExplainer-workflow} ``Why is the patient's A1C important for their Diabetes?'' is identified to be of type - Contrastive explanation and the interpretation along with the identified explanation type are parsed in the delegate stage to run the SHAP explainer~\cite{lundberg2017unified}, a contrastive explainer method registered in the EO. Finally, the outputs of SHAP are passed on to the synthesis stage to construct a contrastive explanation addressing the user question about the importance of A1C. We evaluate the output of each stage since each stage performs different tasks and needs to be assessed differently. We have made the MetaExplainer code available under the open-source MIT license on Github: \url{https://github.com/tetherless-world/MetaExplainer}. In the rest of the section, we describe the technical details of the three stages of the MetaExplainer framework - Decompose, Delegate and Synthesis.

\subsection{Decompose} \label{sec:MetaExplainer-decompose}
In the Decompose stage, the objective is to generate a machine-actionable parse of a user question (s), including the explanation type that best addresses the question and the features being asked about. For example, in a question - ``Why is a 60-year-old woman with a BMI of 28 more likely to have Diabetes?,'' we would want to identify that this question is best addressed by facts in contrastive explanations and has filters applied on features - age, BMI. From the question, we could also infer that the patient's likelihood of Diabetes is high. These attributes, if captured, help ensure that the explainer methods can reliably address the user question. Hence, in the Decompose stage, we set up a task to convert a user question $uq$ into a reframed question $rq$, which is a tuple composed of $\{$ question, explanation type, machine interpretation, action $\}$ (see Listing \ref{lst:decompose}). Our reframed question improves upon the parsed utterance structure from Megaexplainer~\cite{slack2023explaining}, and helps us generate explanations for identified explanation types and filters from fields in $rq$ such as explanation type and machine interpretation.

\begin{algorithm} 
  \caption{\texttt{Decompose}: decompose user questions into prototypical questions along EO and resolve explanation types} \label{lst:decompose}
	\begin{algorithmic}[1]   
	   \Require Explanation Ontology (EO), Data Store (DS)
          \State Explanation types, $t = \{t_i, \forall i \in N\}$
		  \State Prototypical Questions, $qt = \{qt_a, \text{ s.t. } \exists t_a \in t \text{ associated with } qt_a\}$

          \Input
             \Desc{$uq$}{User Question}
             \Desc{$qb$}{Question bank}
          \EndInput \newline 

            \For {each domain} \Comment{\textbf{Prepare $qb$ to fine-tune LLM for inferences}}
          \State  Generate question bank, $qb$ of $uq$:$rq$ pairs for domain, from:
                \State - features from domain dataset (s), $ds$ from DS 
                \State - prototypical questions examples, $qt$, for each $t$ 
                \State Fine-tune LLM on $qb$ 
        \EndFor
          
          \For {Each $uq$} \Comment{\textbf{Decompose $uq$ into $rq$}}
                \State Append $uq$ to LLM prompt
                \State Pass through LLM to infer $rq$
                \State Parse LLM's top-1 response to populate $rq$ with - (question, action, machine interpretation, explanation type) tuple
                \EndFor
            \Output
            \Desc{$rq$}{List of reframed questions with types}
          \EndOutput
    \end{algorithmic}
\end{algorithm}

We experimented with using traditional semantic and syntactic parse approaches to generate the $rq$ from $uq$. However, we found that these methods are cumbersome to implement, as our task is both an entity and relation extraction problem, and we are in an unsupervised setting where there are typically few or no $uq$:$rq$ training pairs. Hence, we decided to leverage LLMs, which have been shown to do well with a few examples (few-shot learning)~\cite{cahyawijaya2024llms, ahmed2022few} and can be trained to generate responses in various formats. To fine-tune the LLMs to generate $rq$ for $uq$, we constructed question banks $qb$. These question banks are generated on a per-dataset basis per domain. The questions in the question bank are built from combinations of dataset features and possible value ranges along the question types or prototypical question ($qt$) for each explanation type supported in the EO (e.g., counterfactual explanations address What if questions). 

As indicated in the pseudocode for Delegate in the list \ref{lst:decompose}, we first build a question bank $qb$ of $uq$: $rq$ pairs to help fine-tune LLMs to generate $rq$ for new user questions, $uq$. We construct this $qb$ in a weak-supervised fashion. As mentioned earlier, we use the prototypical questions mapped for each explanation type in EO (e.g., For contextual explanations - What broader information about the current situation prompted the suggestion of this recommendation?) as a guide to generate a certain number of questions for each explanation type on a per-dataset basis. Instead of constructing these $ uq$: $rq$ pairs entirely by hand, we leverage the GPT family of LLMs~\cite{floridi2020gpt}, specifically GPT 3.5 Turbo, (Listing \ref{lst:gpt-prompt}) to generate these pairs, and we then verify these pairs for quality and validity. Several well-used foundation models are trained in similar weak-supervised or unsupervised fashions wherein LLMs are augmented with aids of high-quality data to generate a few examples to fine-tune other LLM models~\cite{li2024llava}. Also, when MetaExplainer is run for a new use case, a domain expert will have to verify these generated pairs. 

\begin{figure}[hbt!]
\centering
\includegraphics[width=1.0\linewidth]{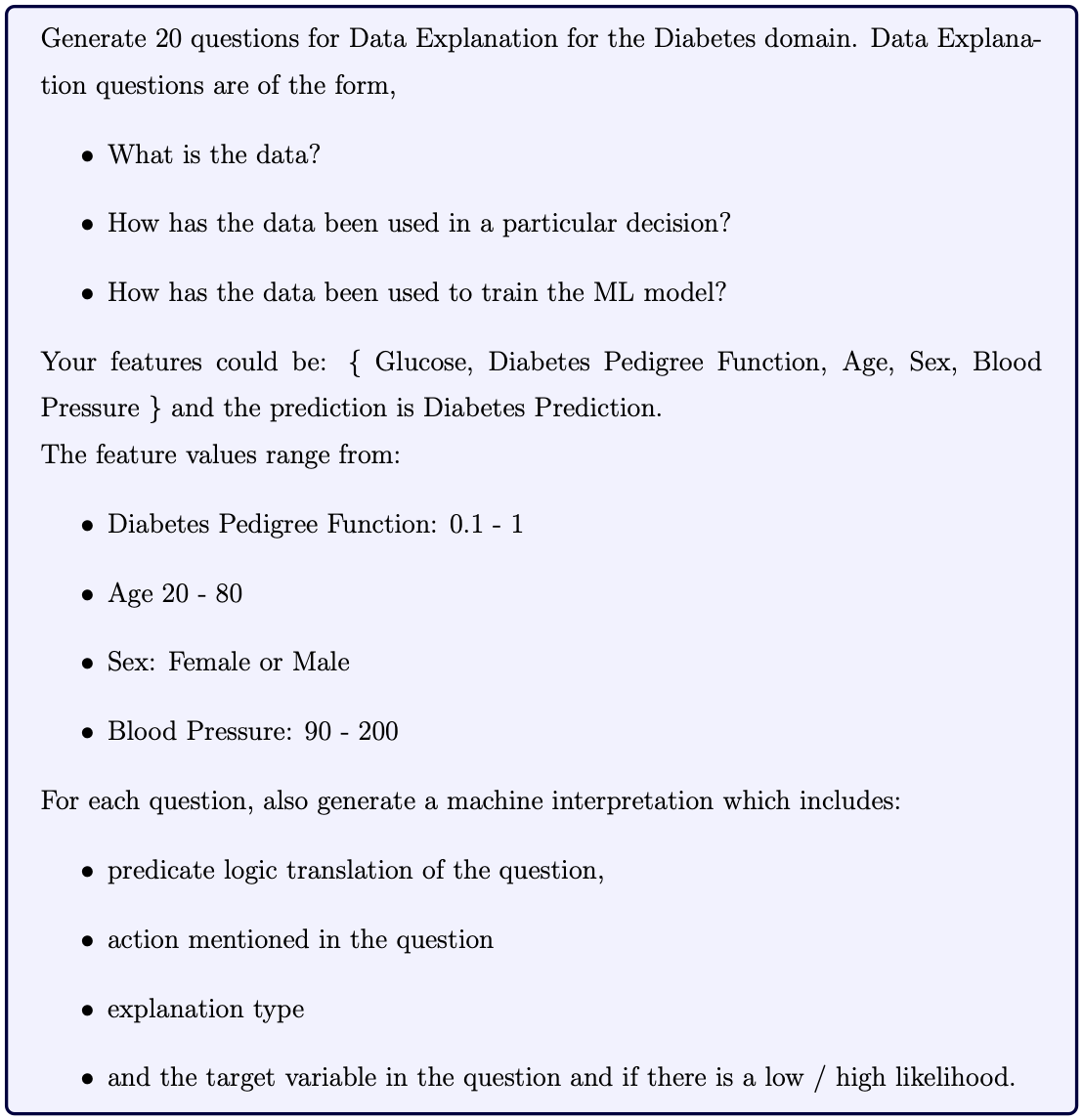}
\caption{Prompt provided to GPT-3.5 Turbo model to generate questions for data explanations to populate question bank, $qb$.}
\label{fig:prompt1}  
\end{figure}

The verification is done by looking over $uq$:$rq$ spreadsheets for each explanation type. In the verification phase, we mainly edit the predicate logic parses of $rq$. We find that the LLMs are generally good at performing identification tasks (i.e., identifying the explanation type) or extracting content directly from the question (i.e., actions and likelihoods) but find it more challenging to generate logical parses, i.e., predicate logic translation of the question, $uq$. Furthermore, we find that these logical translations do not conform to one pattern, and hence, we handle this ambiguity in the predicate logic translations in the next Delegate stage.

\begin{figure}[hbt!]
\centering
\includegraphics[width=1.0\linewidth]{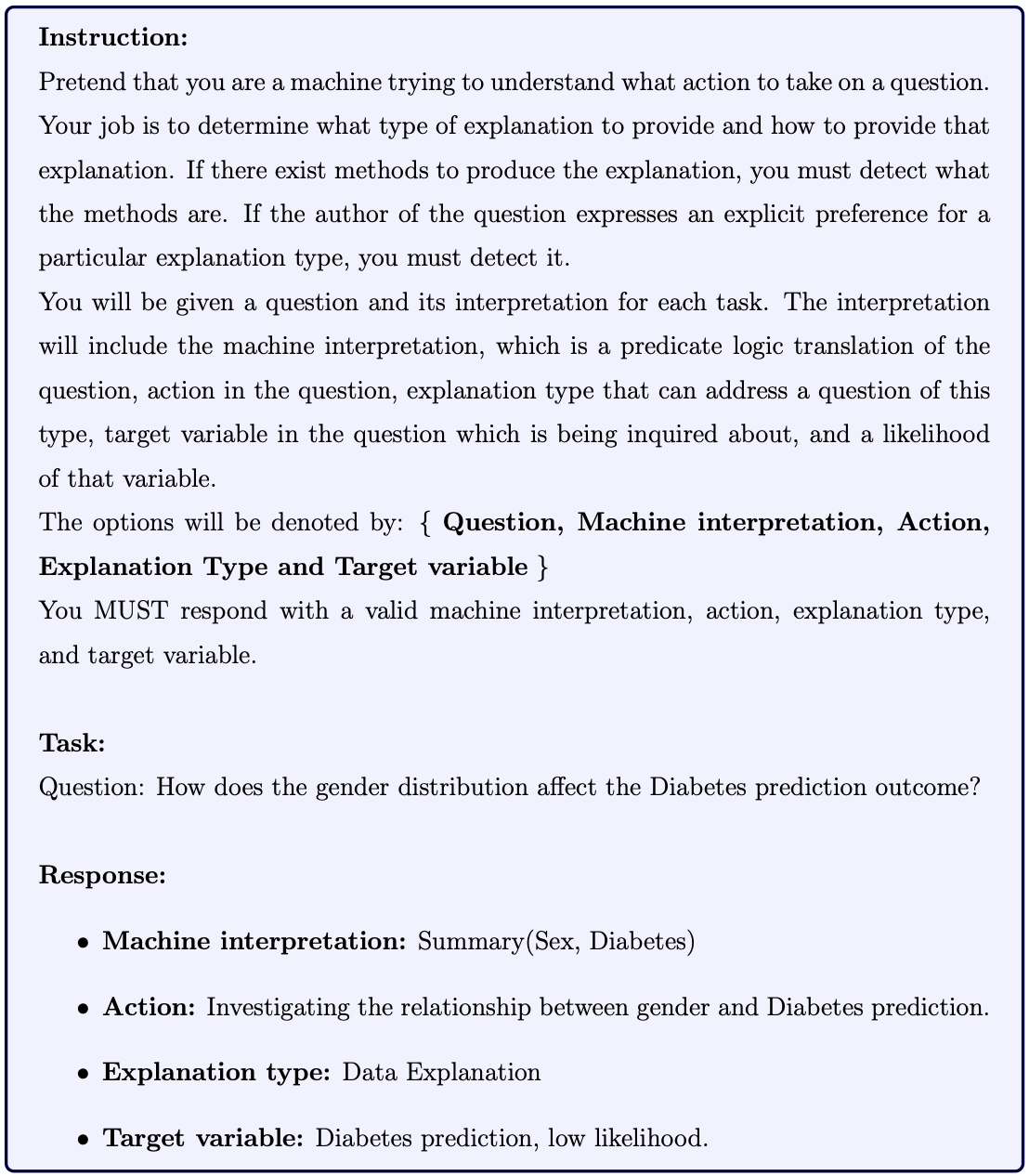}
\caption{Instruction prompt example used for fine-tuning Llama models to decompose question $uq$ into response, $rq$.}
\label{fig:prompt2}  
\end{figure}

Upon verifying the GPT generated $uq$:$rq$ pairs, we fine-tune open-source large LLMs such as Llama2 and Llama3~\cite{touvron2023llama} available on the HuggingFace portal, to generate $uq$:$rq$ pairs for new questions. We fit the $uq$:$rq$ pairs into user: response style (see Listing \ref{lst:decompose-prompt}) instruction prompts~\cite{taori2023stanford} and use Low-Rank adaptations~\cite{hu2021lora} and Supervised Fine-tuning Trainers (SFTTrainer)~\cite{gunel2020supervised} to fine-tune the Llama models. We reserve $20\%$ of the verified $uq$:$rq$ pairs for test inference, and we report the F1 and exact match scores~\cite{schneider2024evaluating} for the $rq$ generated by the fine-tuned Llama models~\cite{wei2021finetuned} (Sec. \ref{sec:MetaExplainer-results}). We process the output from the fine-tuned Llama models to construct $rq$ tuples that can be parsed by Delegate to run explainer methods.

\subsection{Delegate} \label{sec:delegate}
In the Delegate stage, the objective is to execute relevant explainer methods to address a user question. Since it is difficult to directly identify which methods to run and what filters to apply from the user question $uq$, we implemented the Decompose stage to generate a reframed question, $rq$, that contains fields that can help identify which explainer method to run and what actions to perform. Additionally, within the EO, we capture against explanation types explainer methods that could help generate them (e.g., Contrastive explanations can be based on local explanation outputs from Saliency methods~\cite{arya2019one}). Hence, the Delegate stage combines a parsing task, explainers ($em$) execution task, and explainer outputs ($eq$) processing tasks (see Listing \ref{lst:delegate}). In the parsing task, we parse the machine interpretation field of the $rq$ for filters on features, the explanation type ($et$) field to identify what explainer methods ($em$) can be run, and the action field to identify if there are other actions the question asks about (e.g., preprocessing, accuracies, etc.) In the execution task, we run open-source explainers available on Github. Finally, in the output processing step, we apply additional filters to the explainer outputs and persist them uniformly as data frames that the Synthesis stage can read to include in natural-language explanations of the explanation type. 

\begin{algorithm}  [hbt!]
  \caption{\texttt{Delegate}: delegate and retrieve answers from explainer methods along different explanation and data types} \label{lst:delegate}
	\begin{algorithmic}[1] 
		\Require Explanation Ontology (EO), Data Store (DS)
        \State Explanation types, $t = \{t_i, \forall i \in N\}$
		\State Explainers, $em = \{em_j, \forall j \in M\}$
        \State Explanation type - explainer graph, $G^1 = \{(t_i, em_j),  \forall i \in N \text{ and } \forall j \in M\}$  
        \State Data type - explainer graph, $G^2 = \{(d_k, em_j),  \forall k \in K \text{ and } \forall j \in M\}$  \newline

          \Input
             \Desc{$rq$}{List of reframed questions from user question}
          \EndInput  \newline
       
		\State Set $eq$ to NULL   
            \For {Every $rq_a$ in $rq$}
                
                \State Get $t_{a}$ from  $rq_a$
                    \State Set $eq(t_a)$ to NULL if $t_{a}$ doesn't exist
                    \State Fetch all \textit{explainers} $\lbrace\{em\}_a, \{em\}_a \subset em\rbrace$ from $G^1$, for $t_a$

                    \State Get feature groups, $fg_{a}$ from $rq_a$
                    \For {Every $em_j$ in $\{em\}_a$} \Comment{\textbf{Store O/P of explainers against ($fg$, $et$, $em$)}}
                        \State Fetch all \textit{data type} $\lbrace\{d\}_j, \{d\}_j \subset d\rbrace$ from $G^2$, for $em_j$
                        \State $\forall d_k \in \{d\}_j$, update $eq(t_a) = eq(t_a) \cup \{em_j(rq_a, d_k\}$ 
                    
                \EndFor
            \EndFor \newline

             \Output
          \Desc{$eq$}{Explainer outputs stored for each mapped explanation type}
          \EndOutput
            	\end{algorithmic} 
\end{algorithm}

The parsing task of the Delegate stage involves identifying feature groups, which are a combination of {feature: value} pairs within the machine interpretation portion of the reframed question, $rq$. The feature groups could also be a combination of multiple filters or feature combinations, e.g., 60-year-old female patient - (age: 60, sex: female) within the same feature group. Further, the features could be expressed differently in the predicate logic machine interpretations depending on how they were generated by the fine-tuned Llama LLMs. Hence, we had to handle the different variations and combinations of feature groups in our parsing module. To reduce the variance in parsing, we implement a rule-based regular expression-dependent parser to extract various combinations of feature pairs typically encapsulated within round braces `(', but could also occur as feature = value standalone pairs. Our parser also captures additional actions before the feature groups and maps them to be features if the actions happen in the feature column set of the dataset (s) used within the use case. Finally, we pass on the explanation type, $et$, to the explainer identification stage of Delegate. A sample parse of the reframed question, $rq$, can be viewed from Tab. \ref{tab:delegate-parse}.

\begin{table}[hbt!] 
\centering
\caption{Parser output for reframed question, $rq$, that includes the identified feature groups, actions and explanation type.}
\label{tab:delegate-parse}
\begin{tabular}{|l|p{10cm}|} 
\hline
\textbf{Question} & How did the model justify predicting Diabetes for a 45-year-old female with a BMI of 27 and a Diabetes Pedigree Function of 0.2? \\
\hline
\textbf{Machine Interpretation} & Predict(Diabetes, Age = 45, Sex = Female, BMI = 27, DPF = 0.2) \\
\hline
\textbf{Action} & Predict \\
\hline
\textbf{Feature Groups} & \{ \textbf{Diabetes}: `', \textbf{Age}: 45, \textbf{Sex}: `Female', \textbf{BMI}: 27, \textbf{Diabetes Pedigree Function}: 0.2 \} \\
\hline
\textbf{Explanation Type} & Rationale Explanation \\
\hline
\end{tabular}

\end{table}

The \textit{execution task} of the Delegate stage involves executing explainer methods that are known to provide explanations for the identified explanation type, $et$, in reframed question, $rq$. We retrieve these explanation methods from EO using a SPARQL query (Listing \ref{lst:sparqlqueryexplanationmethod}) using the RDFLib library, that extracts the model explainers from equivalence condition restrictions defined against each explanation type. Several explainer methods can generate rules, instances, or features as evidence for or against ML decision outputs or general model behavior. As mentioned in Sec. \ref{MetaExplainer:background}, several explainers are open-sourced and can be applied to custom use cases. These explainers are typically run on trained models and their learned weights or predictions; hence, we also identify the best-performing model on our tabular datasets and run the explainers on these models. We reuse ML models available in the widely-used Scikit-learn library~\cite{pedregosa2011scikit} (e.g., Logistic Regression, Decision Tree, and Random Forest) and run them on our curated tabular datasets. 

\begin{figure}[H]
    \centering
\lstset{language=SPARQL, basicstyle=\ttfamily\fontsize{9}{10}\selectfont, xleftmargin=5mm, framexleftmargin=5mm, numbers=left, stepnumber=1, breaklines=true, breakatwhitespace=false, numbersep=5pt, tabsize=2, frame=lines}
\begin{lstlisting} 
PREFIX rdfs:<http://www.w3.org/2000/01/rdf-schema#>
PREFIX owl:<http://www.w3.org/2002/07/owl#>
PREFIX ep: <http://linkedu.eu/dedalo/explanationPattern.owl#>
PREFIX eo: <https://purl.org/heals/eo#> 
PREFIX sio: <http://semanticscience.org/resource/>
PREFIX prov: <http://www.w3.org/ns/prov#>

select ?taskObject where {
?class (rdfs:subClassOf|owl:equivalentClass)/owl:onProperty ep:isBasedOn .
?class (rdfs:subClassOf|owl:equivalentClass)/owl:someValuesFrom ?object .
?object owl:intersectionOf ?collections .
 ?collections rdf:rest*/rdf:first ?comps .
?comps rdf:type owl:Restriction .
?comps owl:onProperty sio:SIO_000232 .
?comps owl:someValuesFrom ?taskObject .
?class rdfs:label "Rationale Explanation" .
}

\end{lstlisting}
\caption{A SPARQL query run to retrieve explainer methods for explanation types.}
    \label{lst:sparqlqueryexplanationmethod}
\end{figure}

While we support fifteen user-centered explanation types in the EO (Chapt. \ref{chapt:explanation_ontology}), not all have corresponding model explainer implementations yet (e.g., no known explainer methods exist to support safety and performance explanations). In the first version of the MetaExplainer, we have chosen only to include explanation types for which we have identified explainer methods that can generate them, and the six explanation types we support are indicated in Tab. \ref{tab:MetaExplainer-delegate}. Additionally, we have not found a generic implementation for contextual explanations; hence, we do not yet support these types of explanations - we could consider repurposing our guideline contextualization framework (Chapt. \ref{chapt:qa_contextualization}) for extracting contexts from published literature. Hence, it can be considered that we support five explanation types in our current implementation of the MetaExplainer with plans to support more as explainer methods become available (Sec. \ref{sec:MetaExplainer-discussion}).

\begin{table}[hbt!]
\centering
\caption{Explainer methods registered within the MetaExplainer for the currently supported user-centered explanation types from the EO.}
\label{tab:MetaExplainer-delegate}
\begin{tabular}{|l|l|l|}
\hline
Explanation Type & Explainer Method     & Output Type            \\ \hline
Case-based       & Protodash~\cite{gurumoorthy2019efficient}    & Instances                    \\ \hline
Contextual       & None     & Data points                        \\ \hline
Contrastive      & SHAP~\cite{lundberg2017unified}     & Features                    \\ \hline
Counterfactual   & DiCE~\cite{mothilal2020explaining} & Instances                            \\ \hline
Data             & Protodash, Data Analysis Methods & Instances and Features\\ \hline
Rationale        & RuleXAI~\cite{macha2022rulexai}  & Rules                        \\ \hline
\end{tabular}
\end{table}

\begin{table}[ht!]
    \centering
    \caption{Set of rules generated for patient from Tab. \ref{tab:delegate-parse}.}
    \label{tab:rules_summary}
    \begin{tabular}{ll}
        \toprule
        \textbf{Rule} & \textbf{Label} \\
        \midrule
        IF DiabetesPedigreeFunction = (-inf, 0.22) & THEN label = \{0\} \\
        IF Glucose = ($<168.5$, inf) & THEN label = \{0\} \\
        IF BMI = (-inf, 28.25) & THEN label = \{0\} \\
        IF Glucose = (-inf, 123.5) AND BMI = (-inf, 40.25) & THEN label = \{0\} \\
        IF BloodPressure = ($<97.0$, inf) & THEN label = \{1\} \\
        IF Insulin = ($<113.5$, inf) & THEN label = \{1\} \\
        IF BMI = ($<42.05$, inf) & THEN label = \{1\} \\
        IF Glucose = ($<123.5$, 168.5) & THEN label = \{1\} \\
        \bottomrule
    \end{tabular}
    
\end{table}

Upon identifying the explainer to run based on the identified explanation type for $uq$, we pass filtered subsets of the tabular dataset with the feature group restrictions, $fg$, applied to the explainers. If there are no direct matches for the restrictions, we retrieve the closest records whose feature values match the restrictions by leveraging closest match queries from the Pandas library. For example, if there is no 60-year-old patient in the dataset - we show results for a 58-year-old patient if they exist). Further, there could be several feature group restrictions; hence, we run explainers for each feature group combination. An example of rules identified by RuleXAI~\cite{macha2022rulexai} explainer method, for patient from parsed $rq$ (see Tab. \ref{tab:delegate-parse}) can be seen in Tab. \ref{tab:rules_summary}. We evaluate the individual outputs of the explainer methods by the metrics recommended in the literature~\cite{zhou2021evaluating} for the explanation modalities that the methods output (see Tab. \ref{tab:modality-metrics}).

In the output processing step of Delegate, we persist the explainer outputs along with the provenance of feature groups and associated actions, such that in the synthesis stage, these outputs can be reliably combined. This is important because various groups of explainer outputs could address different portions of a user question, $uq$. When the final explanation, $E$, is synthesized, it should accurately address the original question, $uq$. Further, since there is a diverse set of explainer output types (see Tab. \ref{tab:MetaExplainer-delegate}), we had to allow for the output data frames to reflect this diversity. Hence, we support handlers for the various explainer output types in the synthesis stage. We save the explainer outputs in folders titled with a combination of the explanation type, the explainer method, and a unique timestamp since several questions can be addressed by the same explanation type and explainer methods. In the future, we can also include information about the user session in the persistence strategy. The Synthesis stage reads from these output folders. 

\begin{table}[hbt!]
\centering
\caption{Evaluations supported within MetaExplainer for the following explanation modalities.}
\label{tab:modality-metrics}
\begin{tabular}{p{3cm}p{3cm}p{8cm}} 
\toprule
\textbf{Modality} & \textbf{Metric} & \textbf{Definitions} \\ 
\midrule




Representative Samples & Diversity~\cite{nguyen2020quantitative} & Diversity is the distance between example points. \\
Representative Samples & Non Representativeness~\cite{nguyen2020quantitative} & High non-representativeness, however, can also be indicative of factual inaccuracy. A highly diverse set of examples demonstrates the degree of integration of the explanation. \\
Feature Importances & Faithfulness\cite{alvarez2018towards} & Are “relevant” features truly relevant? \\
Feature Importances & Monotonicity\cite{nguyen2020quantitative} & The monotonicity for feature attributions $a_{i}$ is defined as the Spearman’s correlation coefficient $\rho_S$ (a, e) [28]. a = (. . . , $\|a_{i}\|$, . . . ) is a vector containing the absolute values of the attributions. e = (. . . , E(l($y_\ast$, fi); Xi|$x_\ast - i$), . . . ) contains the corresponding (estimated) expectations, as computed in Eq.(1). \\
Rules & Fidelity\cite{van2021evaluating,saad2007neural} & The approximation should correctly capture the black box model behavior in all parts of the feature space. \\
Rules & Average Rule Length\cite{van2021evaluating, lakkaraju2017interpretable} & The average number of antecedents, connected with the AND operator, of the rules contained in each ruleset Lakkaraju et al.~\cite{lakkaraju2017interpretable}. $a_i$ represents the number of antecedents of the ith rule and R = $\|$A$\|$ the number of rules. \\
\bottomrule
\end{tabular}
\end{table}

\subsection{Synthesis} \label{sec:MetaExplainer-synthesis}
The Synthesis stage aims to combine and synthesize the explainer output (s) into natural-language explanations aligned with the identified explanation's expected templates. For example, a contrastive explanation should contain facts, foils, or both supporting model predictions. However, the explainer outputs from Delegate are in data frames and not in natural language; hence, in the Synthesis stage, we need to generate natural-language outputs from data frames. This task can be broken down into several steps (see Listing \ref{lst:synthesize}), including a retrieval step where we extract relevant data points from the explainer outputs to include in the final explanation, an augmentation/alignment step where we align the outputs to fit the identified explanation type's templates, and a generation step where we output natural-language explanation ($\explanation_c$) populated with the retrieved content in line with the template for the explanation type ($et$). The synthesis task lends itself well to an application of the widely popular Retrieval-Augmented-Generation (RAG)~\cite{gao2023retrieval}, ~\cite{lewis2020retrieval} technique of prompting LLMs to output relevant and high-precision answers. 


\begin{algorithm} [hbt!] 
\caption{\texttt{Synthesis}: Combine outputs from different explainers for different data and explanation types to generate natural-language explanations.}  
\label{lst:synthesize}
	\begin{algorithmic}[1]   
	\Require Explanation Ontology (EO)
        \State Explanation types, $t = \{t_i, \forall i \in N\}$
		\State Explainers, $em = \{em_j, \forall j \in M\}$
        \newline
		
     \Input
          \Desc{$eq$}{Explainer outputs against explanation types}
          \Desc{$upp$}{User persona and scenario preference}
    \EndInput  \newline
  
		\State Set $E$ to NULL
              \For {Every $t_i$ in $\{t_a\}$}
                  \State Explainer output $\{eq\}_a$ = Fetch all $\{em_j(.|d_k)\}$ from $eq(t_a)$ 
                 
                  \State Rank $\{eq\}_a$ by accuracy / confidence scores / explainability metrics 
                  
                  \State Filter $\{eq\}_a$ as $\lbrace eq_c, \text{ s.t. } eq_c \in \{eq\}_a \text{ and } \forall c \in \{ 1 \dots C \}\rbrace$ \Comment{\textbf{Retaining top-C answers}}
                  
                 
                  \For {$eq_c$ in $\{eq\}_a$}
                      \State Explanation $\explanation_c$ = Slot fill for $t_a$'s template from $eo_{c}$ content in EO KG 
                      \State Update $E = E \cup ({\explanation}_c; t_a, em_j, d_k, rq_a)$
                  \EndFor
                  
                  \State 
              \EndFor
              
              
               \Output
                    \Desc{$E$}{List of explanation tuples}
             \EndOutput
                    
     \end{algorithmic} 
\end{algorithm}

The synthesis stage of the MetaExplainer is more streamlined than the other stages in that all three steps - retrieval, augmentation, and generation- can be easily chained together. Suppose we are not responding to a live user question and are synthesizing explanations for questions from $qb$ for which the explainers have already been run; in that case, the synthesis module reads from the output folders of the Delegate stage. We read explainer outputs $eq$ from the Delegate output folders. If we were responding to user questions, $uq$, we would still read from the same output folder but wait for Delegate's response before we can run Synthesis. Also, as mentioned in the Delegate, Sec. \ref{sec:delegate}, the explainer outputs are tied to the feature groups $fg$, identified in the machine interpretation parse of the reframed question, $rq$. Hence, we need to read from possibly one or more result sub-directories. Additionally, each explainer also outputs explanations in different modalities, and therefore, in the retrieval step, we handle this diversity among explainer outputs, $eq$. We also rank the explainer outputs by the corresponding weight metrics if applicable, otherwise we use auxillary metrics from Tab. \ref{tab:modality-metrics}, within their associated data frames.

The augmentation step retrieves explanation templates for the identified explanation type, $et$. Here, we run a SPARQL query to retrieve the definitions against each explanation type. Further, we tailor the expected explanation output based on the explainer outputs, i.e., we search for the expected explainer output modality (e.g., Case-based explanations require cases and hence the result frames are processed accordingly before passing to the generation step).

\begin{figure}[hbt!]
\centering
\includegraphics[width=1.0\linewidth]{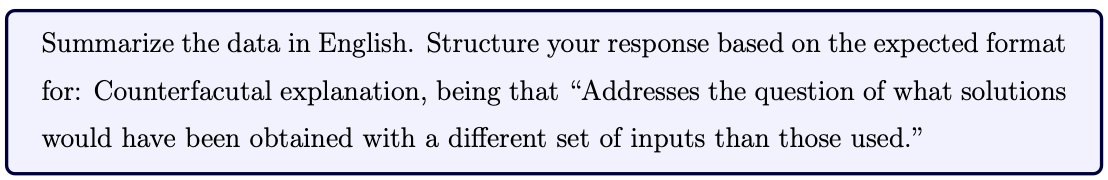}
\caption{Prompt example for generating explanations of filtered datasets for a counterfactual explanation example, ``Why rely on a BMI of 24 for predicting Diabetes in a 40-year-old male with a Diabetes Pedigree Function of 0.15, rather than a BMI of 18 in a 50-year-old female with a Diabetes Pedigree Function of 0.3?''}
\label{fig:prompt4}  
\end{figure}

In the generation step, we construct prompts to retrieve content from the explainer output, $eq$ data frames to generate explanations along explanation templates $et$. The prompts are run against the LLM Indexer (e.g., LlamaIndex~\cite{llamaindex},~\cite{langchain}) executed on the combined explainer outputs to generate the final explanation (generation). The prompts are built from the outputs of the previous stages of the MetaExplainer including the retrieved explanation template ($et$), the feature groups $fg$ and the original user question, $uq$, and run against the combined explainer outputs. A combination of the user question $uq$ and/or feature groups (Listing \ref{lst:rag-prompt-filter}) are passed as a guide to apply filters on the explainer outputs (Listing \ref{lst:rag-prompt-counterfactual}) if they were missed by the Delegate stage. We only pass the $uq$, if the machine interpretation didn't capture a feature restrictions and we perform these checks by comparing the feature groups and $uq$. 

\begin{figure}[hbt!]
\centering
\includegraphics[width=1.0\linewidth]{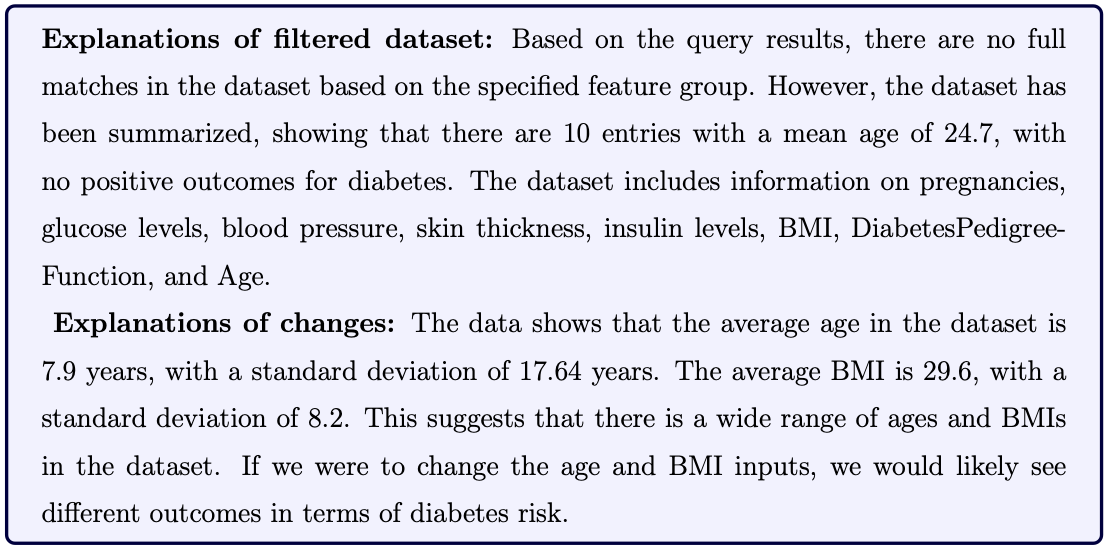}
\caption{Prompt example for generating explanations of counterfactual results to address question, ``Why rely on a BMI of 24 for predicting Diabetes in a 40-year-old male with a Diabetes Pedigree Function of 0.15, rather than a BMI of 18 in a 50-year-old female with a Diabetes Pedigree Function of 0.3?''}
\label{fig:prompt5}  
\end{figure}

Additionally, we applied RAG techniques on the result and filtered dataset frames to summarize the subset of records that matched the feature groups in the question and explain the explainer outputs that address these feature groups. A sample of the natural-language explanation for a counterfactual question is shown in Listing \ref{lst:rag-output}. 

\begin{figure}[hbt!]
\centering
\includegraphics[width=1.0\linewidth]{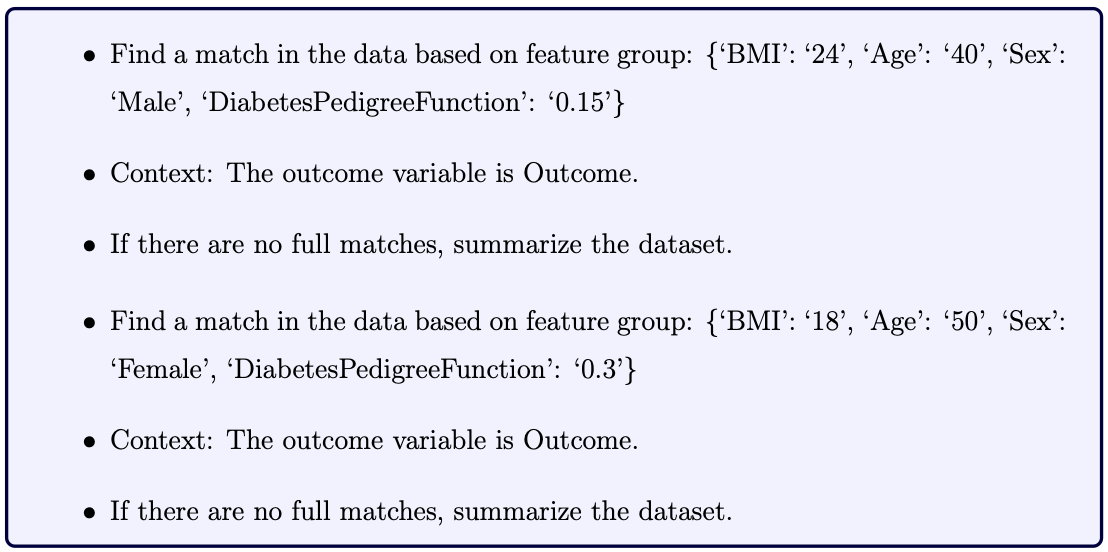}
\caption{Explanation outputs from running RAG on result dataframes of counterfactual explainer method - DiCE~\cite{mothilal2020explaining}.}
\label{fig:prompt3}  
\end{figure}

Finally, the natural-language $\explanation$ is appended with additional provenance, including the identified explanation type $et$, explainer methods $em$ run, and the reframed question $rq$ to populate explanation tuple, $E$; this provenance-affixed explanation can help users better understand how the explanation was generated and potentially where the explanation failed~\cite{mcguinness2004explaining, mueller2021principles} if it is not the expected explanation.

\section{Results}\label{sec:MetaExplainer-results}

\subsection{Data Sources} \label{sec:MetaExplainer-datasets}
For the purpose of demonstration of the utility of the MetaExplainer in a high-precision use case such as healthcare, we wanted to select a comprehensive and simple dataset to be readily used in a single-variable classification task, e.g., whether a patient has Diabetes. We found the Pima Indians Diabetes Dataset~\cite{smith1988using}, a well-cited resource released and collected by the National Institute of Diabetes, Digestive and Kidney Diseases among high-risk Diabetes Mellitus ethnic Pima tribe women at least 21 years or older in New Mexico and Arizona regions in 1988. The dataset could be considered small (~768 records) in today's deep learning age but is still sufficient to apply simpler machine learning models like logistic regression and decision tree classifiers. Recently, Chang et al.~\cite{chang2023pima} published their findings on which models work best with this dataset and we used their findings to decide upon three local models (Logistic Regression - LR, Decision Tree - DT and Random Forest - RF) to apply at each hospital node. 

Further, the Pima Dataset is a dataset with 768 rows and 9 columns - 8 features and 1 outcome variable (see Tab. \ref{tab:dataset-stas}). While there were no direct missing column values in the dataset, like non-numeric values, there were zeroes for columns that weren't collected or were missing for the patient instance. We applied data imputation techniques to fill in the median value of the column for zero values in columns.

\begin{table}[hbtp!]
\centering
\caption{Statistics of patient records in the PIMA Indian Diabetes dataset.}
\label{tab:dataset-stas}
\begin{tabular}{|l|l|l|l|}
\hline
Column                     & Count & Mean   & Standard Deviation (SD) \\ \hline
Total                      & 768   &        &                         \\ \hline
Glucose                    &       & 121.65 & 30.4                    \\ \hline
Blood Pressure             &       & 72.38  & 12.09                   \\ \hline
Skin Thickness             &       & 27.34  & 9.22                    \\ \hline
Insulin                    &       & 94.56  & 105.54                  \\ \hline
BMI                        &       & 32.45  & 6.87                    \\ \hline
Diabetes Pedigree Function &       & 0.47   & 0.33                    \\ \hline
Age                        &       & 33.24  & 11.76                   \\ \hline
Outcome - yes Diabetes     & 500   &        &                         \\ \hline
Outcome - no Diabetes      & 268   &        &                         \\ \hline
\end{tabular}
\end{table}

\subsection{Decompose} \label{sec:decompose-results}
\begin{table}[ht]
    \centering
       \caption{Explanation types of GPT-generated questions, $uq$ and their counts in the Decompose stage of the MetaExplainer.}
    \label{tab:explanation_counts_gpt}
    \begin{tabular}{lc}
        \toprule
        \textbf{Explanation Type} & \textbf{Count} \\
        \midrule
        \textbf{Data Explanation} & \textbf{80} \\
        \textbf{Case Based Explanation} & \textbf{60} \\
        Rationale Explanation & 50 \\
        Contextual Explanation & 35 \\
        Contrastive Explanation & 29 \\
        Counterfactual Explanation & 25 \\
        \bottomrule
    \end{tabular}
 
\end{table}
We evaluate the Decompose stage by how similar the reframed questions $rq$ generated by our fine-tuned LLMs are to the verified $rq$: $uq$ question pairs (see Tab. \ref{tab:explanation_counts_gpt} for distribution of $uq$). We use F1 style metrics for the natural-language fields (Action, Machine Interpretation, and Likelihood) in $rq$ and report the classification accuracy for explanation types. Among the F1 metrics, we calculate the F1 by exact match~\cite{schneider2024evaluating} and Levenshtein distances. While Llama2 (see Tab. \ref{tab:confusion_matrix-llama2}) is slightly better at identifying explanation types than Llama3 (see Tab. \ref{tab:confusion_matrix-llama3}) for certain explanations (e.g., contrastive explanations), we choose Llama3 outputs over Llama2 since the parsing accuracy (see Tab. \ref{tab:results-decompose-llama3}) is far better. Although these results are reported for a single data set in the diabetes domain (PIMA Indians Diabetes~\cite{bennett1971diabetes}), we plan to generate more $uq$:$rq$ pairs for another tabular data set (Sec.\ref{sec:MetaExplainer-datasets}), German credit risk, and we believe that these results will improve with more examples to fine-tune the models.

\begin{table}[hbt!] 
\centering
\caption{Confusion matrix metrics for identification of different explanation types on $53$ sample test set, by Llama-3 fine-tuned model in the Decompose stage of the MetaExplainer.}
\label{tab:confusion_matrix-llama3}
\begin{tabular}{lcccc}
\toprule
\textbf{Explanation Type} & \textbf{Precision} & \textbf{Recall} & \textbf{F1-Score} & \textbf{Support} \\
\midrule
Contextual Explanation     & 0.50 & 0.67 & 0.57 & 6  \\
Data Explanation           & \textbf{0.80} & 0.62 & 0.70 & 13 \\
Contrastive Explanation    & 0.00 & 0.00 & 0.00 & 5  \\
Case Based Explanation     & 0.50 & 0.07 & 0.12 & 14 \\
Rationale Explanation      & 0.69 & \textbf{0.92} & \textbf{0.79} & 12 \\
Counterfactual Explanation & 0.50 & 1.00 & 0.67 & 3  \\
\midrule
micro avg                  & 0.60 & \textbf{0.51} & \textbf{0.55} & 53 \\
macro avg                  & 0.50 & 0.55 & 0.47 & 53 \\
weighted avg               & 0.57 & \textbf{0.51} & \textbf{0.48} & 53 \\
\bottomrule
\end{tabular}
\end{table}

\begin{table}[hbt!]
\centering
\caption{Confusion matrix metrics for identification of different explanation types on $53$ sample test set, by Llama-2 fine-tuned model in the Decompose stage of the MetaExplainer.}
 \label{tab:confusion_matrix-llama2}
\begin{tabular}{lcccc}
\hline
\textbf{Explanation Type} & \textbf{Precision} & \textbf{Recall} & \textbf{F1-Score} & \textbf{Support} \\ \hline
Contextual Explanation    & 0.00               & 0.00            & 0.00              & 6                \\ 
Data Explanation          & 0.67               & 0.46            & 0.55              & 13               \\ 
Contrastive Explanation   & 0.50               & 0.20            & 0.29              & 5                \\ 
Case Based Explanation    & \textbf{1.00 }              & 0.14            & 0.25              & 14               \\ 
Rationale Explanation     & \textbf{1.00 }              & 0.17            & 0.29              & 12               \\ 
Counterfactual Explanation& \textbf{1.00}               & \textbf{1.00 }           & \textbf{1.00 }             & 3                \\ 
\hline
\textbf{micro avg}        & 0.70               & 0.26            & 0.38              & 53               \\ 
\textbf{macro avg}        & 0.69               & 0.33            & 0.39              & 53               \\ 
\textbf{weighted avg}     & 0.76               & 0.26            & 0.35              & 53               \\ \hline
\end{tabular}

\end{table}

\begin{table}[hbt!] 
\centering
\caption{Performance metrics for text fields from the Llama2 fine-tuned model in Decompose stage of the MetaExplainer.}
\label{tab:results-decompose-llama2}
\begin{tabular}{lccc}
\toprule
\textbf{Field} & \textbf{F1 (\%)} & \textbf{Precision (\%)} & \textbf{Recall (\%)} \\
\midrule
\multicolumn{4}{l}{\textbf{F1 Exact Match scores on text fields}} \\
\midrule
Machine Interpretation & 46.34 & 61.29 & 37.25 \\
Action                 & 57.74 & 57.64 & 57.84 \\
Likelihood             & \textbf{50.14} & \textbf{59.35} & \textbf{43.40} \\
\midrule
\multicolumn{4}{l}{\textbf{F1 Levenshtein scores on text fields}} \\
\midrule
Machine Interpretation & 11.32 & 11.32 & 11.32 \\
Action                 & 8.25  & 9.09  & 7.55  \\
Likelihood             & 47.17 & 47.17 & 47.17 \\
\midrule
\multicolumn{4}{l}{\textbf{Exact match on text fields}} \\
\midrule
Machine Interpretation & \multicolumn{3}{c}{23.17} \\
Action                 & \multicolumn{3}{c}{28.87} \\
Likelihood             & \multicolumn{3}{c}{\textbf{25.07}} \\
\bottomrule
\end{tabular}
\end{table}

\begin{table}[hbt!] 
\centering
\caption{Performance metrics for text fields from Llama3 fine-tuned model used in Decompose stage of the MetaExplainer.}
\label{tab:results-decompose-llama3}
\begin{tabular}{lccc}
\toprule
\textbf{Field} & \textbf{F1 (\%)} & \textbf{Precision (\%)} & \textbf{Recall (\%)} \\
\midrule
\multicolumn{4}{l}{\textbf{F1 Exact Match scores on text fields}} \\
\midrule
Machine Interpretation & 59.06 & 55.91 & 62.58 \\
Action                 & 57.48 & 50.00 & 67.60 \\
Likelihood             & \textbf{81.46} & \textbf{84.34} & \textbf{78.77} \\
\midrule
\multicolumn{4}{l}{\textbf{F1 Levenshtein scores on text fields}} \\
\midrule
Machine Interpretation & 18.87 & 18.87 & 18.87 \\
Action                 & 19.23 & 19.61 & 18.87 \\
Likelihood             & \textbf{81.13} & \textbf{81.13} & \textbf{81.13} \\
\midrule
\multicolumn{4}{l}{\textbf{Exact match on text fields}} \\
\midrule
Machine Interpretation & \multicolumn{3}{c}{29.53} \\
Action                 & \multicolumn{3}{c}{28.74} \\
Likelihood             & \multicolumn{3}{c}{\textbf{40.73}} \\
\bottomrule
\end{tabular}
\end{table}

\subsection{Delegate}
\begin{table}[ht]
    \centering
    \caption{Summary of inputs to the Delegate stage by explanation type breakdowns. The explanation types with the highest counts are highlighted in bold.}
    \label{tab:parse_summary}
    \begin{tabular}{ll}
        \toprule
         \multicolumn{2}{c}{\textbf{Parse stats for mode: Generated by fine-tuned Llama3 model}} \\
        Usable passes & 221 \\
        \multicolumn{2}{c}{\textbf{Number of explanation types}} \\
        Explanation type & Count \\
        \midrule
        \textbf{Rationale Explanation} & \textbf{61} \\
        \textbf{Data Explanation} & \textbf{55} \\
        Counterfactual Explanation & 33 \\
        Contextual Explanation & 28 \\
        Contrastive Explanation & 25 \\
        Case Based Explanation & 19 \\
        \midrule
        Length of unusable parses & 58 \\
        \midrule
        \multicolumn{2}{c}{\textbf{Parse stats for mode: human-verified GPT generated}} \\
        \textbf{Usable passes} & \textbf{279} \\
        \multicolumn{2}{c}{\textbf{Number of explanation types}} \\
        Explanation type & Count \\
        \midrule
        \textbf{Data Explanation} & \textbf{80} \\
        Case Based Explanation & 60 \\
        \textbf{Rationale Explanation} & \textbf{50} \\
        Contextual Explanation & 35 \\
        Contrastive Explanation & 29 \\
        Counterfactual Explanation & 25 \\
        \midrule
        \textbf{Length of unusable parses} & \textbf{0} \\
        \bottomrule
    \end{tabular}
\end{table}

We evaluate the Delegate stage by the parser, model trainer and executor sub-stages (see Sec. \ref{sec:delegate}), i.e., how many of the $rq$ were we able to identify explainers for and what are the performance of these explainers and the ML models they explain. We report the number of usable parses for the generated and fine-tuned generation settings of Decompose (see Sec. \ref{sec:decompose-results}) and further break down the results by the explanation types (Tab. \ref{tab:parse_summary}). As can be seen from Tab. \ref{tab:parse_summary}, there are no unusable parses from the human-verified GPT parses but about $21\%$ of the parses from Llama fine-tuned model are unusable (some of these can be fixed by improvements we discuss in Sec. \ref{sec:MetaExplainer-discussion}) and further the distributions for the explanation types match in the human-verified and fine-tuned settings corroborating results we see in Sec. \ref{sec:decompose-results}. Finally, we choose to explain a Logistic Regression classification model, as it performs the best, Tab. \ref{tab:model_performance_delegate}, on the PIMA Indians Diabetes Dataset (Sec. ~\ref{sec:MetaExplainer-datasets}).

\begin{table}[hbt!]
    \centering
       \caption{Model performance metrics on PIMA Indians Diabetes dataset.}
    \label{tab:model_performance_delegate}
    \begin{tabular}{lccccc}
        \toprule
        \textbf{Model} & \textbf{Precision} & \textbf{Recall} & \textbf{F1} & \textbf{Sensitivity} & \textbf{Specificity} \\
        \midrule
        Logistic Regression & \textbf{0.77} & \textbf{0.77} & \textbf{0.77} & 0.61 & \textbf{0.86} \\
        Decision Tree & 0.73 & 0.73 & 0.73 & \textbf{0.63} & 0.79 \\
        Random Forest & 0.75 & 0.75 & 0.75 & \textbf{0.63} & 0.82 \\
        \bottomrule
    \end{tabular}
\end{table}

We evaluate the explainer methods by metrics recommended in the explainable AI community for the modalities outputted by the methods (see Tab. \ref{tab:modality-metrics}). For example, Nguyen and Martinez~\cite{nguyen2020quantitative} recommend using a combination of diversity and non-representativeness for samples outputted by example-based explainers~\cite{zhou2021evaluating}. Further, we report the explainer method performance scores for the parses outputted by the fine-tuned Llama3 model (Sec. \ref{sec:decompose-results}), since the MetaExplainer is expected to respond to a user question in real-time. 

\begin{table}[hbt!] 
    \centering
     \caption{Summary of metrics for explanation modalities outputted by the Delegate stage of the MetaExplainer.}
       \label{tab:metrics_summary}
    \begin{tabular}{lrrr}
        \toprule
        \textbf{Metric} & \textbf{Mean values} & \textbf{Modality} & \textbf{Explainer Method} \\
        \midrule
        Average rule length & 2.39 & Rules & RuleXAI\\
        Fidelity & 0.31 & Rules & RuleXAI  \\
        Non representativeness & 0.026 & Samples & Protodash and DiCE\\
        Diversity & \textbf{340.96} & Samples & Protodash and DiCE\\
        \textbf{Faithfulness} & \textbf{0.71} & Features & SHAP  \\
        Monotonicity & 0.095 & Features & SHAP  \\
        \bottomrule
    \end{tabular}
\end{table}

There is a general lack of existing explainer evaluation benchmarks to compare our scores. However, some resources exist; for example, Vilone and Longo~\cite{vilone2021quantitative} benchmarked various rule-based explainer metrics for $15$ datasets and found a correlation between the average rule length and fidelity. As seen from Tab. \ref{tab:metrics_summary}, in our case, the rule length is lower (\textbf{$2.39$}) than reported in their paper~\cite{vilone2021quantitative}, and hence, our fidelity (\textbf{$0.31$}) is in-line with the average-performing explainers in their survey. Additionally, our faithfulness scores align with the higher values (\textbf{$0.71$}) expected of feature-based explainers' performance; the monotonicity is worrying. There are no reported scores for example-based metrics; however, from the definition of the diversity and non-representativeness metrics~\cite{nguyen2020quantitative}, we can deduce that a diverse and representative set of examples is preferred. Although a high non-representativeness value can also indicate factual inaccuracies since this metric depends on the loss value, so a lower value of non-representativeness, such as ours ($0.09$), can be tolerated. 

\subsection{Synthesis}
We evaluate the natural-language explanations generated by the Synthesis stage by determining how relevant the retrieved answers are to the contexts of the passed delegate result and user question, $uq$. Since we don't have annotated ground truth examples of reasonable explanations for the different explanation types that we have chosen and generated, we utilize unsupervised metrics from the RAG community to assess the relevance, faithfulness, and precision of the explanation to the context and question. Since, in synthesis, we set up the explanation generation task as an RAG problem (Sec. \ref{sec:MetaExplainer-synthesis}), we can leverage RAG-based metrics. We use the RAGAS Python library~\cite{ragas2024},~\cite{es2023ragas} to evaluate responses from the Synthesis module. RAGAS implements the LLM-as-a-judge technique~\cite{zheng2024judging} wherein widely-used LLMs such as GPT-3.5 Turbo and other HuggingFace hosted LLMs such as Llama3 can be used to evaluate the performance of other LLMs. Since our result data frames from Delegate are in structured formats, we convert them to text to populate the contexts used to evaluate the explanations. Further, for each $uq$, we consider the explanations of the filtered subsets and result dataframes separately since the contexts for these explanations are different. The results from Synthesis are as seen in Tab. \ref{tab:synthesis-results}, as can be seen the context-utilization ($0.92$) and relevance scores ($0.68$) are high indicating that the RAG model is not hallucinating however the faithfulness score ($0.07$) is low indicating that the answers can have broader coverage of the contexts.

\begin{table}[hbt!] 
    \centering
    \caption{Results of RAG metrics~\cite{es2023ragas} for small-set of natural-language explanations generated by the Synthesis stage of the MetaExplainer.}
    \label{tab:synthesis-results}
    \begin{tabular}{lr}
        \toprule
        \textbf{Metric} & \textbf{Value} \\
        \midrule
        Faithfulness & 0.071 \\
        Answer relevance & 0.68 \\
        Context-utilization & 0.92 \\
        \bottomrule
    \end{tabular}
\end{table}

\section{Discussion} \label{sec:MetaExplainer-discussion}
MetaExplainer offers a robust framework to respond to user questions with a set of user-centered explanation types (e.g., Contrastive Explanations, Counterfactual Explanations, etc.), and the explanations are enhanced with the provenance of intermediate outputs from each stage, ensuring transparency and traceability. Further, each stage within the framework can be evaluated using quantitative metrics (Sec. \ref{sec:MetaExplainer-results}), allowing for precise performance assessment. The modular nature of the framework will enable developers to integrate additional explainers, explanation types, and data modalities, providing flexibility and scalability. We ensure high-quality explanations by leveraging state-of-the-art (SOTA) methods, including top-performing large language models (LLMs)(e.g., Llama family~\cite{touvron2023llama} and GPT 3.5 Turbo), our widely-cited Explanation Ontology~\cite{chari2023explanation},~\cite{chari2020explanation}, and explainer techniques (e.g., SHAP, Protodash, etc. - Tab. \ref{tab:MetaExplainer-delegate}). We make available the codebase as an open-source repository, facilitating easy adoption and implementation by the community.
\subsection{Evaluation}
The MetaExplainer comprises different stages and AI methods; hence, we evaluate the performance at each stage and ideally at each method to help trace the causes of errors and ensure that each stage performs as intended. While, in our current evaluation (Sec. \ref{sec:MetaExplainer-results}), we have used standard metrics for the methods leveraged at each stage of the MetaExplainer, an overall evaluation of the explanations generated by the MetaExplainer can be helpful. Fewer metrics are proposed for natural-language explanations that are primarily objective, but this is an avenue for future research. For example, we could expand on the general principles of faithfulness, fidelity, and simplicity~\cite{zhou2021evaluating},~\cite{agarwal2022openxai} proposed for model-based explanations (Tab. \ref{tab:modality-metrics}) and apply them to the different types of user-centered natural language explanations, and these could help provide notions of principles such as reliance, performance, and explanation goodness, highlighted in Hoffman et al.'s well-cited explanation evaluation measures paper~\cite{hoffman2023evaluating}. Small-scale user studies can also provide perspectives on end-user satisfaction with explanations such as mental models and their overall satisfaction with the generated explanations. Another exciting area to evaluate is how close the explanation is to addressing the intent of the user question, $uq$.
\subsection{Ease of Implementation}
As mentioned and demonstrated in Sec. \ref{MetaExplainer:methods}, the design and implementation of the MetaExplainer are kept modular to enhance scalability and easier debugging. We also implemented the MetaExplainer on a well-thought-out pseudocode designed by collaborators on the HEALS project. Hence, this helped us ensure that we have a reliable blueprint for the expected outputs and inputs of each stage of the MetaExplainer. Additionally, the Explanation Ontology (EO) offers different structural pieces about explanations (Chapt. \ref{chapt:explanation_ontology}) that we leverage in the three stages of the MetaExplainer, i.e., prototypical questions mapped against each explanation type in Decompose,  explainer methods, corresponding modalities and metrics in Delegate and finally expected structure templates in Synthesis. The increasing capabilities of LLMs across a range of tasks, such as question-answering, generation, and retrieval-augmented-generation (RAG), also helped our implementation of the MetaExplainer, as we were able to leverage both the generation and RAG capabilities to generate a question bank of user questions to respond to and perform relevant retrieval operations on explainer outputs.

Further, while we support six out of the fifteen explanation types (Tab. \ref{tab:MetaExplainer-delegate}) in the current version of the MetaExplainer, developers can add support for more explanation types if they have/know explainer methods that can generate model explanations for the additional types. The explainer method mappings can be added to EO, and corresponding explainer methods should be defined within the Delegate folder of the MetaExplainer codebase. Alternatively, if developers want to apply the MetaExplainer to new use cases, they would need to re-run the question generation stage of Decompose to use the features from their dataset. 
\subsection{Reliably Using LLMs}
LLMs, while powerful, are known to have several issues, including hallucinations~\cite{rawte2023survey} and grounding~\cite{ye2023effective}. Since we use LLMs for a trustworthy AI pillar~\cite{trustworthy-ai-nist} to support the eventual generation of explanations, we adjusted our techniques in stages where we use LLMs within the MetaExplainer to reduce these issues. In the Decompose stage, we use in-context learning~\cite{dong2022survey},~\cite{min2022rethinking}, providing a few examples of prototypical questions for each explanation type, and we find that GPT can reproduce the question types for different feature combinations from the PIMA dataset (Sec. \ref{sec:MetaExplainer-datasets}). Further, in the decompose stage, we also fine-tune the Llama family of LLMs to help them generate machine-interpretation parses of the natural-language questions. We find that the Llama models can be leveraged to create reliable parses (Tab. \ref{tab:results-decompose-llama3}), but their performance is variable in identifying the explanation type. However, if we had a more balanced question split across explanation types (Tab. \ref{tab:explanation_counts_gpt}), the Llama model would better identify explanation types. In the delegate stage, we handle the ambiguity in machine interpretation formats in the parsing sub-stage. Finally, in the Synthesis stage, we reduce the hallucination tendencies of the LLM by using the RAG technique and also passing to the RAG LangChain agent~\cite{topsakal2023creating},~\cite{langchain} a set of high-quality data frames from explainer methods that already address the user question, $uq$ and its machine interpretation, $rq$. Furthermore, since we maintain intermediate outputs at each stage of the MetaExplainer, it is easy to debug if the error was caused by the stages that use LLMs or elsewhere, and this also helps ground the outputs of LLMs.
\subsection{Benefits of an Ontology-enabled Implementation}
In the EO, we encode dependencies of explanations such that they can be easily generated from their components if they are available to the AI system, e.g., the data points within the datasets, the knowledge used by the AI methods, and the predictions of classification/regression models and outputs of explainer methods. In version 2 of the EO~\cite{chari2023explanation}, we added support to represent a taxonomy of explainer methods and their classes of outputs~\cite{arya2019one}, i.e., Local versus Global explanations. Hence, we were able to represent outputs of AIX-360 explainer methods~\cite{arya2022ai}. However, in version 3, we have further edited and added to the EO to extract useful mappings within the MetaExplainer, including explanation type-explainer method, explainer method-explanation modality, and modality-metrics linkages. These additions not only make the EO more easily queryable through code but also improve the coverage of the EO in terms of explainable AI literature~\cite{zhou2021evaluating},~\cite{vilone2021quantitative}. 

Using the EO also helps make the MetaExplainer more scalable, better maintainable, and modular than other related works, such as the Megaexplainer~\cite{slack2023explaining}, wherein the authors define several mapping and configuration files to achieve the same mappings. Also, an ontology approach provides additional capabilities, such as representing explainer outputs in knowledge graphs~\cite{edge2024local}, allowing for capturing more connections and providing structured formats for trying different methods, such as embedding-based approaches for LLM-based generation~\cite{pan2024unifying}. The obvious benefit of using an ontology-enabled implementation is better provenance management. 
\subsection{Potential for Improvements}
Here, we have described an end-to-end implementation of the MetaExplainer and demonstrated its capabilities on a single yet representative tabular dataset, the PIMA Indians Diabetes dataset. Beyond this thesis, we plan to evaluate the MetaExplainer on more tabular datasets typically used for evaluating fairness and explainability algorithms such as the German Credit Risk Prediction~\cite{hofmann1994statlog} and COMPAS~\cite{fabris2022algorithmic} datasets. This is minimal effort since the current set of tabular prediction models and explainer methods we support within Delegate are domain-agnostic and will extend to other tabular datasets. However, we will have to generate a new set of questions for each dataset and verify them, which is not time-consuming and does not require more than five hours. Further, we also plan to expand the MetaExplainer to non-tabular datasets such as text datasets, and this would involve support text prediction models and text explainers (some of the explainers we support have demonstrated use for text, e.g., SHAP~\cite{lundberg2017unified}, RuleXAI~\cite{macha2022rulexai}).

Another direction for expansion is to support more explainer methods for each explanation type (Tab. \ref{tab:MetaExplainer-delegate}) (e.g., support LIME~\cite{ribeiro2016should} in addition to SHAP for Contrastive explanations) and also for more explanation types (e.g., Contextual explanations we couldn't find a reliable open-source explainer). We will need to add to the current suite of explainer implementations in Delegate and new evaluations if these methods produce explanations beyond the supported modalities of instances, rules, and features (Tab. \ref{tab:modality-metrics}). In addition to the data-based explainers, we could also explore integrating knowledge explainers, such as our contextualization framework from Chapt. ~\ref{chapt:qa_contextualization}; however, the evaluation metrics are more challenging to define here - but we can explore proxies such as the quantitative ones~\cite{zhou2021evaluating},~\cite{vilone2021quantitative}, ~\cite{lakkaraju2017interpretable} we currently support. 

We can continually improve upon the evaluation scores of the different stages of the MetaExplainer (Sec. \ref{sec:MetaExplainer-results}) to ensure better, faithful, and more reliable performance. For example, we could experiment with various fine-tuning strategies for LLMs in Decompose and increase the training data size via data augmentation, benchmark and experiment with several parsers and explainer methods and choose the best ones in the Delegate and also benchmark several LLMs for the RAG portion of Synthesis. We could also vary the intermediate representation formats at each stage and identify if they improve performance or, eventually, better explanations. For example, we could consider using knowledge graphs as inputs to the RAG pipeline (similar to Edge et al.~\cite{edge2024local}) in Synthesis and investigate if the RAG approach can generate explanations better based on KG representation of explainer outputs, and this is a relatively easy next step given that we already leverage the EO for several mappings in the MetaExplainer. 

Finally, a framework such as the MetaExplainer, which is user-centered, can benefit from improving and adapting to user feedback. In our original plan for the MetaExplainer, we had two additional stages: inferring which explanation types to provide users based on their saved persona characteristics and persisting user feedback for explanations after they have been generated. However, we decided to leave these stages out of the first implementation of the MetaExplainer, given the challenges of finding relevant users (which, with the PIMA Indians setting, could be clinicians, clinical researchers, or medical students), conducting and establishing user sessions. However, if the MetaExplainer were applied in a project where there is an ongoing collaboration with domain experts, it would be worthwhile to conduct user interviews~\cite{wang2019designing},~\cite{gruen2021designing} to assess the utility and quality of MetaExplainer's explanations and use this feedback to improve the implementation based on what users want. 
\section{Conclusion}
The MetaExplainer implements a vision that we had and a need that we gathered from clinician conversations of supporting and generating user-centered explanations along various types and that address several types of user questions (e.g., What Ifs, What instances, Why this and not That). We have designed and implemented the MetaExplainer as a modular three-stage framework, allowing us to leverage/adapt existing AI methods within each stage and effectively utilize the explanation dependencies defined within the EO. The three stages of the MetaExplainer include Decompose: converting user questions to machine-interpretable equivalents, Delegate: executing explainer methods for the identified explanation type using constraints from the machine interpretation, and Synthesis, generating natural-language explanations from explainer outputs. In this chapter, we have provided sufficient details of our design choices, implementation, and functionality we support within the MetaExplainer framework so that interested system developers could adapt the MetaExplainer to their use cases to serve as an explanation interfacing component for the end-users interacting with the predictions of the ML methods in their AI systems. Further, the modular design of the MetaExplainer provides several avenues for adaptability and improvement; for example, if developers want to add more explanation types/explainer methods, they can do so by using our GitHub repository. In summary, the MetaExplainer is an essential step towards providing end-users multiple ways to probe AI systems to explain their predictions along various types. 

\chapter{DISCUSSION, LIMITATIONS AND FUTURE WORK} \label{chapt:Discussion}
This chapter discusses themes and challenges around the end-user explainability of AI systems. These themes summarize the larger goals in explainability, including the purpose they serve, the dimensions that should be included, such as domain knowledge and contexts, and challenges, such as evaluation of explanations and combination strategies. In our discussions of these themes, we indicate how we have solved some of the challenges under these themes through our three contributions and also discuss what issues remain. 

\section{User-centered Aspects of Explainability and the Challenges}

Explanations are pivotal in helping end-users interact with AI systems in multiple ways. They allow users to trust, debug, or be educated about AI systems~\cite{mittelstadt2019explaining},~\cite{doshi2017accountability}. Hence, due to the wide variety of purposes, explanations serve the range of questions they address, and their presentations are varied. Further, end-users have different needs for explanations; for example, clinicians have different domain expertise and purposes for explanations, as postulated by Dey et al.~\cite{dey2022human}. In this thesis, we make contributions towards supporting user-centered explanations in many ways, including representing different explanation types that address various user questions (such as what ifs, why this or not that, what evidence, etc.), connecting model explanations to authoritative contexts and designing a framework that users can interactively ask questions to and are provided with the best-fit explanations. However, there are still opportunities to support end-users with better needs for explanations. 

\subsection{Future Work for Supporting User-centered Aspects of Explainability}
One of the challenges that remain is capturing user feedback~\cite{lakkaraju2022rethinking} to improve and learn about the presentation of explanations~\cite{krishna2023towards},~\cite{krishna2022disagreement} that end-users prefer and prioritizing important dimensions such as actionability and causality as properties within explanations. Further, the various needs that end-users have for explanations indicate that capturing a representation of user personas~\cite{dey2022human} can be helpful to tailor explanations to the needs of end-users. One way we can capture user personas is by adding user and persona-related characteristics to the user portion of the EO. This would then help us add learning and feedback components to the MetaExplainer such that explanations can be tailored to end-user personas. Additionally, techniques such as active learning~\cite{ghai2021explainable},~\cite{liang2020alice} and uncertainty quantification~\cite{ghosh2022uncertainty} can both help tailor explanations to users' preferences and better quantify the bounds of explanations to help users trust and understand the explanations better. Additionally, end-user explainability is tied to explaining the ever-changing and complex landscape of AI methods, so while the user-centered explanation types might remain the same, the content that is populated into these explanations will have to evolve. For example, safety and performance explanations (see Chapt. \ref{chapt:explanation_ontology} for a list of all explanation types) of LLMs will be different from those of simpler ML models such as decision trees. 


Overall, there are several challenges and opportunities to support explanations for end-users, and they encompass and go beyond dimensions we support in the user-centered explanations in this thesis, i.e., we support the knowledge-enabled and context-aware nature of explanations (Sec. \ref{sec:discussion-knowledge}). However, other dimensions, such as tailoring explanations to users' preferences and personas, lead to exciting research directions that will eventually make explanations more valuable and trustworthy for the end-users of AI systems. 

\section{Grounding Explanations in Domain Knowledge and Context} \label{sec:discussion-knowledge}
Several researchers~\cite{ghassemi2021false, miller2019explanation, doshi2017accountability} propose that explanations need to be grounded in the context of application and domain knowledge to be more usable and actionable. From our interactions with clinicians during requirement gathering and evaluation sessions (see Chapt. \ref{chapt:qa_contextualization}), we find that they often refer to the expertise and literature they are familiar with when reasoning through patient cases, their predictions and explanations. Hence, grounding explanations in domain knowledge and contexts is an essential problem to address to improve the uptake of explanations. Approaches such as our risk prediction dashboard that shows several pieces of information, including AI model predictions, post hoc explanations of patient predictions, and literature-derived contextual explanations along the same screen, can be helpful to help end users, especially domain experts, decide if they should act on AI predictions and explanations or not. Conversely, explanations can boost model predictions in the wrong direction, and rooting these explanations in domain knowledge/evidence could also help provide a negative context to help decide not to trust an AI system for a wrong decision. Further, providing multiple explanation types in a decision-making situation, such as risk prediction (see Fig. \ref{fig:needforexplanations}), can help domain experts decide and take action based on many views. For example, seeing a contrastive explanation that exposes the features that contribute to a decision versus those that contribute negatively can help clinicians assess if the model focuses on the features they know it should have utilized. Similarly, a data explanation exposing the subset of data used for the decision can further help clinicians/domain experts trust a decision. 

\subsection{Future Work for Grounding Explanations in Domain Knowledge and Context}

However, grounding explanations in domain knowledge and contexts can be a challenging and sometimes intractable problem. The problem of including domain knowledge in explanations can be solved on a per-use-case basis by identifying the sources of domain knowledge and contexts in consultation with domain experts and implementing extraction and upstream analysis pipelines to utilize the insights from the identified domain knowledge sources. There are often challenges in extracting content from domain knowledge owing to the various publication formats and licensing policies. However, utilizing and adapting a standard suite of extraction-transform-load (ETL) pipelines~\cite{mccusker2018setlr} can help make the extraction more seamless. Neuro-symbolic approaches~\cite{gaur2021semantics} such as our contextualization QA pipeline (see Chapt. \ref{chapt:qa_contextualization}) can then be implemented and applied to derive insights from the extracted domain-knowledge through a question-answering setup or other conversational approaches such as AI chatbots. These insights from the domain knowledge sources can be transformed to populate explanations of various types, including scientific, everyday, statistical, contextual, case-based explanations, etc.  

Explanations need to be rooted in or be aware of the various contexts of the AI ecosystem they are a part of. For example, in the EO, we support multiple definitions of context in line with Dey et al. ~\cite{dey1998cyberdesk},~\cite{lim2009and} including environmental, system, and application contexts (See Chapt. \ref{chapt:explanation_ontology}). An understanding of contexts addresses some of the user-centered aspects of uptake, including providing curious and satisfactory explanations, i.e., reducing information that end-users might already know and conversely including contexts that might help better reason through the predictions. McGuinness et al. ~\cite{mcguinness2007explaining} captured an encoding of contexts in their Integrated Cognitive Explanation architecture where they explain task processing for learning agents. There are also several context-aware recommender systems; adapting these principles to explanations can help make them more aligned with the cognitive and social aspects of explainability. Hence, while we have proposed a method for contextualizing risk prediction scores and feature importance using authoritative literature such as medical guidelines,  there are opportunities and strategies to apply similar techniques to other application domains. 

\section{Future Work: Disagreement and Overload Problem in Explanations} \label{sec:disagreement}
Several researchers~\cite{krishna2022disagreement},~\cite{roy2022don}, have recently found that explainer methods often disagree with each other. There is often a bias for a particular method based on the system designers’ preferences. While these issues have been identified, recent contributions in advancing user-centered explainability are typically limited to position statements~\cite{ghassemi2021false},~\cite{lakkaraju2022rethinking} or user studies~\cite{liao2022connecting}, rarely more so on implementations. 

However, a few methods exist; Schwarzschild et al.~\cite{schwarzschild2023reckoning} suggest jointly training an ensemble likearchitecture of explainer models to reduce their disagreement. However, the outputs of their jointly trained explanations have yet to be vetted via user studies, so it is hard to interpret if end-users find these explanations more valuable than those of individual explainers. Instead of jointly training to reduce disagreement, we suggest using pairwise similarity methods to understand whether explanations have a high degree of similarity. If so, include them in the same explanation. Also, Roy et al.~\cite{roy2022don} specifically present three metrics - Rank, Feature, and Sign agreements to quantify the disagreement between feature attribution local explainers such as LIME~\cite{ribeiro2016should} and SHAP~\cite{lundberg2017unified}. While this is a useful method to compute the similarity between feature attribution explainers, it doesn't extend beyond explainers that are non-feature based. 

An explanation disagreement and combination module can be a useful module to our Metaexplainer, where we could leverage and build upon these methods for feature-attribution-dependent explanations such as data and contrastive explanations and extend methods for more natural language-based explanations, such as scientific and fairness explanations.

\section{Metrics for Explainable AI} \label{sec:discussion-metrics}
Explanations, while an essential and valuable pillar of Trustworthy AI, have been hard to evaluate owing to the nature of the multiple forms/types they can take and the variety of data and AI method sources they are populated from. Explanations cannot be assessed by traditional ground-truth comparison approaches as it is hard to know all possibilities of correct explanations beforehand. However, there are a few applications where explanations can be treated as a closed-world problem, such as providing common-sense rationales for tasks performed by AI methods; in such cases, explanations similar to those in wide-coverage common-sense banks~\cite{dalvi2021explaining} can be considered precise. More generally, it is hard to accurately predict all possible explanation paths when explanations are generated for education, discovery, or debugging purposes. Hence, explanations are evaluated by several approaches, including borrowing from associated fields of cognitive sciences and psychology, especially explanations that focus on the user-centered aspects of AI~\cite{miller2019explanation},~\cite{mittelstadt2019explaining},~\cite{doshi2017accountability}.

Further, there are different evaluation approaches for model~\cite{zhou2021evaluating, vilone2021quantitative} and user-centered explanations. As we have indicated and demonstrated in the MetaExplainer chapter (Sec. \ref{sec:MetaExplainer-results}), several evaluation metrics for different explanation modalities capture how faithful or diverse the model explanations are. However, these quantitative metrics do not necessarily translate to or indicate whether end-users can benefit or are satisfied by these explanations. Further, while in the MetaExplainer, we use RAG-based metrics~\cite{ragas2024} to evaluate the quality of natural-language explanations generated by the Synthesis stage; there are opportunities to use more user-centered and purpose-oriented metrics such as those proposed by Hoffman et al.~\cite{hoffman2023evaluating}.

\subsection{Future Work for Metrics}
Hoffman et al.~\cite{hoffman2023evaluating},~\cite{hoffman2018metrics}'s widely-cited papers provide a comprehensive set of evaluations for explanations that evolved from the various explanation approaches in the DARPA XAI project~\cite{gunning2017explainable}. They have an explanation scorecard that captures important dimensions to identify how useful the explanations are to end-users including satisfaction, user trust, curiosity, mental models, etc. However, the dimensions they evaluate are subjective and complex; for example, user satisfaction and mental models are hard to capture. Still, they have associated checklists (e.g., a curiosity checklist with a list of questions that can be asked to assess the curiosity that the explanations cover) to determine the explanation's coverage in structured formats. Hence, there are opportunities to convert the explanation scorecard into an eventual explanation metrics ontology, similar to the design of the explanation ontology. Further, while we have evaluated explanations via quantitative and qualitative measures across the contextualization and MetaExplainer frameworks in this thesis, in the future, we can consider supporting multiple perspectives of explanation evaluations by leveraging potential knowledge representations of explanation measures such as an eventual explanation metrics ontology. 

In summary, it is hard to define a one-size-fits-all approach to explanation evaluations which is also similar to the need for multiple explanation types~\cite{mittelstadt2019explaining} and explanation approaches~\cite{arya2019one}. The evaluation metrics for explanations should be decided upon after considering the primary purpose the explanations are to serve in the use case and how best this purpose could be evaluated. Maintaining common leaderboards tracking metrics used to assess explanations across various use cases and approaches can help improve the standardization of explanation evaluation metrics. Such leaderboards have proven beneficial for other disruptive technologies, such as LLMs~\cite{helm2023}. 

\section{Ontology-enabled Explanations}
Thus far, in this chapter, we have discussed topic areas within Explainable AI that have been addressed to some extent by this thesis, but there are challenges that remain, and hence, we have discussed plans for addressing some of these challenges. However, we have yet to specifically discuss AI technologies and methods that have helped us support user-centered explainability. At the crux of each contribution in this thesis is the use of ontologies, either the EO we developed or other general-purpose and domain ontologies. Ontologies, or the symbolic AI method, in conjunction with ML and neural approaches, such as LLMs, can help us support the need for user-centered explainability. For example, as we have discussed through several sections before, the EO can capture user representations and hence help address user-centered challenges of explainability (Chapt. \ref{chapt:explanation_ontology}); further, the EO can be adapted to various domains to capture the explanation needs in the application domain (Chapt. \ref{chapt:qa_contextualization}), and finally, as we have seen in the MetaExplainer (Chapt. \ref{chapt:Metaexplainer}), the EO serves as a mapping for the various connecting components of explanations, including explanation methods, explanation modalities, and metrics. Hence, updating the EO, with evolving needs for explanations, is valuable to ensure minimal effort is needed to support either new explanation types/methods or implement explanation methods for a new use case. 

As Tiddi et al.~\cite{tiddi2022knowledge} outline, Ontologies and their adjacent technologies, such as KGs, can provide several benefits for user-centered explainability, including and not limited to enabling more understandable explanations as they ``provide explanations in the form of symbolic, human-readable rules'' (even a neuro-symbolic approach such as the MetaExplainer can ensure that the explanations adhere to expected patterns), the reuse large-scale knowledge graphs can help the system's accuracy (e.g., we found improvements in using knowledge-augementations to improve the semantic coherence of answers in our contextualization framework) and finally that knowledge-enabled systems ``bring reactivity'' to natural-language explanations. Hence, there is value in using ontology-enabled methods to address the critical goal of explainability, which will always be an essential pillar of Trustworthy AI~\cite{trustworthy-ai-nist},~\cite{gunning2017explainable}.

\section{Future Work: Directions for Applications of User-Centered Explanations to Domains Beyond This Thesis}
While we have focused on developing methods to represent, contextualize, and generate user-centered explanations along various literature-derived explanation types, we have also demonstrated the usage of these explanations in high-precision domains spanning finance, healthcare, and nutrition. We have supported explanations for question-driven~\cite{liao2020questioning} settings; however, as AI systems and their applications evolve, explanations could also be required in use cases where the interactions between AI systems and users are passive (e.g., self-driving~\cite{gilpin2021explaining}) or that explanations contribute insight towards the understanding of one or several hypotheses (e.g., drug discovery). Investigating and identifying the broader need for explanations in such settings is interesting. Here, we discuss two use cases from high-precision domains and motivate the need to think about supporting explanations beyond reactionary and conversational settings. 

Fields such as drug discovery, where tasks such as candidate identification inherently benefit from explanations, can significantly benefit from explainable AI approaches. However, given that drug discovery differs from typical benchmarked domains, challenges, and unmet needs exist beyond using traditional AI benchmark datasets (e.g., SQUAD~\cite{rajpurkar2016squad}). In a well-cited position paper, Jimenez et al.~\cite{jimenez2020drug} suggest principles that will be important for explainable AI approaches in drug discovery and review the useful XAI literature for the domain. The principles they identify are: ``Transparency, knowing how the system reached a particular answer; justification, elucidating why the answer provided by the model is acceptable; Informativeness - providing new information to human decision-makers; and Uncertainty estimation - quantifying how reliable a prediction is''~\cite{jimenez2020drug}. Although these principles are addressed by one or more of the explanation types we support in the EO, for example, justifications can be provided by rationale-based explanations; what is not possible yet with our three approaches is that we cannot trigger an explanation to address these principles when results of various experiments in the drug discovery pipeline are made available. However, what can be developed here is a mechanism to identify the situations or contexts that require explanations. For example, Jimenez et al.~\cite{jimenez2020drug} describe an instance of when explanations might be required when ``certain decision-making processes, such as the replacement of animal testing and in vitro to in vivo extrapolation, where human-understandable generated explanations constitute a crucial element''. A modification of the EO (Sec. \ref{sec:eomodel}) for scenario-driven explanations and representing the situations in the drug discovery pipeline where explanations might be required is beneficial. The MetaExplainer, especially the Decompose step, can then be adapted to work with results in situations that trigger the need for explanations. Instead of working with reframing questions and identifying relations, we can adapt a similar reframed question, $rq$, data frame structure (Sec. \ref{sec:MetaExplainer-decompose}) to identify what needs to be explained, the data modality, the background knowledge, and what explanation type would best work. The rest of the MetaExplainer pipeline can be leveraged as is, to generate natural-language explanations for helping domain experts at various stages of drug discovery. The challenges and opportunities to introduce explanations in drug discovery also apply to other similar scientific and multi-modal domains such as materials design. 

Explaining awareness and early warning systems in domains such as climate change~\cite{bommer2024finding} and self-driving~\cite{gilpin2021explaining} includes challenges of understanding expected behavior of systems and analyzing real-time behaviors to explain when the system is deviating from expected behavior or is making an error. Typically, counterfactual explanation techniques are suggested in real-time settings that help address the what-if questions of what might go wrong because of a system result. The GeCO interactive counterfactual explanation framework is one such system demonstrated in a loan application setting~\cite{schleich2021geco}). Further, for example, climate scientists need more robust explanations for unfamiliar situations or critical warnings; that is, suppose that an AI system finds a decline in the population of a particular animal or plant species - they require reasons as to what caused the decline. In such situations, climate scientists might want to trigger simulations for how this would negatively impact the surrounding environment (or simulation-based explanations) or ask the ``what if'' question for potential impact on human health (or counterfactual explanation) (Tab. \ref{tab:explanationtypes}). Hence, comprehensive explanations addressing various questions, such as Why, What if, and What evidence, can contextualize AI system decisions and enhance usability. Moreover, the emergence of AI dashboards~\cite{wang2019designing, liao2022connecting} provides an opportunity to present explanations and seamlessly integrate AI methods into practical applications. 

In a similar setting, Kuznetsov et al.~\cite{kuznietsov2024explainable} outline the requirements to generate autonomous driving explanations that involve various stages of model development and provide explanations at different stages of these pipelines. These stages include interpretable models, interpretable design and surrogate models, interpretable monitoring, auxiliary explanations, and interpretable safety validation. Regarding supporting auxiliary explanations, Gilpin et al.~\cite{gilpin2021explaining} design a taxonomy of explaining errors to generate natural-language explanations of errors that occur at various levels of self-driving systems. This notion of hierarchies in systems can be added to the system portion of the EO model (Sec. \ref{sec:eomodel}), such that the origin of ``system recommendation'' being explained can be traced back. Furthermore, contextual explanations (Chapt. \ref{chapt:qa_contextualization}) that link to published literature, such as manufacturer manuals and in the context of climate change - papers and reports, can be helpful to alert end-users with contexts about their situation. A system such as the MetaExplainer (Chapt. \ref{chapt:Metaexplainer}), which provides multiple explanations, is beneficial. However, the MetaExplainer must be adapted to work with the outputs of various AI methods, including several prediction and surrogate models and additional outputs from multiple data sources, including potentially real-time data streams. Hence, in real-time settings,  there are several interesting challenges for explanations, and they can take various forms, including reactive, multi-dimensional, and modal- explanations. 
\chapter{CONCLUSION} \label{chapt:Conclusion}
In conclusion, user-centered explainability is crucial due to its impact on AI systems' transparency, trust, and usability. In this thesis, we make three contributions to advance the field and understanding of computational methods to support user-centered explainability, ranging from methods to represent, contextualize, and combine explanations. We first reviewed the trustworthy AI literature comprehensively to identify foundations and directions for knowledge-enabled explanations. We identified where the gaps are at to make our contributions, and we defined explanations to include essential dimensions such as data, knowledge, and context. We then developed a semantic representation, the Explanation Ontology (EO), to capture explanations' supporting components, including their user-, system- and interface- dependencies. We have released multiple iterations of the EO by adding support for more explanation components, such as more explanation types, explainer methods, and explanation modalities and metrics. In our second contribution, we developed an end-end contextualization pipeline to provide explanations from high-quality content such as medical guidelines, to aid clinicians in decision-making in a comorbidity risk prediction setting. In the contextualization pipeline, we implement a question-answering task and leverage large language models (LLMs) and their clinical variants in conjunction with domain ontologies to provide contextual explanations that help clinicians reason through feature importance and risk prediction scores. We evaluated the QA by both quantitative and qualitative metrics and identified essential dimensions, such as actionablity, that clinicians would want to see highlighted in explanations. Finally, as the last contribution, we designed a general-purpose framework, MetaExplainer, that generates natural language explanations in response to user queries. In the modular and three-stage MetaExplainer framework, we leverage our EO, state-of-the-art explainer methods, and LLMs to identify the relevant explanation types to address user questions (decompose), execute the corresponding explainer methods (delegate), and finally populate explanations tailored to the questions posed (synthesis). Our contributions significantly advance the field, and we have shared our findings through several publications and associated documentations on supporting user-centered and knowledge-enabled explanations through ontology-enabled AI approaches. 
 
%
\specialhead{REFERENCES}
\bibliographystyle{IEEEtran}           
\bibliography{main}        

 
\begin{appendices}
\chapter{CONTEXTUALIZATION MODEL EXPLANATIONS VIA A \\ KNOWLEDGE-AUGMENTED QUESTION-ANSWERING METHOD}
\section{QA Architecture}
\label{sec:qamethod:appendix}

Here we describe the processes and modules, seen in part B) of Fig. \ref{fig:guidelineqa}, that are involved in generating questions based on question types for each patient (see Tab. \ref{tab:questiontypes}) and the answers from the different domain sources of context in our risk prediction setting.

\subsection{Question Generation}
The \textit{question generation module} almost always creates templated questions using Python's native support for String Templates,~\footnote{\url{https://docs.python.org/3/library/string.html}} and does so based on patient data, more specifically from the patient's diagnoses codes, lab values, and medication list. The patient's diagnoses codes are sometimes abstracted from their higher-level disease groupings supported by the CCS scheme. A subset of these diagnosis codes can be included in the features that the post-hoc explanation module found were contributing to the patient's predicted risk. In an attempt to provide more context around these features, we create instances of the type 3 question , e.g., ``What can be done for this patient's essential hypertension?'' We also support the creation of two standard, non-variant questions for each patient, i.e., whose values don't change from patient data, that can help clinicians easily interpret their predicted risk (question type 1) and their {\dm} state (question type 2).

Moreover, as can be seen from Tab. \ref{tab:questiontypes}, each of the question types that we support on a per patient basis is populated from different data sources. Hence, we have developed different answering methods for each, including simple lookups and knowledge augmented language model capabilities, including combinations of either a LLM + value range comparison or LLM + knowledge augmentation. We provide examples of questions and answers for each question type in \ref{sec:result_settings:appendix}.

\subsection{Answer Generation}
In our answer generation module of our QA approach, we support different submodules that can output answers to questions related to the question types. The answer generation module is capable of inputting questions generated by the previous question generation module and interacting with extracted content from our supported data sources. 

\subsubsection{Template-based Answer Generation} Question types 1 and 2 from Tab. \ref{tab:questiontypes} can be addressed by simple query lookups of our supported data sources. We populate the Python String Template object with the results of the queried components retrieved by using the widely-used Python Pandas library~\cite{mckinney2011pandas}. This process of creating natural language templates that can then be populated with values on a per-patient basis is supported by the Template-based Answer Generation module of our QA pipeline (Fig. \ref{fig:guidelineqa}). The results of these questions can be summarized or built from structured datasets, like patient data, their model outputs, like risk predictions, and features contributing to their predicted risk and population averages. Hence, there is no fuzziness in the results, which is why we don't evaluate the accuracy of this submodule. 

This submodule is also leveraged in combination with other answer generation submodules when there is a pattern in the answers and slots to be filled. We discuss these details shortly after we set up our knowledge augmented language model capabilities and their usage. 

\subsubsection{Numerical Range Comparison} BERT LLMs cannot currently determine if a question that has a numerical value comparison, e.g., ``What can be done for patients, whose Hemoglobin A1C $>$ 10?'', falls in the range of the answer returned. However, clinicians often look for recommendations that match patients' lab values in clinical settings such as ours. Further, within the ADA 2021 guidelines, there are mentions for suggestions based on different ranges for lab values, e.g., a recommendation from the Pharmacological Chapter of these guidelines has a recommendation with the mention of ``when A1C levels ($>10\%$ $[$ 86 mmol/mol $]$." Hence, for question type 3 from Tab. \ref{tab:questiontypes}, we need to determine if the patient's lab values are in the range of the answer returned by the LLM module. To address this requirement of performing numerical range comparison between the question and answer produced by LLM, we leverage syntactic parsing capabilities (similar to ~\cite{chen2021personalized}) to identify numerical phrases in the question and answer and then determine if each numerical phrase from the question is in range of the same in the answer. 

We use Natural Language Toolkit (NLTK) chunking and parsing functionalities to identify noun phrases, comparatives, and numerical mentions within both the question and answer. We then write regular expression (regex) rules to identify the patterns of the positional tags returned by NLTK that can constitute numerical phrases. For each of the numerical phrases, we convert them into a tuple of ``(noun phrase, [upper bound, lower bound]).'' This tuple representation allows us to go through the phrases between the question and answer iteratively, and for those that match on the noun phrase dimension, identify if the ranges are in agreement. With these steps, we can then populate an answer using the Template-based Answer generation module, which says if the answer outputted by LLM is in/out of the range of the question. Hence, in this manner, we enhance the capabilities of BERT LLMs for numerical range comparisons via rule-based syntactic methods, which is also why we consider this step a rule augmentation of LLMs' capabilities.

\subsubsection{Knowledge Augmentations to LLMs} 
Transformer based LLM approaches like BERT work on sequences of words that are often seen together and their surrounding words, but don't leverage the semantics of whether these words are diseases, medications, or biological processes. We found that in the absence of this semantic knowledge, we would often get answers from the LLM that don't correlate on a semantic level with the question. For example, a sentence from the Comorbidities chapter of the ADA 2021 CPG on Dementia was returned as a valid answer to a question asking about an Abdominal Hernia. To eliminate such answers, we explored options for a biomedical semantic mapper and zeroed in on the National Library of Medicine (NLLM)'s Metamap tool~\cite{aronson2010overview}. We choose Metamap because of its extensive coverage of biomedical semantic types and its ability to capture entity mentions within the ADA 2021 CPG. Within our pipeline, we have integrated a Python wrapper for Metamap\footnote{PyMetamap: \url{https://github.com/AnthonyMRios/pymetamap}} that can recognize biological entities within the guideline text and their semantic types (e.g., dsyn: disease or syndrome, bpoc: biological processes, etc. for a complete list of types returned by Metamap see: \footnote{\url{https://lhncbc.nLLM.nih.gov/ii/tools/MetaMap/Docs/SemanticTypes_2018AB.txt}}).

We run Metamap on question types 3 and 4 from Tab. \ref{tab:questiontypes} to only output answers from BERT when there is a valid semantic match between the question and answer. Specifically, using this knowledge augmentation module, for question type 3, we only output answers whose matched term is a noun and is recognized as a disease term by Metamap, and similarly, for question type 5, we only output answers whose matched term is a noun and is recognized as medication by Metamap. We have observed that depending on the mention of a biological entity in the text, a disease term can be recognized as a disease, biological process, or a finding by Metamap. Hence, we allow for flexibility among semantic types, in filtering disease matches for question type 3. For example, we want to allow answers with the mention of the term `hypertensive' for a question on hypertension, although hypertensive is identified as a finding by Metamap.

Additionally, given this ability to filter based on semantic types, we want to allow additional answers with mentions of related diseases. To provide more broad answers, we use the UMLS Concept Unique Identifier (CUI) codes from the Metamap returned outputs to map to Snomed-CT disease codes~\cite{donnelly2006snomed}. From the mapped Snomed-CT disease codes, we can traverse the Snomed-CT disease tree to identify how many hops apart question and answer disease codes are and if the answer codes are an ancestor of those in the question. We operate on the idea that answers about the parent disease code apply to children nodes. For example, a question about ``What can be done for Asthma'' can borrow from an answer on ``What can be done for respiratory diseases?'' Conversely, if disease codes in the question and answer are far apart in the Snomed tree, it would signify that they are semantically less related. In addition to Metamap codes, we append to the candidate guideline sentences the hop distances from each question computed by applying the Dijkstra's algorithm~\cite{dijkstra1959note} on an uploaded Snomed graph in Python package NetworkX~\cite{hagberg2008exploring} and ancestor values derived from using Python library PyMedTermino's~\cite{lamy2015pymedtermino} is\_ancestor function.

We use the outputs of these knowledge augmentation modules to both pre-filter and post-sort the LLMs answers. The LLM and LLM + post sorting settings 1, 2 and 5, were run against $410$ passage chunks of guideline text, of average length $267$ tokens, since BERT has a 512 token limit for an answer passage. The LLMs on the pre-filtering settings 3 and 5 were run on passage chunks of variable length, depending on the number of filtered sentences to be passed to the LLM model. In the pre-filtering settings, we varied the values of the features that we were filtering by to understand which feature values improve accuracy. In essence, the pre-filtering settings can be thought of as algorithmic knobs to control the set of answers that the LLM has to process. In contrast, in the post-filtering settings we sorted the LLM's answers by feature values, and here we could control the ordering of answers to be outputted. In the pre-filtering setting 2, we filter the guideline sentences by length of disease overlap with the question. In pre-filtering setting 4, we have more possibilities in the feature column because the number of Snomed disease hops between a question and answer can range between a continuous range of integer values. We report if restricting the number of hops to allow for more general yet precise answers improves accuracy. Similarly, in the post-sorting settings, 2 and 4, we use the feature values from the knowledge augmentation modules in addition to the LLM's own confidence scores to rank answers. Specifically, in setting 2, we sort the LLMs answerset on variations to a combination of length of disease overlap between question and answer Metamap phrases and the LLM confidence scores. In setting 5, we sort the LLM's answerset based on variations to a combination of sum of hops between question and answer Snomed disease codes, number of Snomed ancestors in the answer and LLM confidence scores. 

We report the accuracies for answers that address questions of question types 3 from Tab. \ref{tab:questiontypes}, that use these knowledge augmentations in the results section (Sec. \ref{sec:results}). We have written functions that use the NLTK toolkit in our evaluation submodule to generate standard, natural language processing (NLP), accuracy scores like F1, precision, recall, and BLEU. Overall, the integration of a semantic mapping tool helps us enhance the capabilities of BERT LLMs for more precise and better semantic matches via knowledge-driven methods.

\section{Results}
\label{sec:result_settings:appendix}
Here we present additional material to support the results of QA module described in Sec. \ref{sec:guidelineqaresults}. 

\begin{table}[!htbp]
\centering
\caption{Performance of Guideline QA with different language model approaches reported at mean average precision (map), F1 and recall at top-10 answers and precision at top-1 and top-5 for $12$ expert validated questions. Best and second-best values for each column is highlighted in green and blue color, respectively.}
\label{tab:expertresults-native}
\small
\small
\begin{tabular}{lrrrrrr} 
{} & {bleu} & {P@1} & {P@5} & {map} & {f1} & {recall} \\
{model} & {} & {} & {} & {} & {} & {} \\
BERT & \color{green} 0.155 & \color{blue} 0.363 & \color{green} 0.262 & \color{green} 0.267 & \color{blue} 0.224 & \color{blue} 0.365 \\
BioBERT & 0.121 & 0.296 & 0.200 & 0.222 & 0.186 & 0.342 \\
BioBERT-BioASQ & 0.131 & 0.259 & 0.200 & 0.205 & 0.192 & 0.363 \\
BioClinicalBERT-ADR & 0.112 & 0.227 & 0.178 & 0.179 & 0.171 & 0.351 \\
SciBERT & \color{blue} 0.153 & \color{green} 0.366 & \color{blue} 0.216 & \color{blue} 0.244 & \color{green} 0.235 & \color{green} 0.463 \\
\end{tabular}
\end{table}

\subsection{Guideline QA Results on Expert Validated Set}
As seen in Tab. ~\ref{tab:expertresults-native} and Tab. ~\ref{tab:expertresults-kmidiseases}, we provide results numbers on a small set of expert validated answers from our guideline annotations of $12$ questions and $47$ candidate answers. We are considering methods like weak supervision to increase the expert validation coverage of our annotations. 

\begin{table}[!htbp]
\centering
\caption{Performance of Guideline QA with different language model approaches + knowledge augmentations reported at mean average precision (map), F1 and recall at top-10 answers and precision at top-1 and top-5 for $12$ expert validated questions. Best and second-best values for each column are highlighted in green and blue color, respectively. Language model (e.g. BERT) suffixed with KA represents the corresponding knowledge augmented model (e.g. BERT-KA).}
\label{tab:expertresults-kmidiseases}
\begin{tabular}{lrrrrrr}
{} & {bleu} & {P@1} & {P@5} & {map} & {f1} & {recall} \\
{model} & {} & {} & {} & {} & {} & {} \\
BERT-KA & 0.021 & 0.296 & \color{green} 0.296 & \color{green} 0.296 & 0.127 & 0.081 \\
BioBERT-KA & 0.143 & \color{green} 0.440 & 0.215 & 0.258 & 0.222 & 0.319 \\
BioBERT-BioASQ-KA & \color{blue} 0.147 & \color{blue} 0.366 & \color{blue} 0.249 & 0.272 & \color{blue} 0.227 & 0.335 \\
BioClinicalBERT-ADR-KA & 0.123 & 0.321 & 0.209 & 0.221 & 0.201 & \color{blue} 0.384 \\
SciBERT-KA & \color{green} 0.201 & 0.356 & 0.209 & \color{blue} 0.284 & \color{green} 0.297 & \color{green} 0.600 \\
\end{tabular}
\end{table}

We find that generally the results on the expert validated answers (Tab.~\ref{tab:expertresults-native} and ~\ref{tab:expertresults-kmidiseases}) follow the trend of accuracy values in the larger annotation set, in that the precision is highest in the pre-filtering by number of Snomed disease hop settings (setting 4) - $0.29$ and that the recall and bleu is high in post-filtering by number of disease overlaps between question and answer (SciBERT + KA in Tab. \ref{tab:expertresults-kmidiseases}), setting 3 - $0.6$ and $0.2$ respectively. However, contrary to the larger set of results the F1 are highest in both the vanilla BERT and BioBERT-BioASQ + KA settings, $0.22$. 

\chapter{PERMISSION FILES}
The following three permission files include license details and terms for reproducing contents in this dissertation.

\section*{Permission for Chapters 2 and 3}
\noindent Permissions for re-using the paper ``Explanation ontology: A general-purpose, semantic representation for supporting user-centered explanations'' in Chapter 2.1 and Chapter 3.

\noindent\textbf{File name:} IOS\_permissions.pdf \\
\textbf{File type:} Portable Document Format (PDF)\\
\textbf{File size:} 135 KB \\
\textbf{Required application software:} Adobe Acrobat or other standard PDF viewer \\
\textbf{Special hardware requirements:} None

\section*{Permission for Chapters 2 and 4}
\noindent Permissions for re-using the paper ``Informing clinical assessment by contextualizing post-hoc explanations of risk prediction models in type-2 diabetes'' in Chapter 2.2 and Chapter 4.

\noindent\textbf{File name:} Elsevier\_permissions.pdf \\
\textbf{File type:} Portable Document Format (PDF)\\
\textbf{File size:} 82 KB \\
\textbf{Required application software:} Adobe Acrobat or other standard PDF viewer \\
\textbf{Special hardware requirements:} None

\section*{Permission for Chapters 3}
\noindent Permissions for re-using the paper ``Explanation ontology: A model of explanations for user-centered A'' in Chapter 3.

\noindent\textbf{File name:} Springer\_permissions.pdf \\
\textbf{File type:} Portable Document Format (PDF)\\
\textbf{File size:} 29 KB \\
\textbf{Required application software:} Adobe Acrobat or other standard PDF viewer \\
\textbf{Special hardware requirements:} None

\end{appendices}

 
\end{document}